\pdfoutput=1
\documentclass{article} %
\usepackage{proactegomem,times}

\usepackage{hyperref}
\usepackage{url}
\usepackage[all]{hypcap}

\usepackage[T1]{fontenc}
\usepackage[utf8]{inputenc}
\usepackage{amsthm,amsmath,amsfonts,amssymb}
\usepackage{pifont}
\usepackage[dvipsnames]{xcolor}
\usepackage{graphicx}
\usepackage{wrapfig}
\usepackage{fontawesome5}
\usepackage{colortbl}
\usepackage{booktabs}
\usepackage{changepage}
\usepackage{tabularx}
\usepackage{caption}
\usepackage{mdframed}
\usepackage{tcolorbox}
\tcbuselibrary{breakable,skins}
\usepackage{textcomp}
\definecolor{StanfordRed}{HTML}{8C1515}
\definecolor{orange}{HTML}{FF8000}
\definecolor{citegray}{HTML}{666666}

\paperfinalcopy %
\newif\ifpaperpreprint
\paperpreprinttrue

\usepackage{amsmath,amsfonts,bm}

\def\eqref#1{equation~\ref{#1}}

\def\1{\bm{1}}

\def\ra{{\textnormal{a}}}
\def\rb{{\textnormal{b}}}

\DeclareMathAlphabet{\mathsfit}{\encodingdefault}{\sfdefault}{m}{sl}
\SetMathAlphabet{\mathsfit}{bold}{\encodingdefault}{\sfdefault}{bx}{n}

\usepackage{subcaption}
\usepackage{algorithm}
\usepackage{algorithmic}
\usepackage{mathrsfs}
\usepackage{setspace}
\usepackage{mathtools}
\usepackage{multirow}
\usepackage{array}
\usepackage{placeins}    %

\usepackage{needspace}
\usepackage{tikz}
\usetikzlibrary{calc,arrows.meta,shadows,positioning,fit,backgrounds,decorations.markings}
\usepackage{pgfplots}
\pgfplotsset{compat=1.18}
\usepackage{xspace}
\usepackage{enumitem}

\newcommand{\vctask}{\textsc{IC}}
\newcommand{\sttask}{\textsc{ST}}
\newcommand{\lttask}{\textsc{LT}}
\newcommand{\na}{--}
\newcommand{\method}{\textbf{ProactEgoMem}\xspace}
\newcommand{\cmark}{\ding{51}}
\newcommand{\xmark}{\textcolor{red}{\ding{55}}}

\newcommand{\dailymethod}{\textbf{ProactEgoMem-Daily}\xspace}
\newcommand{\scimethod}{\textbf{ProactEgoMem-Sci}\xspace}
\newcommand{\dailyclips}{300}
\newcommand{\dailyclipspertask}{100}
\newcommand{\dailyvideos}{239}
\newcommand{\dailyhours}{16.00}
\newcommand{\dailymeanclip}{192.0}
\newcommand{\dailysegments}{1242}
\newcommand{\dailyclosed}{924}
\newcommand{\dailyopen}{318}
\newcommand{\dailyavgseg}{31.82}
\newcommand{\dailymedseg}{13.47}
\newcommand{\dailyobjects}{292}
\newcommand{\dailyqa}{1000}
\newcommand{\dailysegic}{278}
\newcommand{\dailysegst}{483}
\newcommand{\dailyseglt}{481}
\newcommand{\dailyqaic}{507}
\newcommand{\dailyqast}{241}
\newcommand{\dailyqalt}{252}
\newcommand{\sciclips}{100}
\newcommand{\sciicclips}{30}
\newcommand{\scistclips}{35}
\newcommand{\sciltclips}{35}
\newcommand{\scivideos}{100}
\newcommand{\scihours}{6.47}
\newcommand{\scimeanclip}{233.0}
\newcommand{\scisegments}{1377}
\newcommand{\sciclosed}{1218}
\newcommand{\sciopen}{159}
\newcommand{\sciavgseg}{13.33}
\newcommand{\scimedseg}{4.77}
\newcommand{\sciobjects}{89}
\newcommand{\sciqa}{500}
\newcommand{\scisegic}{175}
\newcommand{\scisegst}{831}
\newcommand{\sciseglt}{371}
\newcommand{\sciqaic}{299}
\newcommand{\sciqast}{104}
\newcommand{\sciqalt}{157}

\definecolor{teaserNavy}{HTML}{13243B}
\definecolor{teaserBlue}{HTML}{2E6F9E}
\definecolor{teaserTeal}{HTML}{00B894}
\definecolor{teaserSoft}{HTML}{F4F8FA}
\definecolor{teaserLine}{HTML}{BFD1DC}
\definecolor{teaserGold}{HTML}{D89A34}
\definecolor{modeBlue}{HTML}{2D75B6}
\definecolor{modeGreen}{HTML}{2F9559}
\definecolor{taskGray}{HTML}{F1F3F5}
\definecolor{qaAnswerBg}{HTML}{F3E1E4}
\definecolor{qaAnswerText}{HTML}{8A4652}
\definecolor{taskShade}{HTML}{F6EAEC}
\newcommand{\correctoption}[1]{%
  \begingroup\setlength{\fboxsep}{1.2pt}%
  \colorbox{qaAnswerBg}{\textcolor{qaAnswerText}{\strut\textbf{#1}}}%
  \endgroup}
\newsavebox{\denseqaBox}
\newcommand{\denseqa}[1]{%
  \sbox{\denseqaBox}{\strut #1}%
  \ifdim\wd\denseqaBox>\linewidth
    \resizebox{\linewidth}{!}{\usebox{\denseqaBox}}%
  \else
    \usebox{\denseqaBox}%
  \fi}
\newcommand{\tasktitle}[1]{%
  \tikz[baseline=(tasktitle.base)]
    \node[anchor=base, fill=taskShade, rounded corners=3pt,
      inner xsep=5pt, inner ysep=2pt] (tasktitle) {\textbf{#1}};%
}
\newcommand{\taskbadge}[1]{%
  \tikz[baseline=(taskbadge.base)]
    \node[anchor=base, fill=taskShade, rounded corners=2pt,
      inner xsep=3pt, inner ysep=1.5pt] (taskbadge) {\textbf{#1}};%
}
\newcommand{\toolcard}[1]{%
  \begin{tikzpicture}
    \node[fill=taskGray, rounded corners=4pt, outer sep=0pt,
      inner sep=3pt, text width=\dimexpr\linewidth-6pt\relax, font=\scriptsize,
      align=left] {#1};
  \end{tikzpicture}%
}
\newcommand{\Notif}{\textcolor{modeBlue}{Notification}\xspace}
\newcommand{\Interv}{\textcolor{modeGreen}{Intervention}\xspace}
\newcommand{\NotifLevel}{\textcolor{modeBlue}{Notification} Level\xspace}
\newcommand{\IntervLevel}{\textcolor{modeGreen}{Intervention} Level\xspace}
\newcommand{\NotifLevelAdj}{\textcolor{modeBlue}{Notification}-level\xspace}
\newcommand{\IntervLevelAdj}{\textcolor{modeGreen}{Intervention}-level\xspace}
\newcommand{\NotifLevelPlain}{Notification Level\xspace}
\newcommand{\IntervLevelPlain}{Intervention Level\xspace}

\newcommand{\IntervLevelAdjPlain}{Intervention-level\xspace}

\definecolor{takeawayBg}{HTML}{EFF7EF}
\definecolor{takeawayHead}{HTML}{1F5C35}
\definecolor{takeawayLine}{HTML}{BFDCC4}
\newcommand{\takeaway}[2]{%
\par\vspace{0.7em}\noindent
\begin{tikzpicture}
  \node[fill=takeawayBg, draw=takeawayLine, line width=0.7pt,
        inner sep=7pt, outer sep=0pt, align=left,
        text width=\dimexpr\linewidth-16pt\relax,
        drop shadow={shadow xshift=1.6pt, shadow yshift=-1.6pt,
                     fill=black!22, opacity=0.55}]
  {\small
    {\bfseries\color{takeawayHead}Takeaway: #1}\par
    \vspace{2.5pt}{\color{takeawayLine}\rule{\linewidth}{0.6pt}}\par
    \vspace{1.5pt}#2\par};%
\end{tikzpicture}
\par\vspace{0.7em}}

\newcommand{\agenttitlemark}{%
\raisebox{-0.15em}{\includegraphics[width=1.5em]{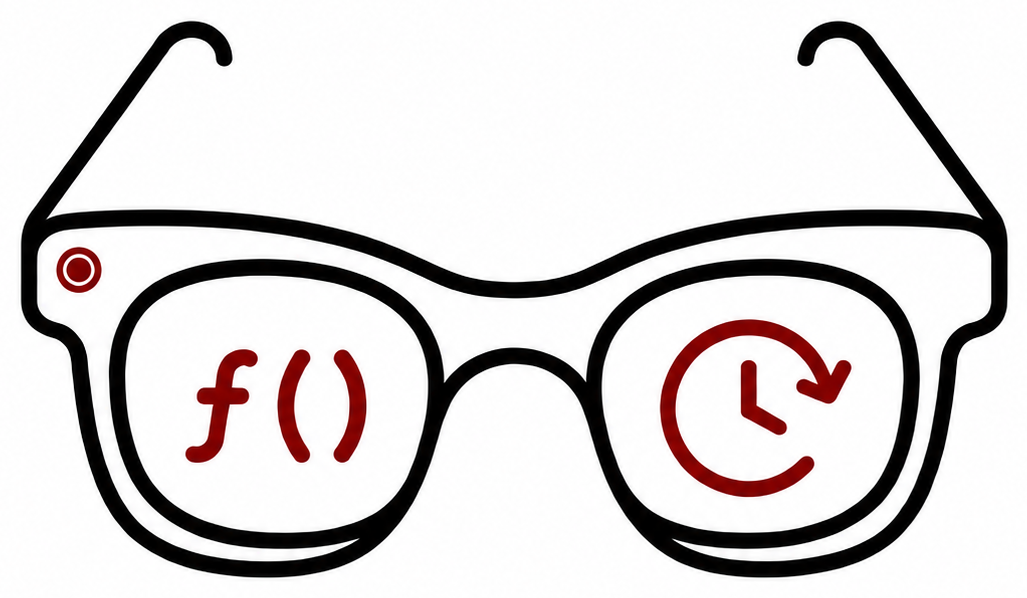}}%
}

\newcommand{\firstpageteaser}{%
\begin{minipage}{\textwidth}
\centering
\includegraphics[width=0.84\linewidth]{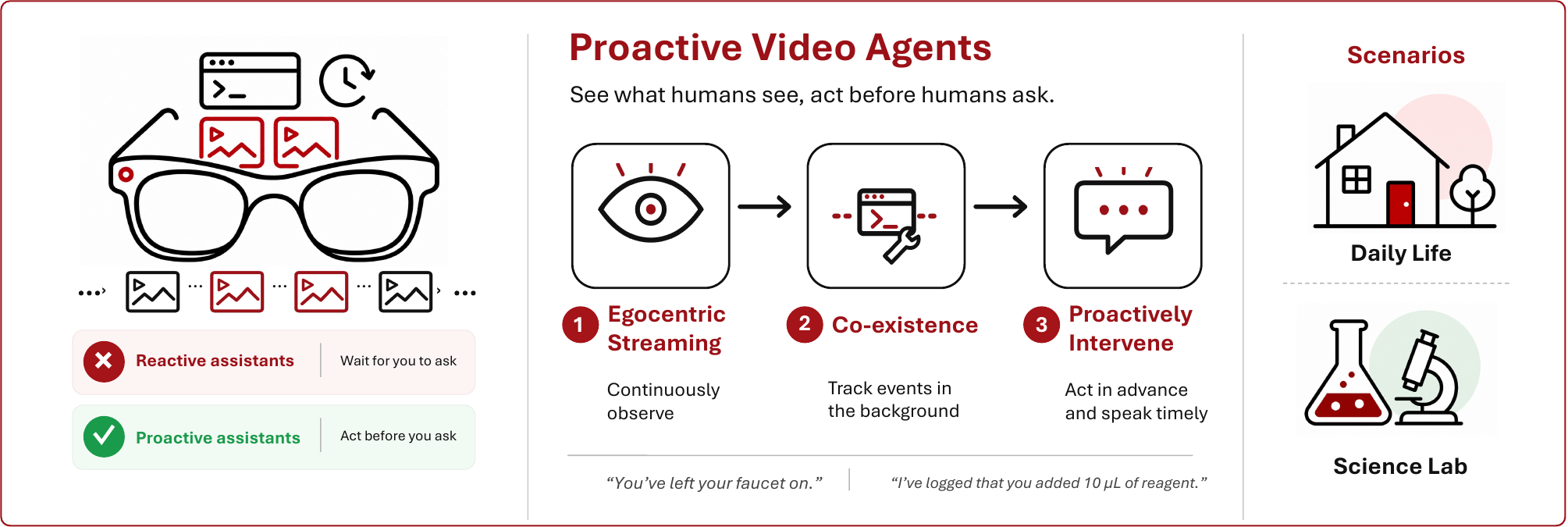}%
\captionof{figure}{\label{fig:teaser}%
Proactive agents observe video, co-exist in the background, and act before users ask.}
\end{minipage}%
}

\title{\agenttitlemark\hspace{0.25em}\method: Benchmarking Proactive Memory in Streaming Egocentric Video}

\author{%
\textbf{Ruotong Liao$^{1,*}$, \enspace Meikel Kokowski$^{4,*}$, \enspace Xiaoxiao Sun$^{2,\text{\faEnvelope[regular]}}$, \enspace Yongqin Xian$^{3}$,} \\
\textbf{Laura Bravo Sánchez$^{2}$, \enspace Shiye Su$^{2}$, \enspace Federico Tombari$^{3}$, \enspace Serena Yeung-Levy$^{2,\text{\faEnvelope[regular]}}$, \enspace Volker Tresp$^{1}$}\\
{\normalfont\small
$^{1}$LMU Munich \quad $^{2}$Stanford University \quad $^{3}$Google \quad $^{4}$Technical University of Munich}%
\ifpaperpreprint\\[2pt]
{\normalfont\small\ttfamily\hypersetup{pdfborder={0 0 0}}%
Link:\quad
\href{https://proactegomem.github.io/}{\textcolor{StanfordRed}{Project Page}}\quad|\quad
\href{https://github.com/mayhugotong/ProactEgoMem}{\textcolor{StanfordRed}{Code}}\quad|\quad
\href{https://huggingface.co/datasets/Meikel/ProactEgoMem}{\textcolor{StanfordRed}{Hugging Face}}\quad|\quad
\href{https://www.codabench.org/competitions/18340/}{\textcolor{StanfordRed}{Leaderboard}}}%
\fi}

\begin{document}
\maketitle
\lhead{Preprint under review}
\ifpaperfinal
  {\renewcommand{\thefootnote}{$*$}%
  \footnotetext{Equal contribution.}}
  {\renewcommand{\thefootnote}{\faEnvelope[regular]}\hypersetup{pdfborder={0 0 0}}%
  \footnotetext{Corresponding authors. Email: \texttt{liao@dbs.ifi.lmu.de},
  \texttt{\{\href{mailto:xxsun@stanford.edu}{xxsun},
  \href{mailto:syyeung@stanford.edu}{syyeung}\}@stanford.edu}}}
  \vspace{-16pt}
\fi

\begin{abstract}
Today's multimodal assistants are typically prompted to interpret the visual evidence presented to them. To offload the cognitive burden of remembering across long horizons, wearable assistants should proactively acquire and maintain useful memory as the user's surroundings evolve. Remembering the ambience on the user's behalf requires identifying what matters, understanding in real time, and initiating memory maintenance in the background while the user focuses on activities. We introduce \method, a benchmark that evaluates proactive memory in streaming egocentric video. Three tasks ground the evaluation in ambient environmental needs: \textsc{Information Capture} (\vctask) preserves transient yet dense visual information, \textsc{State Tracking} (\sttask) follows changing object states, and \textsc{Location Tracking} (\lttask) records object movements. The benchmark proposes the protocol at two levels of 
proactiveness: the \emph{\NotifLevel} asks for one timely notification that satisfies the user's intent; the deeper \emph{\IntervLevel} asks
for memory maintenance with an active start-to-stop tool lifecycle on the user's behalf. Parallel \dailymethod{} and \scimethod{} tracks contain \dailyclips{}/\sciclips{} clips for both daily and scientific use cases, paired with \dailyqa{}/\sciqa{} questions in ProactEgoMem-QA, which measures the quality of proactively acquired memory. Across proprietary,
open-source, video-language, and streaming baselines, even the strongest models fail to proactively capture the memory timely and accurately. \method is a rigorous evaluation foundation for multimodal
proactive assistance and co-existing video agents.
\end{abstract}

\firstpageteaser

\section{Introduction}
\label{sec:intro}

A wearable assistant on AI glasses can reduce the burden of remembering everyday details while the user attends to other activities \citep{engel2023aria,yang2025egolife}: for example, alert when a faucet remains running unattended while cooking, remembering where a tool is placed during multitasking
(Figure~\ref{fig:teaser}). Such assistance requires continuous background observation, where the assistant infers the user's intent and detects relevant events, maintains ambient environment states, and alerts or remembers proactively for later use without explicit instructions. Useful information may be transient, and the dynamically complex ambient evidence may be intertwined continuously. This motivates our central question: \emph{How can a wearable assistant proactively maintain memory in the background as the user's surroundings change over long time horizons?}

\begin{figure}[!t]
\centering
\includegraphics[width=0.98\textwidth]{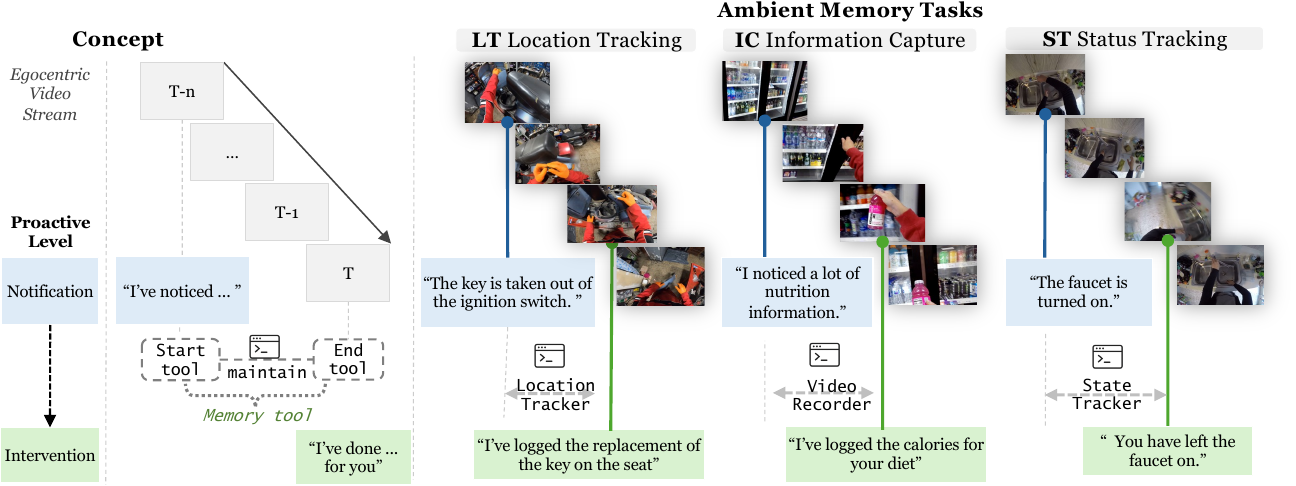}
\captionof{figure}{\label{fig:overview}
\textbf{\method measures proactive memory as lifecycle behavior in egocentric
streams.} At the \NotifLevel{} the agent sends one timely message; at the
\IntervLevel{} it starts, maintains, and ends a memory tool, and releases a message when the task is completed.}
\end{figure}
Remembering on the user’s behalf requires more than recognizing relevant information in a video. An assistant must infer intents, capture transient details before they disappear, and keep track of changing states and object locations as activities unfold. Such memory needs extend over time: the assistant must determine when to begin acquisition, maintain tracking through subsequent observations, and stop when the relevant event ends, all while the user attends to other activities. This calls for two levels of proactive assistance: notifying the user when relevant information warrants attention, and intervening on the user’s behalf to acquire it as memory.

We formulate this capability as \emph{proactive memory}: autonomously acquiring and maintaining useful evidence from an unfolding egocentric stream. Conventional video QA evaluates answers from post-hoc complete videos \citep{mangalam2023egoschema,fu2025videomme}, while streaming QA evaluates responses as video arrives \citep{lin2024streamingbench}. ProactEgoMem evaluates how useful memory is acquired in the first place, how an assistant initiates and maintains background processing and intervenes proactively, and whether the resulting memory supports subsequent assistance.

We introduce \method (Figure~\ref{fig:overview}), a benchmark for proactive memory over
streaming egocentric video, built around three object-centric tasks for ambient memories from first-person recordings:
\textsc{Information Capture} (\vctask), \textsc{State Tracking} (\sttask),
and \textsc{Location Tracking} (\lttask). We further propose to evaluate two levels of easy-to-hard proactiveness: the \emph{\NotifLevel} asks for a single timely, semantically
correct notification once the observation satisfies the user's intent; the
advanced \emph{\IntervLevel} requires the agent to track the situation in the background and operate memory tools on the user's behalf, such as maintaining lifecycles for recording or status and location tracking. %
The benchmark comprises two parallel tracks: \dailymethod{}, built from everyday
first-person recordings, with \dailyclips{} clips and \dailyqa{} questions, and
\scimethod{}, built from egocentric recordings in wet-labs as co-scientist assistants, with
\sciclips{} clips and \sciqa{} questions.

Our Benchmark makes three contributions:
\begin{itemize}[itemsep=1pt, topsep=3pt, parsep=0pt]

\item \textbf{Formulation.} We pose \emph{proactive memory} for egocentric assistants: deciding from the visual stream when to intervene and what to remember. It has three core components, streaming egocentric
perception, proactive initiative, and co-existent background
operation, framing it as object-centric ambient memory with a 
\emph{measurable memory lifecycle}.
\item \textbf{Evaluation protocol.} We design a protocol built on
multi-level proactiveness, three ambient-memory tasks, and a question
answering dataset that tests the quality of proactive memory; evaluation metrics score jointly on timing and semantics of proactive initiatives.
\item \textbf{Benchmarking.} We benchmark
proprietary, open-source, video-language, and streaming models on both daily and scientific scenarios and
identify failure modes: \emph{missed calibrations}, \emph{delayed initiation}, \emph{false alarms}, \emph{lifecycle maintenance}, and \emph{invalid tool calls}.
\end{itemize}

\section{Related Work}
\label{sec:related}

\textbf{Streaming and egocentric video understanding.}
First-person datasets \citep{9879279,damen2022epic,grauman2024egoexo4d,perrett2025hdepic}
and always-on wearables \citep{engel2023aria} support evaluation of
long-horizon egocentric QA \citep{mangalam2023egoschema,chandrasegaran2024hourvideo},
long-context video QA \citep{fu2025videomme,wu2024longvideobench,zhou2025mlvu},
streaming QA \citep{lin2024streamingbench,li2025ovobench}, and egocentric
episodic memory \citep{forte2026egostream,wang2026sember}. All of them are
query-driven (Table~\ref{tab:benchmark-comparison}); in \method the agent
decides \emph{when} to act with no query and maintains a memory across a
start--maintain--stop lifecycle.

\textbf{Video-language models.}
Proprietary \citep{openai2024gpt4o,geminiteam2024gemini15} and open
video-language models
\citep{chen2024internvl,llavaonevision_2024,llavaonevision_15_2025,lin2024videollava,wang2024qwen2vl,qwen25vl_2025,qwen35_2026,qwen36_2026,videollama3_2025,gemma4_2026}
consume a whole clip and emit an answer. Streaming systems decide online
through respond-versus-wait objectives \citep{chen2024videollmonline,wang2025mmduet},
real-time memory \citep{zhang2025flashvstream}, decoupled perception and
reaction \citep{qian2025dispider}, when-to-speak activation
\citep{kim2026stride}, offline-to-proactive conversion
\citep{wang2025streambridge}, proactive companions \citep{yan2026proactvl},
and full-duplex omni-models \citep{minicpmo45_2026,qwenteam2025qwen25omni};
each reduces proactivity to a single respond-or-not event.

\textbf{Proactive assistance and proactivity benchmarks.}
Proactive assistants span mixed-initiative interaction
\citep{horvitz1999principles,allen1999mixed}, conversational recommendation
\citep{christakopoulou2016towards,jannach2021survey}, proactive dialogue systems
\citep{deng2023proactive} and LLM agents \citep{lu2024proactiveagent}, and view-sharing wearables
\citep{yang2025egolife}, all triggered by a conversational turn; HoloAssist
\citep{wang2023holoassist} predicts intervention from first-person video
offline. Multimodal proactivity benchmarks
\citep{zhang2025proassist,wang2025omnimmi,wang2025proactivevideoqa} score one
``should I respond now?'' decision, refined by user profiles
\citep{ran2026egoprobench}, out-of-plan recovery
\citep{kundu2026planwatchrecover}, and one-second grading
\citep{du2026proactivebench}; Ego2Web \citep{yu2026ego2web} acts on user
instructions, and concurrent work studies intervene-or-abstain decisions
\citep{gong2026vinci2} and memory as an intervention mechanism
\citep{wu2026proactivememory}. \method is, to our knowledge, the first
benchmark in which proactivity \emph{produces a memory}: the agent decides
when to act, sustains a memory through a start--maintain--stop
lifecycle, and is scored on whether it answers later questions.

\section{Problem Formulation}
\label{sec:problem}

\subsection{Streaming Setting}
\label{sec:streaming-setting}

\textbf{Streaming setting.}
An episode is a timestamped stream of egocentric frames
$V=(v_1,\ldots,v_T)$. At time $t$, the agent receives only frames from oldest to newest up to the
current moment and a logical tool state. At every moment the agent decides what to initiate, operations of memory tool call in the background, a message to the user, or nothing at all.
\begin{itemize}
\item \textbf{Tool calling.}
\label{sec:tools}
The first kind of initiative is an operation of a tool call (recording,
state tracking, or location tracking) that acquires memory in the background. At every moment the agent
outputs one JSON object whose list of tool calls, possibly empty, starts or
stops tools; several tool instances may be active in parallel.

\item \textbf{Proactive message.}
The second kind of initiative is a message to the user in the \texttt{message} field: grounded in its
observations and acquired memory, the agent generates notifications or interventions. The agent keeps silent when no action is required.
\end{itemize}
\[
  \begin{aligned}
  y_t = \texttt{\{}&\texttt{"message": null | "<text>",}\\
  &\texttt{"tool\_calls": [\{"name": \ldots, "arguments": \{\ldots\}\}]\}}
  \end{aligned}
\]

\textbf{Proactive memory.}
We use \emph{proactive memory} that the agent acquires and
maintains on its own initiative: the agent decides what to
store, keeps memory tools active in the background, and surfaces the memory when the question answering (QA) warrants, with no re-inspection of the full video.

\textbf{Question answering.}
Each event is paired with multiple-choice questions
(Section~\ref{sec:qa-construction}) whose answers must be grounded in what the
agent's memory tool calls captured during the stream.

\Needspace*{25\baselineskip}
\subsection{Two Proactive Modes}
\label{sec:modes}

\begin{wrapfigure}[14]{r}{0.46\textwidth}
\vspace{-0.8em}
\centering
\resizebox{0.44\textwidth}{!}{%
\begin{tikzpicture}[
  font=\small,
  frame/.style={rounded corners=2pt, draw=black!35, fill=taskGray,
    line width=0.6pt, minimum width=1.25cm, minimum height=0.92cm,
    align=center, font=\scriptsize},
  chip/.style={rounded corners=3pt, line width=0.8pt, align=center,
    inner xsep=5pt, inner ysep=3pt, font=\footnotesize\bfseries},
  bubble/.style={rounded corners=4pt, draw=black!30, fill=white,
    line width=0.6pt, align=left, inner sep=4pt, font=\footnotesize\itshape},
  lcstep/.style={rounded corners=2pt, draw=black!45, fill=white,
    line width=0.6pt, inner xsep=4pt, inner ysep=2.5pt,
    font=\footnotesize},
  go/.style={-{Stealth[length=4pt]}, line width=0.8pt},
  lbl/.style={font=\footnotesize\itshape, color=black!65,
    fill=white, inner xsep=3pt, inner ysep=1pt},
]

\node[frame] (f0) at (0.25,-0.55) {$T{-}n$};
\node[frame] (f1) at (1.55,-0.55) {$\cdots$};
\node[frame] (f2) at (2.85,-0.55) {$T{-}2$};
\node[frame] (f3) at (4.15,-0.55) {$T{-}1$};
\node[frame] (f4) at (5.45,-0.55) {$T$};
\draw[go, draw=black!55] (-0.35,-1.18) -- (6.25,-1.18);
\node[lbl, anchor=east] at (6.25,-1.40) {time};
\node[lbl, anchor=west] at (-0.35,0.18) {egocentric stream};

\draw[dotted, draw=black!35, line width=0.7pt] (2.85,-1.18) -- (2.85,-3.55);
\draw[dotted, draw=black!35, line width=0.7pt] (5.45,-1.18) -- (5.45,-4.42);

\node[chip, draw=modeBlue, fill=modeBlue!12, text=modeBlue,
  anchor=west] (notif) at (-0.75,-2.42) {Notification};
\node[bubble, anchor=center] (nmsg) at (2.85,-2.42) {``I've noticed \ldots''};
\draw[go, draw=modeBlue] (notif) -- (nmsg);
\node[lbl, anchor=west] at ($(notif.north west)+(0,0.33)$)
  {one timely grounded message};

\node[chip, draw=modeGreen, fill=modeGreen!12, text=modeGreen,
  anchor=west] (interv) at (-0.75,-3.55) {Intervention};
\node[lcstep, anchor=center] (stop) at (5.45,-3.55) {stop tool};
\node[lcstep, left=0.26cm of stop] (keep) {maintain};
\node[lcstep, left=0.26cm of keep] (start) {start tool};
\draw[go, draw=modeGreen] (interv) -- (start);
\draw[go, draw=modeGreen] (start) -- (keep);
\draw[go, draw=modeGreen] (keep) -- (stop);
\node[lbl, anchor=west] at ($(start.north west)+(-0.05,0.4)$)
  {memory kept in the background};
\node[bubble, anchor=center] (imsg) at (5.45,-4.42) {``I've done \ldots''};
\draw[go, draw=modeGreen] (stop.south) -- (imsg.north);

\draw[dashed, draw=black!25, line width=0.6pt]
  (-0.85,-1.92) -- (6.25,-1.92);
\draw[dashed, draw=black!25, line width=0.6pt]
  (-0.85,-2.88) -- (6.25,-2.88);
\end{tikzpicture}}
\caption{\label{fig:two-modes}
\textbf{Two- level proactive modes}.}
\vspace{-0.6em}
\end{wrapfigure}
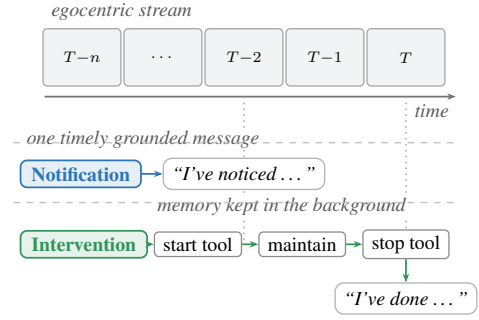
Offloading mental burden can be multi-faceted. Inspired by the distinction between proactively informing a user and acting
autonomously on their behalf \citep{kraus2021trust}, we operationalize
proactivity at two depths (Figure~\ref{fig:two-modes}; Table~\ref{tab:mode-examples}):\\
\textbf{\NotifLevel.}
The agent should infer the user's intent and send a timely notification in
response to a visual cue. For example, after inferring that the user is shopping for items, the agent sends the proactive assistant message \emph{``I noticed a lot of nutrition information''} when the user shows interest in an item. The
ground truth is a timestamped opportunity and a semantic target message.

\textbf{\IntervLevel.}
At this advanced level, the agent should infer the user's intent, act proactively on their behalf, maintain a memory tool
lifecycle for proactive memory acquisition, and report the result. For
example, when the user is performing multitasking cooking, turns on a faucet, an oven, etc, the agent starts status trackers for each object in the background,
keeps them active while the items remain on, stops them once they are
turned off, and reports that the tracking has been resolved.

\begin{table}[!t]
\centering
\caption{Illustrative examples of proactive messages for each task and mode.}
\label{tab:mode-examples}
\scriptsize
\setlength{\tabcolsep}{2.5pt}
\renewcommand{\arraystretch}{1.72}
\begin{tabularx}{\textwidth}{@{}c l l l
  >{\centering\arraybackslash}X@{}}
\toprule
\textbf{Task} &
\textbf{Scenarios} &
\textbf{\textcolor{modeBlue}{Notification}} &
\textbf{\textcolor{modeGreen}{Intervention}} &
\textbf{Illustration} \\
\midrule
\multirow{3}{*}{\taskbadge{\vctask{}}} &
Shopping &
``This label contains dense nutrition facts.'' &
``Nutrition facts recorded.''  &
\multirow{3}{*}{\providecolor{taskShade}{HTML}{F6EAEC}%
\providecolor{modeBlue}{HTML}{2D75B6}%
\resizebox{\linewidth}{!}{%
\begin{tikzpicture}[
  badge/.style={rounded corners=2pt, fill=taskShade, inner xsep=5pt,
    inner ysep=2.5pt, font=\bfseries\small, text=teaserNavy},
  name/.style={font=\bfseries\footnotesize, color=teaserNavy,
    anchor=west, align=left, text width=2.15cm, inner sep=0pt},
  lbl/.style={font=\large\itshape, color=black!60},
  ink/.style={draw=teaserNavy, line width=0.7pt},
  soft/.style={rounded corners=3pt, draw=teaserLine, line width=0.8pt,
    fill=teaserSoft},
  flow/.style={-{Stealth[length=4.5pt]}, line width=0.9pt},
  sep/.style={dashed, draw=black!22, line width=0.6pt},
]
\path[use as bounding box] (-0.05,-0.12) rectangle (4.05,2.62);
\begin{scope}[shift={(-3.30,-0.10)}]
  \begin{scope}[shift={(3.35,0.18)}, rotate=-2.5]
    \draw[ink, rounded corners=2pt, fill=white] (0,0) rectangle (2.05,1.62);
    \foreach \y/\w in {1.42/1.74, 1.24/1.18, 1.06/1.66, 0.88/0.92,
                       0.70/1.70, 0.52/1.30, 0.34/1.60, 0.16/0.84} {
      \draw[draw=teaserNavy, line width=1pt, opacity=0.5] (0.16,\y) -- ({0.16+\w},\y);
    }
    \fill[teaserBlue, opacity=0.9] (0.16,1.18) rectangle (0.74,1.30);
    \fill[teaserBlue, opacity=0.9] (0.95,0.64) rectangle (1.55,0.76);
    \fill[teaserBlue, opacity=0.9] (0.16,0.28) rectangle (0.60,0.40);
  \end{scope}

  \draw[dotted, draw=black!30, line width=0.7pt] (3.45,2.34) -- (7.25,2.34);
  \draw[draw=teaserBlue, line width=5pt, opacity=0.30] (4.24,2.34) -- (4.52,2.34);
  \draw[draw=teaserBlue, line width=1.2pt] (4.24,2.34) -- (4.52,2.34);
  \draw[draw=teaserBlue, line width=1pt] (4.24,2.29) -- (4.24,2.39);
  \draw[draw=teaserBlue, line width=1pt] (4.52,2.29) -- (4.52,2.39);
  \node[lbl, anchor=south] at (4.38,2.40) {one glance};
  \fill[teaserBlue] (4.24,2.14) -- (4.52,2.14) -- (4.38,1.86) -- cycle;
  \draw[dotted, draw=teaserBlue!60, line width=0.7pt] (4.38,1.84) -- (4.38,1.60);

  \draw[flow, draw=teaserBlue] (5.55,1.00) -- (6.25,1.00);
  \node[lbl, anchor=south] at (5.90,1.30) {keep};

  \draw[soft] (6.35,0.55) rectangle (7.25,1.45);
  \foreach \y/\w in {1.28/0.64, 1.13/0.42, 0.98/0.58, 0.83/0.34, 0.68/0.50} {
    \draw[draw=teaserNavy, line width=0.9pt, opacity=0.5] (6.48,\y) -- ({6.48+\w},\y);
  }
\end{scope}
\end{tikzpicture}}} \\
& Construction &
``This sign lists work-zone instructions.'' &
``Work-zone instructions saved.'' & \\
& Car mechanic &
``This document lists repair specifications.'' &
``Repair specifications recorded.'' & \\
\midrule
\multirow{3}{*}{\taskbadge{\sttask{}}} &
Cooking &
``The faucet is still running.'' &
``Faucet tracked until off.''  &
\multirow{3}{*}{\providecolor{taskShade}{HTML}{F6EAEC}%
\providecolor{modeBlue}{HTML}{2D75B6}%
\resizebox{\linewidth}{!}{%
\begin{tikzpicture}[
  badge/.style={rounded corners=2pt, fill=taskShade, inner xsep=5pt,
    inner ysep=2.5pt, font=\bfseries\small, text=teaserNavy},
  name/.style={font=\bfseries\footnotesize, color=teaserNavy,
    anchor=west, align=left, text width=2.15cm, inner sep=0pt},
  lbl/.style={font=\large\itshape, color=black!60},
  ink/.style={draw=teaserNavy, line width=0.7pt},
  soft/.style={rounded corners=3pt, draw=teaserLine, line width=0.8pt,
    fill=teaserSoft},
  flow/.style={-{Stealth[length=4.5pt]}, line width=0.9pt},
  sep/.style={dashed, draw=black!22, line width=0.6pt},
]
\path[use as bounding box] (-0.05,-0.18) rectangle (4.45,2.05);
\begin{scope}[shift={(-3.30,0.02)}]
  \draw[ink, fill=white] (3.38,1.50) rectangle (3.62,1.74);
  \draw[ink, fill=white] (3.50,1.10) circle (0.13);
  \draw[ink, fill=white] (3.50,0.55) -- (3.63,0.68) -- (3.50,0.81) -- (3.37,0.68) -- cycle;

  \foreach \y/\a/\b/\c in {1.62/3.95/5.75/teaserTeal,
                           1.10/4.55/7.05/teaserBlue,
                           0.68/5.30/7.55/teaserGold} {
    \draw[sep] (3.88,\y) -- (7.70,\y);
    \draw[draw=\c, line width=4.5pt, opacity=0.28] (\a,\y) -- (\b,\y);
    \draw[draw=\c, line width=1pt] (\a,\y) -- (\b,\y);
    \fill[\c] ($(\a,\y)+(-0.11,0.21)$) -- ($(\a,\y)+(0.11,0.21)$) -- (\a,\y) -- cycle;
    \draw[draw=\c, line width=0.8pt, fill=white]
          ($(\b,\y)+(-0.11,-0.21)$) -- ($(\b,\y)+(0.11,-0.21)$) -- (\b,\y) -- cycle;
  }

  \draw[draw=teaserNavy, line width=0.7pt, dash pattern=on 2pt off 2pt]
        (5.55,0.24) -- (5.55,1.86);
  \node[lbl, anchor=south] at (5.55,1.82) {all on};

  \draw[flow, draw=black!55] (3.88,0.24) -- (7.70,0.24);
  \node[lbl, anchor=east] at (7.70,0.06) {time};
\end{scope}
\end{tikzpicture}}} \\
& Baking &
``The dough mixer is still running.'' &
``Mixer tracked until off.'' & \\
& Cleaning &
``The vacuum is on.'' &
``Vacuum tracked until off.'' & \\
\midrule
\multirow{3}{*}{\taskbadge{\lttask{}}} &
Carpentry &
``The hammer is taken from the workbench.'' &
``Hammer's last location: on the ground.''  &
\multirow{3}{*}{\providecolor{taskShade}{HTML}{F6EAEC}%
\providecolor{modeBlue}{HTML}{2D75B6}%
\resizebox{\linewidth}{!}{%
\begin{tikzpicture}[
  badge/.style={rounded corners=2pt, fill=taskShade, inner xsep=5pt,
    inner ysep=2.5pt, font=\bfseries\small, text=teaserNavy},
  name/.style={font=\bfseries\footnotesize, color=teaserNavy,
    anchor=west, align=left, text width=2.15cm, inner sep=0pt},
  lbl/.style={font=\large\itshape, color=black!60},
  ink/.style={draw=teaserNavy, line width=0.7pt},
  soft/.style={rounded corners=3pt, draw=teaserLine, line width=0.8pt,
    fill=teaserSoft},
  flow/.style={-{Stealth[length=4.5pt]}, line width=0.9pt},
  sep/.style={dashed, draw=black!22, line width=0.6pt},
]
\path[use as bounding box] (-0.05,-0.15) rectangle (4.55,2.40);
\begin{scope}[shift={(-3.20,0.02)}]
  \draw[soft, fill=taskGray] (3.25,0.22) rectangle (4.40,0.52);
  \node[lbl] at (3.82,0.04) {A};
  \draw[ink, fill=white] (3.35,0.52) rectangle (3.57,0.74);
  \draw[ink, fill=teaserGold!40] (3.85,0.67) circle (0.15);
  \draw[ink, fill=white] (4.16,0.55) -- (4.28,0.67) -- (4.16,0.79) -- (4.04,0.67) -- cycle;

  \draw[soft, fill=taskGray] (4.85,0.22) rectangle (6.00,0.52);
  \node[lbl] at (5.42,0.04) {A};
  \draw[ink, fill=white] (4.95,0.52) rectangle (5.17,0.74);
  \draw[ink, dash pattern=on 2pt off 2pt, fill=white] (5.45,0.67) circle (0.15);
  \draw[ink, fill=white] (5.76,0.55) -- (5.88,0.67) -- (5.76,0.79) -- (5.64,0.67) -- cycle;

  \draw[soft, fill=taskGray] (6.45,0.22) rectangle (7.60,0.52);
  \node[lbl] at (7.02,0.04) {B};
  \draw[ink, fill=teaserGold!40] (7.02,0.67) circle (0.15);

  \fill[modeBlue] (5.33,0.94) -- (5.57,0.94) -- (5.45,1.18) -- cycle;
  \fill[teaserGold] (6.90,1.26) -- (7.14,1.26) -- (7.02,1.02) -- cycle;
  \draw[draw=teaserGold, line width=0.9pt, dash pattern=on 3pt off 2.5pt]
        (5.62,1.04) .. controls (6.02,1.56) and (6.48,1.56) .. (6.80,1.12);

  \draw[dotted, draw=teaserNavy, line width=0.8pt] (3.60,2.06) -- (7.20,2.06);
  \draw[dotted, draw=teaserNavy!55, line width=0.7pt] (3.85,2.00) -- (3.85,0.90);
  \draw[dotted, draw=teaserNavy!55, line width=0.7pt] (5.45,2.00) -- (5.45,1.28);
  \draw[dotted, draw=teaserNavy!55, line width=0.7pt] (7.02,2.00) -- (7.02,1.38);
  \foreach \x in {3.85,5.45,6.30,7.02} { \fill[teaserNavy] (\x,2.06) circle (1.7pt); }
  \node[lbl, anchor=south] at (5.45,2.12) {watched throughout};
  
\end{scope}
\end{tikzpicture}}} \\
& Car mechanic &
``The hose was placed beside the car.'' &
``Hose's last location: on the grass.'' & \\
& Cleaning\slash laundry &
``The detergent bottle was in the cabinet.'' &
``Bottle's last location: on washing machine.'' & \\
\bottomrule
\end{tabularx}
\end{table}

\subsection{Three Ambient-Memory Tasks}
\label{sec:tasks}

From 4,000 hours of video, we define three core
ambient-memory tasks that offload mental burden by proactively preserving
transient information, object states, and location movements
(Table~\ref{tab:mode-examples}).

\noindent\tasktitle{\vctask{}: Information Capture.}\enspace
The agent actively preserves transient dense visual information in the ambient environment that may be useful later, such as a
document, sign, label, screen, poster, recipe, or instructions, etc. QA questions
ask about the proactively captured dense information, etc.

\noindent\tasktitle{\sttask{}: State Tracking.}\enspace
The agent tracks object states such as on\slash off, open\slash closed,
active\slash inactive, filled\slash empty, or connected\slash disconnected during multitasking. QA questions ask how often a state
changed, what remains active, the order of activations, or which object was
active longest, etc.

\noindent\tasktitle{\lttask{}: Location Tracking.}\enspace
The agent remembers where an object was placed elsewhere. The task requires tracking where an object is originally placed, when it is picked up, placed, or last observed if placed at a different location.
QA questions ask where an object was placed, its original location, the
order of locations, or total times it moved, etc.

\section{\method: Benchmark Construction}
\label{sec:construction}

\label{sec:dataset}
\begin{figure}[!t]
\centering
\includegraphics{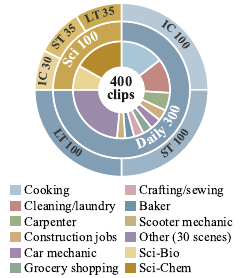}\hfill%
\includegraphics{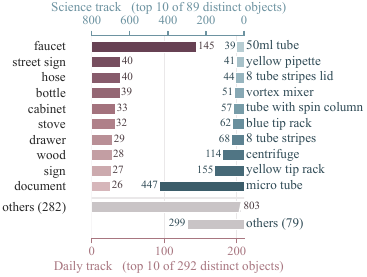}\hfill%
\includegraphics{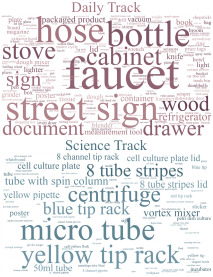}%
\caption{\label{fig:dataset-overview}
\method composition and target objects. \textbf{Left}: composition by scenario
(inner ring), track (middle), and task (outer). \textbf{Centre}: the ten most
frequent target objects per track, counted in annotated events.
\textbf{Right}: the full object vocabulary, sized by annotated events.}
\end{figure}

\textbf{Dataset source and annotation.}
\method is organized around clips in which an ambient memory would
help a user, and both tracks are built by the pipeline of
Figure~\ref{fig:construction-pipeline} (Appendix~\ref{sec:benchmark-comparison}): (1a)~native action labels are greedily paired into candidate events while (1b)~a VLM (Qwen3.6-27B) densely captions Daily clips; (2)~annotators correct these events and task families (Ego4D lab clips are annotated from scratch); (3)~QA is generated. The (4)~review by $\geq$3 annotators is visible in Table~\ref{fig:stats-agreement} and the composition and target objects in
Figure~\ref{fig:dataset-overview}. \dailymethod{}, the Daily track, is built
from Ego4D \citep{9879279}. \scimethod{}, the Science track, is built
from egocentric recordings in wet-labs
\citep{yagi2025finebio,nishimoto2024biovlqr} and from lab videos in Ego4D \citep{9879279} for a co-scientist setting. All annotations on raw video streams and evaluation QA pairs are newly annotated, ensuring zero label leakage from pre-training corpora (details in Appendix~\ref{sec:data-leakage-mitigation}).

\textbf{ProactEgoMem-QA.}\label{sec:qa-construction}
Each annotated event is paired with multiple-choice questions that test
whether the acquired memory supports later recall;
Table~\ref{tab:qa-template-examples-dense} lists the question categories of
each task with a benchmark example. (1)~\vctask{} questions are generated by an LLM
from dense human annotations of visible information and reviewed in three
rounds, and their categories are the forms these questions recur in.
(2)~\sttask{} questions are chiefly instantiated from templates over the reviewed
state events, filling object, count, time, duration, and order slots.
(3)~\lttask{} questions are chiefly instantiated from templates over the reviewed
object-location transitions, destination, placement, and count
slots. Every question is reviewed by at least three annotators above bachelor's degree level.

\begin{table}[!t]
\centering
\caption{\label{tab:qa-template-examples-dense}\textbf{ProactEgoMem-QA question types and benchmark examples}}
\tiny
\setlength{\tabcolsep}{2.2pt}
\renewcommand{\arraystretch}{0.95}
\setlength{\aboverulesep}{0.45ex}\setlength{\belowrulesep}{0.55ex}
\renewcommand{\correctoption}[1]{\begingroup\setlength{\fboxsep}{0.9pt}%
  \colorbox{qaAnswerBg}{\textcolor{qaAnswerText}{\textbf{#1}}}\endgroup}%
\renewcommand{\denseqa}[1]{\sbox{\denseqaBox}{#1}%
  \ifdim\wd\denseqaBox>\linewidth\resizebox{\linewidth}{!}{\usebox{\denseqaBox}}%
  \else\usebox{\denseqaBox}\fi}%
\begin{tabularx}{\textwidth}{@{}
  >{\centering\arraybackslash}p{0.035\textwidth}
  >{\raggedright\arraybackslash}p{0.105\textwidth}
  >{\raggedright\arraybackslash}X@{}}
\toprule
\textbf{Task} & \textbf{Category}
& \textbf{Question and candidates} \\
\midrule
\multirow{6}{*}{\taskbadge{\vctask{}}}
& Text
& \denseqa{\emph{What were the Rental Center hours on Sunday?}\enspace
  \textcolor{black!70}{a.\ 10a--4p \enspace b.\ 1p--6p \enspace
  \correctoption{c.\ 12p--5p} \enspace d.\ 10a--6p}} \\
\cmidrule(l){2-3}
& Sequence
& \denseqa{\emph{What was the first packaged product likely to be?}\enspace
  \textcolor{black!70}{a.\ Bacon strips \enspace b.\ Olive oil \enspace
  c.\ Ham \enspace \correctoption{d.\ Mushrooms}}} \\
\cmidrule(l){2-3}
& Attribute
& \denseqa{\emph{What kind of oil did I pour into the frying pan?}\enspace
  \textcolor{black!70}{a.\ Sesame oil \enspace b.\ Canola oil \enspace
  \correctoption{c.\ Olive oil} \enspace d.\ Sunflower oil}} \\
\cmidrule(l){2-3}
& Price
& \denseqa{\emph{How much was the Pop Mandalorian item?}\enspace
  \textcolor{black!70}{a.\ \$20 \enspace b.\ \$11 \enspace c.\ \$9.50
  \enspace \correctoption{d.\ \$10}}} \\
\cmidrule(l){2-3}
& Legibility
& \denseqa{\emph{How readable were the Dominoes rules?}\enspace
  \textcolor{black!70}{\correctoption{a.\ Barely readable} \enspace
  b.\ Written in large print \enspace c.\ Completely unreadable \enspace
  d.\ Clearly readable}} \\
\cmidrule(l){2-3}
& Direction
& \denseqa{\emph{Which way did the sign point for Outdoor Center Rentals?}\enspace
  \textcolor{black!70}{\correctoption{a.\ To the left} \enspace
  b.\ Back behind you \enspace c.\ To the right \enspace d.\ Straight ahead}} \\
\midrule
\multirow{6}{*}{\taskbadge{\sttask{}}}
& Frequency
& \denseqa{\emph{How often did I open the drawer?}\enspace
  \textcolor{black!70}{a.\ 0 times \enspace b.\ 1 time \enspace
  c.\ 3 times \enspace \correctoption{d.\ 2 times}}} \\
\cmidrule(l){2-3}
& Final state
& \denseqa{\emph{How many objects remained on or open at the clip's end?}\enspace
  \textcolor{black!70}{a.\ 0 objects \enspace b.\ 3 objects \enspace
  c.\ 1 object \enspace \correctoption{d.\ 2 objects}}} \\
\cmidrule(l){2-3}
& Unattended state
& \denseqa{\emph{How many objects did I open and leave unattended while open?}\enspace
  \textcolor{black!70}{\correctoption{a.\ 1 object} \enspace
  b.\ 3 objects \enspace c.\ 0 objects \enspace d.\ 2 objects}} \\
\cmidrule(l){2-3}
& Duration
& \denseqa{\emph{How long was the light turned on?}\enspace
  \textcolor{black!70}{\correctoption{a.\ 10 seconds} \enspace
  b.\ 25 seconds \enspace c.\ 100 seconds \enspace d.\ 30 seconds}} \\
\cmidrule(l){2-3}
& Event order
& \denseqa{\emph{What was the on-order (L=light, F=faucet, C=cabinet)?}\enspace
  \textcolor{black!70}{a.\ L$\rightarrow$F$\rightarrow$F$\rightarrow$C
  \enspace b.\ F$\rightarrow$F$\rightarrow$L$\rightarrow$C
  \enspace c.\ F$\rightarrow$C$\rightarrow$F$\rightarrow$L
  \enspace \correctoption{d.\ C$\rightarrow$F$\rightarrow$F$\rightarrow$L}}} \\
\cmidrule(l){2-3}
& Before/after
& \denseqa{\emph{What did I open after I turned on the stove?}\enspace
  \textcolor{black!70}{a.\ oven door \enspace b.\ cabinet \enspace
  \correctoption{c.\ refrigerator} \enspace d.\ drawer}} \\
\midrule
\multirow{6}{*}{\taskbadge{\lttask{}}}
& Placement
& \denseqa{\emph{Where did I place small metal cogs?}\enspace
  \textcolor{black!70}{\correctoption{a.\ on the workbench} \enspace
  b.\ in the case \enspace c.\ in the drawer \enspace d.\ in the box}} \\
\cmidrule(l){2-3}
& Source
& \denseqa{\emph{Where did I pick up the white bowl before putting it in the sink?}\enspace
  \textcolor{black!70}{a.\ from the drawer \enspace
  \correctoption{b.\ from the table} \enspace c.\ from the cabinet \enspace
  d.\ from the sink}} \\
\cmidrule(l){2-3}
& Destination
& \denseqa{\emph{Where did I place the black crankset after taking it from the box?}\enspace
  \textcolor{black!70}{a.\ mounted on the bicycle \enspace b.\ in the case
  \enspace \correctoption{c.\ on the workbench} \enspace d.\ in the box}} \\
\cmidrule(l){2-3}
& Last location
& \denseqa{\emph{Where did I place the black pliers the last time?}\enspace
  \textcolor{black!70}{a.\ in the box \enspace b.\ in the drawer \enspace
  \correctoption{c.\ on the wall} \enspace d.\ in the case}} \\
\cmidrule(l){2-3}
& Location count
& \denseqa{\emph{How many times did I place the knife at a different location?}\enspace
  \textcolor{black!70}{\correctoption{a.\ 1 time} \enspace b.\ 0 times
  \enspace c.\ 3 times \enspace d.\ 2 times}} \\
\cmidrule(l){2-3}
& First location
& \denseqa{\emph{What was the hex key's original location?}\enspace
  \textcolor{black!70}{a.\ hanging on the wall \enspace
  \correctoption{b.\ on the workbench} \enspace
  c.\ on the tool tray by the machine \enspace d.\ in the drawer}} \\
\bottomrule
\end{tabularx}
\end{table}

\begin{table}[!t]
\centering
\caption{\label{fig:stats-agreement}\label{fig:agreement}
\method data statistics and inter-annotator agreement. Paired entries read
Daily/Sci. $N$: reference-accepted GT events; agreement is averaged over three
annotators.}
\setlength{\tabcolsep}{4pt}%
\ifdefined\statspad\else\newlength{\statspad}\newsavebox{\statsR}\newsavebox{\statsL}\fi
\setlength{\statspad}{0pt}%
\newcommand{\statsleft}{%
\begin{tabular}[t]{@{}lr@{\hskip 6pt}lr@{}}
\toprule
\textbf{Statistic} & \textbf{Value} & \textbf{Statistic} & \textbf{Value} \\[\statspad]
\midrule
Unique clips         & \dailyclips{}/\sciclips{}         & \quad closed-ended    & \dailyclosed{}/\sciclosed{} \\[\statspad]
Source videos & \dailyvideos{}/\scivideos{}       & \quad open-ended      & \dailyopen{}/\sciopen{} \\[\statspad]
Total clip len.   & \dailyhours{}/\scihours{} h       & Mean event len.  & \dailyavgseg{}/\sciavgseg{} s \\[\statspad]
Mean clip len.    & \dailymeanclip{}/\scimeanclip{} s & Median event len. & \dailymedseg{}/\scimedseg{} s \\[\statspad]
Annotated events   & \dailysegments{}/\scisegments{}   & Distinct objects      & \dailyobjects{}/\sciobjects{} \\
\bottomrule
\end{tabular}}%
\sbox{\statsR}{\begin{tabular}[t]{@{}lrrr@{\hskip 6pt}rrrrrrr@{}}
\toprule
& \multicolumn{3}{c}{\textbf{Task Breakdown}}
& \multicolumn{7}{c}{\textbf{Inter-Annotator Agreement}} \\
\cmidrule(r{4pt}){2-4}\cmidrule(l){5-11}
\textbf{Task} & \textbf{Clips} & \textbf{Events} & \textbf{QA}
& $N$ & mIoU$\uparrow$ & $\sigma_{\mathrm{IoU}}{\downarrow}$
& S@2$\uparrow$ & E@2$\uparrow$ & $|\Delta_s|{\downarrow}$ & $|\Delta_e|{\downarrow}$ \\
\midrule
\vctask{} & \dailyclipspertask{}/\sciicclips{} & \dailysegic{}/\scisegic{} & \dailyqaic{}/\sciqaic{}
& 323 & 0.9403 & 0.1880 & 0.9761 & 0.9283 & 0.2012 & 2.0305 \\
\sttask{} & \dailyclipspertask{}/\scistclips{} & \dailysegst{}/\scisegst{} & \dailyqast{}/\sciqast{}
& 465 & 0.9957 & 0.0614 & 0.9957 & 0.9957 & 0.1497 & 0.2380 \\
\lttask{} & \dailyclipspertask{}/\sciltclips{} & \dailyseglt{}/\sciseglt{} & \dailyqalt{}/\sciqalt{}
& 547 & 0.9940 & 0.0589 & 0.9843 & 0.9935 & 0.3530 & 0.2382 \\
\midrule
Overall & \dailyclips{}/\sciclips{} & \dailysegments{}/\scisegments{} & \dailyqa{}/\sciqa{}
& 1335 & 0.9817 & 0.1084 & 0.9865 & 0.9786 & 0.2429 & 0.6675 \\
\bottomrule
\end{tabular}}%
\sbox{\statsL}{\statsleft}%
\setlength{\statspad}{\dimexpr(\ht\statsR+\dp\statsR-\ht\statsL-\dp\statsL)/5\relax}%
\resizebox{\textwidth}{!}{\statsleft\hspace{14pt}\usebox{\statsR}}
\end{table}

\section{Experiments}
\label{sec:experiments}

\subsection{Metrics and Baselines}
\label{sec:metrics}
\label{sec:protocol}

\textbf{Common metrics.}
Predictions are matched and scored separately at the two proactive levels.
Predictions at both levels are matched to ground truth by Hungarian
matching~\citep{kuhn1955hungarian}, which requires that an LLM-judge's semantic
similarity exceeds a threshold and that the temporal offset falls within the
matching window; among admitted edges the assignment minimizes the cost
$\alpha(1-\mathrm{sim})+(1-\alpha)\,(|\Delta|/W)^{p}$~\citep{zhang2025proassist}. Intervention
predictions are first converted into intervals and matched on temporal
overlap. Three metric groups are then shared by the two levels:
(1)~\textbf{Timing} (Start@$\tau$): the fraction of matched actions whose
start falls within various $\tau$ seconds of the ground-truth moment. We
report Start@0.2 as the main result: the 200\,ms tolerance reflects the
residual inter-annotator disagreement on notification timing
(Table~\ref{fig:stats-agreement}) and is widely used in real-time streaming
models.
(2)~\textbf{Abs Error} ($|\Delta_s|$, $\sigma$): the mean absolute start
offset and its standard deviation, in seconds, measuring how far and how
consistently a model deviates from the annotated moment.
(3)~\textbf{Detection} (Prec, Recall, F1): precision over emitted actions,
recall over ground-truth events, and their harmonic mean F1,
penalizing spurious interruptions and missed ones.

\textbf{\IntervLevelAdj specific metrics.}
The Intervention-level metrics score the lifecycle itself through two further groups:
(4)~\textbf{Overlap} (mIoU): interval IoU averaged over \emph{all}
ground-truth opportunities, scoring unmatched opportunities as zero, so a high
score requires starting, maintaining, and stopping the tool lifecycle at the
right boundaries; various thresholds in IoU@$\tau$ are also reported.
(5)~\textbf{QA} (Acc): accuracy of the ProactEgoMem-QA dataset requiring
proactive memory.

\textbf{Baselines.}\label{sec:baselines}
The comparison includes proprietary models (Gemini 2.5 Flash \citep{comanici2025gemini25pushingfrontier} and
GPT-5.4-mini \citep{openai2026gpt54mini}), general open-source models (Gemma4-31B \citep{gemma4_2026},
Qwen3.8-27B \citep{qwen38}, Qwen3.6-27B \citep{qwen36_2026},
Qwen3.5-9B \citep{qwen35_2026}), video-language models (VideoLLaMA3-7B \citep{videollama3_2025}, LLaVA-OV-1.5-8B \citep{llavaonevision_15_2025},
MiniCPM-o-4.5 \citep{minicpmo45_2026}), and streaming models (MMDuet \citep{wang2025mmduet}, VideoLLM-Online \citep{chen2024videollmonline}, Inkling-Small \citep{thinkingmachines2026inklingsmall}, and JoyAI-VL \citep{yao2026joyaivlinteractionrealtimevisionlanguageinteraction}).
All baselines are run with the exact same streaming setting as follows.

\subsection{Streaming Evaluation Protocol}
\label{app:streaming-protocol}

Every baseline receives the same prompt for a given task and proactive level; model-specific code
only wraps it in the model's chat template. FPS rate of 1 is used throughout the evaluation. At every step the prompt is followed by a JSON
assistant context and the window of video frames, and the model answers with one JSON object. Complete description in Appendix~\ref{app:prompts}. \textbf{Per turn:} one request per video second simulating the streaming setting, with the task prompt,
a JSON context, and the 10 most recent
frames up to current timestamp $t$, sampled at 1 fps,
each with frame number and timestamp. \textbf{Earlier turns:} only through
the context, listing active tool instances and the $5$ latest
completed intervals (Intervention) or notifications (Notification). \textbf{Output:} one JSON object determining the tool calls at timestamp $t$; unparseable output counts as action of silence.

\subsection{Main Results}
\label{sec:main-results}

\definecolor{vintageAccent}{HTML}{9C6B72}
\definecolor{vintageHL}{HTML}{EADCDE}
\definecolor{rcBarBlue}{HTML}{BCD7EC}
\providecommand{\rcbarfmt}[2]{%
\begin{tikzpicture}[
  baseline=(num.base),
  inner sep=0pt,
  outer sep=0pt,
  x=1cm,
  y=1cm
]
  \def\W{1.15}
  \def\H{0.30}
  \pgfmathsetmacro{\barend}{min(max(#2,0),100)/100*\W}
  \fill[rcBarBlue]
    (0,-0.5*\H) rectangle (\barend,0.5*\H);
  \node[font=\small,inner sep=0pt] (num) at ({0.5*\W},0) {#1{#2}};
  \path[use as bounding box]
    (0,-0.5*\H) rectangle (\W,0.5*\H);
\end{tikzpicture}%
}
\providecommand{\rcident}[1]{#1}
\providecommand{\rcbful}[1]{\textbf{\underline{#1}}}
\providecommand{\rcbar}[1]{\rcbarfmt{\rcident}{#1}}
\providecommand{\rcbarb}[1]{\rcbarfmt{\textbf}{#1}}
\providecommand{\rcbaru}[1]{\rcbarfmt{\underline}{#1}}
\providecommand{\rcbarbu}[1]{\rcbarfmt{\rcbful}{#1}}
\providecommand{\grouprule}{%
  \arrayrulecolor{black!30}\cmidrule[0.02em](lr){2-17}\arrayrulecolor{black}}

\begin{table}[!t]
\centering
\caption{\label{tab:merged-main-results}%
\textbf{\dailymethod{} main results} for notification and intervention levels.
\textbullet~\textbf{Start@0.2}: matched actions within $0.2$\,s of the ground-truth start.
\textbullet~\textbf{mIoU}: interval IoU over \emph{all} opportunities (unmatched scored~$0$).
\textbullet~Error columns are seconds, all others percentages; \textbf{bold}: best in a task.}
\vspace{2pt}
\setlength{\tabcolsep}{4pt}
\resizebox{\textwidth}{!}{%
\small
\begin{tabular}{@{}lllrrrrrr!{\vrule width 1.1pt}rrrrrrrr@{}}
\toprule
\multirow{3}{*}{Task} & \multirow{3}{*}{Group} & \multirow{3}{*}{Model}
& \multicolumn{6}{c!{\vrule width 1.1pt}}{\textbf{\NotifLevel}}
& \multicolumn{8}{c}{\textbf{\IntervLevel}} \\
\cmidrule(lr){4-9}\cmidrule(l){10-17}
& & & \multicolumn{1}{c}{Timing} & \multicolumn{2}{c}{Abs Error} & \multicolumn{3}{c!{\vrule width 1.1pt}}{Detection}
& \multicolumn{1}{c}{Overlap} & \multicolumn{1}{c}{Timing} & \multicolumn{2}{c}{Abs Error} & \multicolumn{3}{c}{Detection} & \multicolumn{1}{c}{QA} \\
\cmidrule(lr){4-4}\cmidrule(lr){5-6}\cmidrule(lr){7-9}\cmidrule(lr){10-10}\cmidrule(lr){11-11}\cmidrule(lr){12-13}\cmidrule(lr){14-16}\cmidrule(l){17-17}
& & & S@0.2$\uparrow$ & $|\Delta_s|{\downarrow}$ & $\sigma{\downarrow}$ & Prec$\uparrow$ & Recall$\uparrow$ & F1$\uparrow$
& mIoU$\uparrow$ & S@0.2$\uparrow$ & $|\Delta_s|{\downarrow}$ & $\sigma{\downarrow}$ & Prec$\uparrow$ & Recall$\uparrow$ & F1$\uparrow$ & Acc$\uparrow$ \\
\midrule
\multirow{13}{*}{IC}
& \multirow{2}{*}{Proprietary} & Gemini 2.5 Flash & \rcbar{14.2} & 2.09 & 2.04 & \rcbar{3.4} & \rcbar{66.2} & 6.5 & \rcbar{3.4} & \rcbar{16.7} & 3.34 & 2.88 & \rcbar{3.9} & \rcbar{10.8} & 5.8 & \rcbar{54.1} \\
& & GPT-5.4-mini & \rcbar{17.0} & 1.70 & 2.09 & \rcbar{1.3} & \rcbar{65.6} & 2.6 & \rcbar{2.0} & \rcbar{5.3} & 3.83 & 2.22 & \rcbar{5.1} & \rcbar{6.9} & 5.9 & \rcbar{52.5} \\
\grouprule
& \multirow{4}{*}{Open Source} & Gemma4-31B & \rcbar{11.8} & 2.16 & 2.25 & \rcbar{6.1} & \rcbar{62.8} & 11.1 & \rcbar{4.5} & \rcbar{8.6} & 3.01 & 2.71 & \rcbar{3.0} & \rcbarb{37.9} & 5.5 & \rcbarb{66.5} \\
& & Qwen3.8-27B & \rcbar{14.9} & 1.70 & 1.85 & \rcbar{3.2} & \rcbar{74.8} & 6.1 & \rcbarb{5.6} & \rcbarb{20.0} & 2.50 & 2.64 & \rcbar{5.9} & \rcbar{10.8} & \textbf{7.6} & \rcbar{62.7} \\
& & Qwen3.6-27B & \rcbar{13.7} & 1.74 & 1.89 & \rcbar{6.1} & \rcbar{70.3} & 11.2 & \rcbar{4.3} & \rcbar{5.9} & 3.04 & 2.42 & \rcbar{2.9} & \rcbar{30.7} & 5.3 & \rcbar{60.4} \\
& & Qwen3.5-9B & \rcbar{19.1} & 1.99 & 2.24 & \rcbar{1.9} & \rcbar{58.2} & 3.7 & \rcbar{0.7} & \rcbar{11.1} & 2.76 & \textbf{1.71} & \rcbar{5.8} & \rcbar{3.2} & 4.2 & \rcbar{54.0} \\
\grouprule
& \multirow{3}{*}{VidLLM} & VideoLLaMA3-7B & \rcbarb{28.1} & 2.18 & 3.11 & \rcbar{0.5} & \rcbar{35.3} & 1.1 & \rcbar{0.6} & \rcbar{9.1} & 3.83 & 2.55 & \rcbar{0.7} & \rcbar{4.0} & 1.2 & \rcbar{50.5} \\
& & LLaVA-OV-1.5-8B & \rcbar{0.0} & 8.50 & 0.00 & \rcbar{16.7} & \rcbar{0.3} & 0.6 & \rcbar{0.3} & \rcbar{0.0} & 7.09 & 1.94 & \rcbar{8.8} & \rcbar{2.2} & 3.5 & \rcbar{41.0} \\
& & MiniCPM-o-4.5 & \rcbar{10.3} & 4.23 & 2.84 & \rcbarb{29.6} & \rcbar{9.0} & \textbf{13.8} & \rcbar{0.0} & \rcbar{0.0} & 3.12 & 0.00 & \rcbar{20.0} & \rcbar{0.4} & 0.7 & \rcbar{36.1} \\
\grouprule
& \multirow{4}{*}{Streaming} & MMDuet & \rcbar{10.5} & 2.62 & 2.47 & \rcbar{1.7} & \rcbar{11.8} & 2.9 & \rcbar{0.2} & \rcbarb{20.0} & \textbf{2.40} & 3.32 & \rcbar{4.9} & \rcbar{1.8} & 2.6 & \rcbar{40.2} \\
& & VideoLLM-Online & \rcbar{28.0} & \textbf{1.59} & \textbf{1.77} & \rcbar{1.3} & \rcbar{7.7} & 2.2 & \rcbar{0.0} & \rcbar{0.0} & 0.00 & 0.00 & \rcbar{0.0} & \rcbar{0.0} & 0.0 & \rcbar{0.0} \\
& & Inkling-Small & \rcbar{16.7} & 1.72 & 1.90 & \rcbar{1.9} & \rcbarb{79.5} & 3.7 & \rcbar{1.7} & \rcbar{6.8} & 4.29 & 2.79 & \rcbar{1.6} & \rcbar{15.9} & 2.9 & \rcbar{49.5} \\
& & JoyAI-VL & \rcbar{5.3} & 2.06 & 2.01 & \rcbar{3.1} & \rcbar{27.3} & 5.6 & \rcbar{0.3} & \rcbar{0.0} & 0.00 & 0.00 & \rcbarb{22.2} & \rcbar{0.7} & 1.4 & \rcbar{52.3} \\
\midrule[\heavyrulewidth]
\multirow{13}{*}{ST}
& \multirow{2}{*}{Proprietary} & Gemini 2.5 Flash & \rcbar{14.1} & 2.07 & 2.33 & \rcbar{6.4} & \rcbar{35.1} & 10.9 & \rcbar{5.1} & \rcbar{14.1} & 2.15 & \textbf{2.06} & \rcbar{8.8} & \rcbar{14.7} & 11.0 & \rcbar{28.8} \\
& & GPT-5.4-mini & \rcbar{16.9} & 1.57 & 1.77 & \rcbar{3.2} & \rcbar{40.9} & 5.9 & \rcbar{0.8} & \rcbar{16.0} & 3.46 & 3.26 & \rcbar{9.5} & \rcbar{5.2} & 6.7 & \rcbar{21.5} \\
\grouprule
& \multirow{4}{*}{Open Source} & Gemma4-31B & \rcbar{7.0} & 3.13 & 2.55 & \rcbar{5.8} & \rcbar{27.7} & 9.6 & \rcbarb{17.2} & \rcbarb{17.2} & \textbf{1.76} & 2.23 & \rcbar{10.1} & \rcbarb{33.7} & 15.5 & \rcbarb{34.2} \\
& & Qwen3.8-27B & \rcbar{23.4} & 1.98 & 2.53 & \rcbar{4.1} & \rcbar{40.8} & 7.4 & \rcbar{13.8} & \rcbar{11.0} & 2.02 & 2.23 & \rcbarb{14.7} & \rcbar{31.9} & \textbf{20.1} & \rcbar{32.4} \\
& & Qwen3.6-27B & \rcbar{15.4} & 1.82 & 2.14 & \rcbarb{14.8} & \rcbar{26.6} & \textbf{19.0} & \rcbar{9.0} & \rcbar{11.9} & 2.33 & 2.53 & \rcbar{10.5} & \rcbar{26.1} & 15.0 & \rcbar{28.3} \\
& & Qwen3.5-9B & \rcbarb{27.2} & 1.10 & 1.68 & \rcbar{1.6} & \rcbarb{46.1} & 3.0 & \rcbar{0.0} & \rcbar{0.0} & 0.00 & 0.00 & \rcbar{0.0} & \rcbar{0.0} & 0.0 & \rcbar{17.7} \\
\grouprule
& \multirow{3}{*}{VidLLM} & VideoLLaMA3-7B & \rcbar{24.8} & 1.34 & 1.87 & \rcbar{0.8} & \rcbar{30.5} & 1.5 & \rcbar{0.4} & \rcbar{16.7} & 3.27 & 2.99 & \rcbar{0.5} & \rcbar{2.5} & 0.9 & \rcbar{33.3} \\
& & LLaVA-OV-1.5-8B & \rcbar{0.0} & 0.00 & 0.00 & \rcbar{0.0} & \rcbar{0.0} & 0.0 & \rcbar{0.0} & \rcbar{0.0} & 0.00 & 0.00 & \rcbar{0.0} & \rcbar{0.0} & 0.0 & \rcbar{21.9} \\
& & MiniCPM-o-4.5 & \rcbar{11.1} & 2.60 & 2.65 & \rcbar{1.6} & \rcbar{1.9} & 1.8 & \rcbar{0.0} & \rcbar{0.0} & 0.00 & 0.00 & \rcbar{0.0} & \rcbar{0.0} & 0.0 & \rcbar{19.4} \\
\grouprule
& \multirow{4}{*}{Streaming} & MMDuet & \rcbar{4.0} & 2.04 & 1.74 & \rcbar{1.6} & \rcbar{5.4} & 2.5 & \rcbar{0.1} & \rcbar{0.0} & 2.00 & 0.00 & \rcbar{3.2} & \rcbar{0.2} & 0.4 & \rcbar{29.1} \\
& & VideoLLM-Online & \rcbar{25.4} & \textbf{0.83} & \textbf{1.51} & \rcbar{2.8} & \rcbar{12.8} & 4.6 & \rcbar{0.0} & \rcbar{0.0} & 0.00 & 0.00 & \rcbar{0.0} & \rcbar{0.0} & 0.0 & \rcbar{0.0} \\
& & Inkling-Small & \rcbar{18.7} & 1.67 & 2.17 & \rcbar{2.1} & \rcbar{44.3} & 4.1 & \rcbar{4.3} & \rcbar{6.6} & 2.99 & 2.52 & \rcbar{3.1} & \rcbar{31.3} & 5.6 & \rcbar{21.9} \\
& & JoyAI-VL & \rcbar{12.5} & 1.40 & 1.73 & \rcbar{4.8} & \rcbar{19.9} & 7.7 & \rcbar{0.0} & \rcbar{0.0} & 0.00 & 0.00 & \rcbar{0.0} & \rcbar{0.0} & 0.0 & \rcbar{18.6} \\
\midrule[\heavyrulewidth]
\multirow{13}{*}{LT}
& \multirow{2}{*}{Proprietary} & Gemini 2.5 Flash & \rcbar{10.6} & 1.97 & 2.29 & \rcbar{4.3} & \rcbar{34.0} & 7.6 & \rcbar{0.6} & \rcbar{2.5} & 3.88 & 2.80 & \rcbar{9.1} & \rcbar{8.3} & 8.7 & \rcbar{18.3} \\
& & GPT-5.4-mini & \rcbar{16.9} & 1.61 & 2.21 & \rcbar{2.7} & \rcbar{55.9} & 5.1 & \rcbar{0.1} & \rcbarb{14.3} & 1.77 & 1.79 & \rcbar{9.7} & \rcbar{1.5} & 2.5 & \rcbar{24.6} \\
\grouprule
& \multirow{4}{*}{Open Source} & Gemma4-31B & \rcbar{9.2} & 1.77 & 1.71 & \rcbarb{13.2} & \rcbar{32.3} & \textbf{18.7} & \rcbar{4.4} & \rcbar{9.3} & 2.73 & 2.77 & \rcbar{5.2} & \rcbar{33.5} & 9.0 & \rcbar{29.0} \\
& & Qwen3.8-27B & \rcbar{16.4} & 1.38 & 1.89 & \rcbar{4.6} & \rcbarb{62.2} & 8.5 & \rcbarb{5.2} & \rcbar{5.4} & 3.08 & 2.66 & \rcbar{8.9} & \rcbarb{35.0} & \textbf{14.2} & \rcbar{30.6} \\
& & Qwen3.6-27B & \rcbar{13.8} & 1.53 & 1.66 & \rcbar{9.9} & \rcbar{33.6} & 15.3 & \rcbar{2.1} & \rcbar{3.2} & 1.82 & 2.05 & \rcbar{12.0} & \rcbar{13.1} & 12.5 & \rcbar{24.2} \\
& & Qwen3.5-9B & \rcbar{24.9} & 0.91 & 1.51 & \rcbar{1.3} & \rcbar{35.9} & 2.5 & \rcbar{0.6} & \rcbar{0.0} & 4.41 & 2.22 & \rcbar{11.5} & \rcbar{0.6} & 1.2 & \rcbar{21.8} \\
\grouprule
& \multirow{3}{*}{VidLLM} & VideoLLaMA3-7B & \rcbarb{28.8} & \textbf{0.73} & 1.32 & \rcbar{0.5} & \rcbar{20.6} & 0.9 & \rcbar{2.2} & \rcbar{4.2} & 3.93 & 2.65 & \rcbar{3.6} & \rcbar{19.8} & 6.1 & \rcbarb{36.1} \\
& & LLaVA-OV-1.5-8B & \rcbar{0.0} & 0.00 & 0.00 & \rcbar{0.0} & \rcbar{0.0} & 0.0 & \rcbar{0.5} & \rcbar{0.0} & 2.66 & 1.26 & \rcbar{7.0} & \rcbar{0.6} & 1.1 & \rcbar{23.8} \\
& & MiniCPM-o-4.5 & \rcbar{20.0} & 1.25 & \textbf{1.14} & \rcbar{2.8} & \rcbar{1.9} & 2.2 & \rcbar{0.2} & \rcbar{0.0} & \textbf{0.69} & \textbf{0.10} & \rcbarb{20.0} & \rcbar{0.4} & 0.8 & \rcbar{18.3} \\
\grouprule
& \multirow{4}{*}{Streaming} & MMDuet & \rcbar{4.7} & 2.96 & 2.30 & \rcbar{4.4} & \rcbar{16.0} & 6.9 & \rcbar{0.0} & \rcbar{0.0} & 0.00 & 0.00 & \rcbar{0.0} & \rcbar{0.0} & 0.0 & \rcbar{29.0} \\
& & VideoLLM-Online & \rcbar{7.3} & 1.71 & 2.16 & \rcbar{0.9} & \rcbar{10.2} & 1.7 & \rcbar{0.0} & \rcbar{0.0} & 0.00 & 0.00 & \rcbar{0.0} & \rcbar{0.0} & 0.0 & \rcbar{0.0} \\
& & Inkling-Small & \rcbar{9.2} & 2.29 & 2.13 & \rcbar{1.7} & \rcbar{31.8} & 3.2 & \rcbar{1.2} & \rcbar{5.7} & 3.38 & 2.76 & \rcbar{1.0} & \rcbar{21.9} & 2.0 & \rcbar{29.0} \\
& & JoyAI-VL & \rcbar{12.7} & 2.61 & 2.84 & \rcbar{2.3} & \rcbar{11.4} & 3.8 & \rcbar{0.0} & \rcbar{0.0} & 0.00 & 0.00 & \rcbar{0.0} & \rcbar{0.0} & 0.0 & \rcbar{30.6} \\
\bottomrule
\end{tabular}%
}
\providecommand{\mainresultstakeaway}{}%
\vspace{1pt}
\mainresultstakeaway\vspace{-0.8em}
\end{table}

\begin{figure}[!t]
\centering
\includegraphics{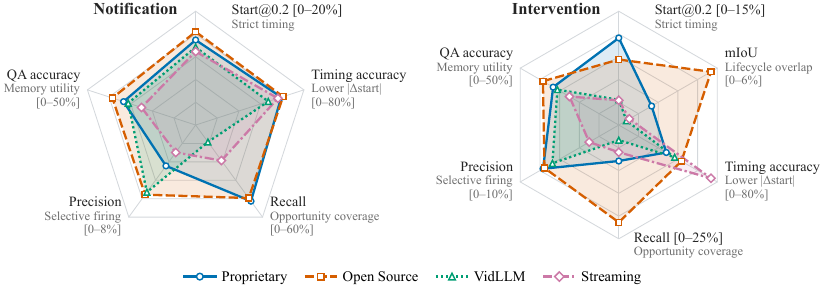}
\caption{\label{fig:radars}
\textbf{Proactive competence by model family}, averaged over the three tasks:
proprietary, open-source, video-language, and streaming models on the
notification and intervention axes.}
\end{figure}

Table~\ref{tab:merged-main-results} reports the \dailymethod{} main results for
every model at both proactive levels, and Figure~\ref{fig:radars} summarises
them by model family. The appendix gives per-level detail in
Appendix~\ref{sec:proactive-level-results}, per-task detail in
Appendix~\ref{sec:task-results}, message quality in
Appendix~\ref{sec:proactive-message}, the timing tolerance sweeps and
start-error distributions in Appendix~\ref{sec:timing-thresholds}, and full
\scimethod{} results in Appendix~\ref{app:sci-main-results}.

\noindent\textbf{No model
dominates and covers all aspects of proactive memory acquisition. The
\IntervLevel{} is uniformly harder than the \NotifLevel{}.} Across all thirteen baselines, covering opportunities never
coincides with acting precisely: the best \NotifLevelPlain{} F1 reaches only
$19.0\%$ (Qwen3.6-27B on \sttask{}), the best \IntervLevelPlain{} overlap only
$17.2\%$ mIoU (Gemma4-31B on \sttask{}), and the strongest strict-timing score
only $28.8\%$ Start@0.2 (VideoLLaMA3-7B on \lttask{}). \IntervLevelPlain{} is spiking on one or
two axes while the rest collapse toward the low scores. 

\noindent\textbf{General VLMs perform better regarding metric coverage, while streaming models excel in timing accuracy.}
(1) Open-source models generally cover more than other models at both level, especially significantly more at the Intervention level. (2) Prompted video-language models are the worst on Recall. (3) VLMs, including open-source and proprietary ones, perform better than native-trained video models regarding metric coverage. (4) The streaming models spike at the timing accuracy, showing the smallest \IntervLevelAdjPlain{} start
error ($1.3$\,s against $2.5$--$3.1$\,s). This indicates that streaming-native training acts
timelier than general-purpose training.

\subsection{Effects of Proactive Memory Acquisition}
\label{sec:proactive-memory-effect}

\begin{figure}[!t]
\centering
\definecolor{qaPropGroup}{HTML}{F8F2E5}
\definecolor{qaOpenGroup}{HTML}{E9E3DC}
\definecolor{qaVidGroup}{HTML}{DDCDB0}
\definecolor{qaStreamGroup}{HTML}{CBBAA8}

\begin{tikzpicture}[
  font=\scriptsize,
  qaNo/.style={draw=black, line width=1.5pt, line cap=round},
  qaMem/.style={draw=orange!45!black, line width=1.5pt, line cap=round},
  qaOracle/.style={draw=black!45, line width=1.5pt, line cap=round},
  qaNoDot/.style={circle, draw=black, fill=white, line width=0.5pt,
    inner sep=0.9pt},
  qaMemDot/.style={circle, draw=orange!45!black, fill=white, line width=0.5pt,
    inner sep=0.9pt},
  qaOracleDot/.style={circle, draw=black!45, fill=white, line width=0.5pt,
    inner sep=0.9pt}
]
\def\ColW{0.315}      %
\def\AxisW{0.75}      %
\def\PlotW{4.095}     %
\def\PanelStep{4.40}  %
\def\PlotH{1.75}      %
\pgfmathsetmacro{\YU}{0.01*\PlotH}  %

\newcommand{\QAStick}[4]{%
  \pgfmathsetmacro{\topnm}{max(#2,#3)}
  \draw[qaNo] (#1,0) -- (#1,#2);
  \draw[qaMem] (#1,#2) -- (#1,#3);
  \draw[qaOracle] (#1,\topnm) -- (#1,#4);
  \node[qaNoDot] at (#1,#2) {};
  \node[qaMemDot] at (#1,#3) {};
  \node[qaOracleDot] at (#1,#4) {};
}

\newcommand{\QAPanel}[3]{%
  \begin{scope}[shift={({\AxisW+#1*\PanelStep},0)}]
    \foreach \ca/\cb/\col in {0/1/qaPropGroup,2/5/qaOpenGroup,6/8/qaVidGroup,9/12/qaStreamGroup} {
      \fill[\col] ({\ca*\ColW},0) rectangle ({(\cb+1)*\ColW},\PlotH);
    }
    \node[font=\bfseries\footnotesize, anchor=south] at (0.5*\PlotW,\PlotH) {#2};
    \begin{scope}[y=\YU cm]
      \foreach \yt in {25,50,75} {
        \draw[black!25, densely dashed] (0,\yt) -- (\PlotW,\yt);
      }
      \draw[black!60] (0,0) -- (\PlotW,0);
      \draw[black!60] (0,0) -- (0,100);
      \foreach \yt in {0,50,100} { \draw[black!60] (-0.06,\yt) -- (0,\yt); }
      \foreach \i/\nomem/\mem/\oracle in {#3} {
        \pgfmathsetmacro{\xc}{(\i+0.5)*\ColW}
        \QAStick{\xc}{\nomem}{\mem}{\oracle}
      }
    \end{scope}
    \foreach \i/\lab in {0/{Gem.\ 2.5F}, 1/{GPT-5.4m}, 2/{Gemma 31B},
        3/{Q3.8 27B}, 4/{Q3.6 27B}, 5/{Q3.5 9B}, 6/{VLLaMA3},
        7/{LLaVA-OV}, 8/{MiniCPM}, 9/{MMDuet}, 10/{VLLM-Onl.},
        11/{Inkling}, 12/{JoyAI-VL}} {
      \node[rotate=60, anchor=east, inner sep=1pt] at ({(\i+0.5)*\ColW},-0.05) {\lab};
    }
  \end{scope}
}

\begin{scope}[y=\YU cm]
  \foreach \yt in {0,50,100} {
    \node[anchor=east, text=black!70, inner sep=1pt] at ({\AxisW-0.08},\yt) {\yt};
  }
  \node[rotate=90, anchor=south, text=black!70, inner sep=1pt] at (0.02,50) {Accuracy (\%)};
\end{scope}

\QAPanel{0}{\vctask{}}{
  0/50.0/52.9/88.2,
  1/51.7/52.5/77.9,
  2/58.6/66.5/89.5,
  3/57.0/62.7/94.9,
  4/56.8/60.4/88.0,
  5/48.1/54.0/79.9,
  6/39.1/50.5/77.3,
  7/37.7/41.0/74.0,
  8/35.5/36.1/79.3,
  9/38.7/40.2/65.7,
  10/0.0/0.0/0.0,
  11/48.7/49.5/90.7,
  12/49.1/52.3/89.2}

\QAPanel{1}{\sttask{}}{
  0/21.9/28.8/56.8,
  1/22.4/21.5/54.0,
  2/27.4/34.2/61.6,
  3/32.0/32.4/57.7,
  4/18.6/28.3/59.5,
  5/29.1/17.7/55.3,
  6/30.0/33.3/43.0,
  7/27.0/21.9/56.1,
  8/16.9/19.4/41.4,
  9/33.8/29.1/39.7,
  10/0.0/0.0/0.0,
  11/23.2/21.9/61.6,
  12/23.6/18.6/54.9}

\QAPanel{2}{\lttask{}}{
  0/24.0/18.3/58.7,
  1/20.2/24.6/62.7,
  2/34.1/29.0/61.1,
  3/36.1/30.6/67.5,
  4/25.8/24.2/65.9,
  5/25.0/21.8/63.5,
  6/22.6/36.1/62.7,
  7/21.0/23.8/61.1,
  8/17.5/18.3/58.3,
  9/23.0/29.0/60.3,
  10/0.0/0.0/0.0,
  11/27.0/29.0/63.1,
  12/27.0/30.6/64.3}

\begin{scope}[shift={(\AxisW+0.3,{\PlotH+0.62})}]
  \draw[qaNo, line width=2pt] (0,0) -- (0.3,0);
  \node[anchor=west, inner sep=1pt] at (0.35,0) {No Mem};
  \draw[qaMem, line width=2pt] (1.60,0) -- (1.90,0);
  \node[anchor=west, inner sep=1pt] at (1.95,0) {Proact Mem};
  \draw[qaOracle, line width=2pt] (3.55,0) -- (3.85,0);
  \node[anchor=west, inner sep=1pt] at (3.90,0) {Oracle Mem};
  \fill[qaPropGroup, draw=black!30] (5.65,-0.1) rectangle (5.9,0.1);
  \node[anchor=west, inner sep=1pt] at (5.95,0) {Proprietary};
  \fill[qaOpenGroup, draw=black!30] (7.45,-0.1) rectangle (7.7,0.1);
  \node[anchor=west, inner sep=1pt] at (7.75,0) {Open Source};
  \fill[qaVidGroup, draw=black!30] (9.40,-0.1) rectangle (9.65,0.1);
  \node[anchor=west, inner sep=1pt] at (9.70,0) {VidLLM};
  \fill[qaStreamGroup, draw=black!30] (10.85,-0.1) rectangle (11.1,0.1);
  \node[anchor=west, inner sep=1pt] at (11.15,0) {Streaming};
\end{scope}
\end{tikzpicture}
\captionsetup{skip=2pt}\caption{\label{fig:qa-results}
\textbf{QA accuracy by task}. The No-Memory-to-Oracle
gap is the acquisition headroom.}
\end{figure}
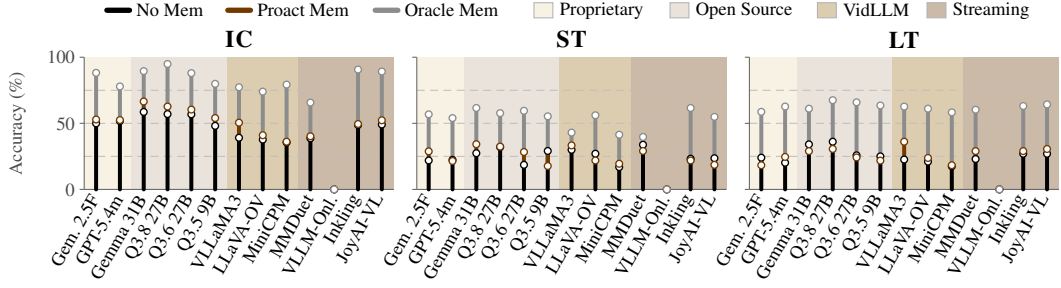

\noindent\textbf{The headroom is in proactive memory acquisition.} Timing and
lifecycle metrics score when an agent acts to acquire memory; ProactEgoMem-QA asks what that action leaves behind. A proactive assistant is useful only if the evidence it chooses to keep can answer the user's questions later, so we test the acquired memory directly. Every question is answered
after the stream ends under three memory conditions
(Figure~\ref{fig:qa-results}): uniform frames (\emph{No Memory}), the frames the agent kept throughout its tool calls (\emph{Proact Memory}), and the ground-truth event windows (\emph{Oracle Memory}). Proact Memory beats No Memory on
\vctask{} for every model with valid output and splits evenly on \sttask{} and
\lttask{}.  In $34$
of the $36$ valid model--task cells the oracle memory sits at least $20$ accuracy
points above the proactive one; with oracle memory ceiling at \vctask{}
$94.9\%$ while \sttask{} and \lttask{} stay below $70\%$. Even with oracle memory
fed to the video QA model, the benchmark remains hard to solve, and the gap
between No Memory and Oracle Memory defines the improvable headroom for proactive memory.

\subsection{From Daily Life to the Laboratory}
\label{sec:sci-transfer-main}

\begin{figure}[!t]
\centering
\noindent\begin{minipage}[b]{0.585\textwidth}
\centering\includegraphics{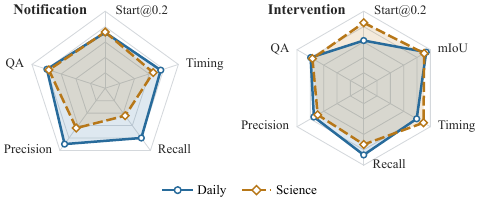}
\end{minipage}\hfill
\begin{minipage}[b]{0.405\textwidth}
\centering\includegraphics{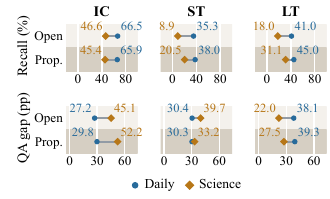}
\end{minipage}\par\nointerlineskip\vspace{-\parskip}
\noindent\begin{minipage}[t]{0.585\textwidth}\centering\scriptsize
\textbf{(a)} Mean of the $13$ models; shared axis ranges
\end{minipage}\hfill
\begin{minipage}[t]{0.405\textwidth}\centering\scriptsize
\textbf{(b)} By task; open-source and proprietary means
\end{minipage}\par
{\captionsetup{skip=3pt}\captionof{figure}{\label{fig:daily-sci-compact}\label{fig:radar-daily-sci}
\textbf{From daily life to the laboratory}: (a) the two tracks on the radar
axes of Figure~\ref{fig:radars}; (b) notification recall and QA acquisition
gap per task.}\par}
\end{figure}

\noindent\textbf{Scientific scenarios expose the central bottleneck in
proactive memory acquisition.}
We measure it as the \emph{memory acquisition gap}, oracle QA accuracy minus QA
accuracy under proactively acquired memory. On IC the gap nearly doubles in
the laboratory, from $27$--$30$ to $45$--$52$ points for the open-source and
proprietary groups (Figure~\ref{fig:daily-sci-compact}b, \vctask{}). The additional complexity of managing and maintaining the video data in a scientific context adds to the overall burden of proactive memory.

\noindent\textbf{This burden concentrates on IC and ST.} Laboratory video
lowers notification recall for both groups on all three tasks, halving the model-mean
recall (Figure~\ref{fig:daily-sci-compact}a), most sharply on
ST, where open-source recall falls from $35.3\%$ to $8.9\%$
(Figure~\ref{fig:daily-sci-compact}b, \sttask{}); the ST gap widens to $33$--$40$ points,
while LT gaps narrow to $22$--$27$ points
(Figure~\ref{fig:daily-sci-compact}b, \lttask{}). \vctask{} and \sttask{} are harder in the lab, whose objects are hard to recognize; \lttask{} is easier, as lab objects rarely move.

\subsection{Failure Modes}
\label{sec:failure-modes}

\begin{figure}[!t]
\centering
\noindent\begin{minipage}[b]{0.625\textwidth}
\centering
\includegraphics[width=\linewidth]{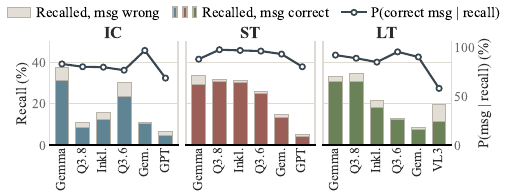}
\end{minipage}\hspace{0.005\textwidth}%
\begin{minipage}[b]{0.37\textwidth}
\centering
\includegraphics[width=\linewidth]{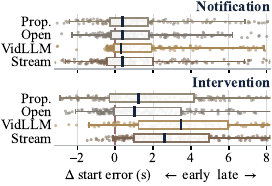}
\end{minipage}\par\nointerlineskip\vspace{-\parskip}
\noindent\begin{minipage}[t]{0.625\textwidth}
\captionsetup{skip=3pt}\captionof{figure}{\label{fig:conditional-stopmsg}
\textbf{Message accuracy conditioned on recall}}
\end{minipage}\hspace{0.005\textwidth}%
\begin{minipage}[t]{0.37\textwidth}
\captionsetup{skip=3pt}\captionof{figure}{\label{fig:start-families}\textbf{Start error by family}}
\end{minipage}
\end{figure}

\vspace{2pt}

\noindent\textbf{Detection of the memory window is the main bottleneck: models that could detect an event mostly explain it correctly.}
At the \IntervLevelPlain{}, a text-only judge Deepseek-V4-Flash checks whether the message closing
a tool call names the right object and outcome against a template reference
(Appendix~\ref{sec:proactive-message}), for models with recall above $5\%$.
Conditioned on its own true positives, a model explains a recalled intervention
correctly $87.8\%$ of the time on average (Figure~\ref{fig:conditional-stopmsg}),
nearly independent of recall (Spearman $r=0.23$).

\noindent\textbf{Proactive actions arrive late, and models are still reactive by nature.} Figure~\ref{fig:start-families} pools every matched start error by model family: medians fall near $+0.4$\,s at the \NotifLevelPlain{} and from $+1.0$ to
$+3.5$\,s at the \IntervLevelPlain{}, where $67$--$93\%$ of starts are late. The families that
enter the fewest lifecycles are also the latest to enter them. The lag reflects the models' inherent reactivity.

\vspace{2pt}
\par\vspace{2pt}
\begin{figure}[!t]
{\centering
\input{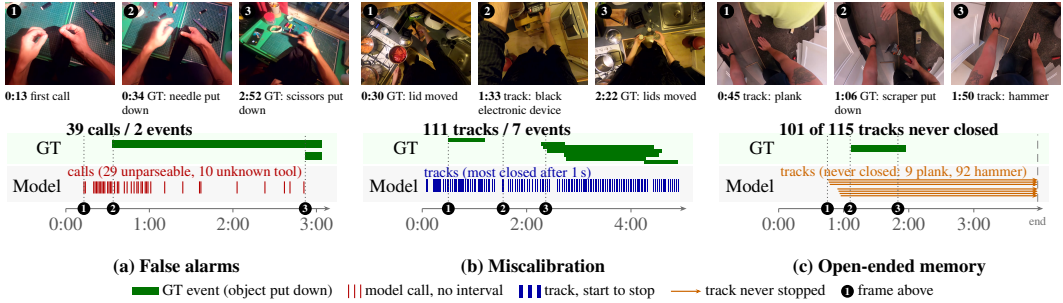}\par
\captionsetup{skip=3pt}\captionof{figure}{\label{fig:false-alarm}\textbf{Failure
modes on \lttask{} clips} (\IntervLevelPlain{}): (a) MMDuet, $39$ calls for $2$ GT
events; (b) Gemma4-31B, $111$ one-second tracks for $7$ events;
(c) VideoLLaMA3-7B, tracks never stopped.}\par}
\end{figure}
\vspace{2pt}

\noindent\textbf{Uncalibrated interruption: coverage is bought with false alarms.}
At the \NotifLevelPlain{}, recall reaches $79.5\%$ at single-digit
precision, capping the best F1 at $19.0\%$
(Appendix~\ref{sec:proactive-level-results}). Gemma4-31B opens $111$ short
tracks in one clip; MMDuet fires $39$ calls for $2$ events
(Figures~\ref{fig:false-alarm}a--b and~\ref{fig:fc-overtrigger}).

\noindent\textbf{Lifecycle maintenance: interventions are opened but rarely held.}
Once a tool must start, persist, and stop, overlap stays
single-digit and recall peaks at $37.9\%$
(Appendix~\ref{sec:proactive-level-results}). VideoLLaMA3-7B re-opens a track
every second its target is visible and never closes it
(Figures~\ref{fig:false-alarm}c, \ref{fig:fc-lifecycle}, and~\ref{fig:fc-stop}).

\noindent\textbf{Video-language models fail to follow instructions as an agent.}
Some models leave the tool protocol altogether: VideoLLM-Online narrates the
scene (``You open the fridge.'') and MMDuet states its decisions in free text,
so neither emits a parsable tool call, and VideoLLM-Online opens no valid
intervention on any task (Figure~\ref{fig:fc-protocol};
Appendix~\ref{app:failure-cases}), showing the limitations of acting as agents.

\section{Conclusion}
\label{sec:conclusion}

\method evaluates proactive ambient memory in streaming egocentric video through
two proactive modes and three task families, \vctask{}, \sttask{} and
\lttask{}, across daily-life and laboratory co-scientist tracks. The
experiments show that today's multimodal models remain far from reliable
proactive assistants: they identify the right moment only under loose timing
windows, and they struggle with precision, event tracking over long-horizon
location, and proactive memory acquisition. It charts a concrete path toward
wearable agents that remember useful context without being asked.

\FloatBarrier

\section*{AI Use Statement}

In this work, we used generative AI tools as components of the benchmark
pipeline: a VLM writes the dense captions from which candidate
events are filtered (Section~\ref{sec:dataset}), an LLM generates the \vctask{}
questions from human-authored dense annotations of visible information
(Section~\ref{sec:qa-construction}), an LLM-judged semantic similarity gates the
matching of predicted to ground-truth actions (Section~\ref{sec:metrics}). Regarding LLM, we use DeepSeek-V4-Flash judges the correctness of proactive messages
(Appendix~\ref{sec:proactive-message}). We have not used generative AI tools
for other tasks with required disclosure, and the rest of the
required disclosure tasks are not applicable to this work. Additionally, we
used generative AI tools for improving the manuscript text and its \LaTeX{}
layout and for polishing the scripts that draw the figures. We have reviewed all
AI-assisted work. Three independent annotators per clip mark the event
boundaries, the LLM-generated questions were reviewed in three rounds, and AI-assisted text and figure code were checked, e.g.\ figure values
verified against the result tables by the authors according to the raw experiment logs. We take responsibility for
the final content of this work, including text, claims or artifacts produced
with the aid of generative AI.

\section*{Reproducibility}
The ground-truth answers stay private: an evaluation
server accepts blind submissions and returns only the scores, which keeps the
benchmark out of web-scale training data for foundation models.

{
\small
\bibliography{main}
}

\clearpage
\appendix
\raggedbottom
\section{Comparison with Existing Benchmarks}
\label{sec:benchmark-comparison}

Table~\ref{tab:benchmark-comparison} situates \method among existing video
benchmarks.
Offline suites
\citep{fu2025videomme,wu2024longvideobench,zhou2025mlvu,mangalam2023egoschema,chandrasegaran2024hourvideo,yang2025egolife}
evaluate query-driven answers over completed clips; streaming benchmarks
\citep{lin2024streamingbench,li2025ovobench} still prompt the model at
chosen timestamps; streaming episodic-memory benchmarks
\citep{forte2026egostream,wang2026sember} test recall from a continuous
stream but supply the question; and a fast-growing proactive line
\citep{wang2025omnimmi,zhang2025proassist,wang2025proactivevideoqa,gong2026vinci2,ran2026egoprobench,kundu2026planwatchrecover,du2026proactivebench}
drops the query yet scores only a single respond-now decision.
\method uniquely combines streaming egocentric input, query-free
proactivity, and a start--maintain--stop tool lifecycle whose acquired
memory is then verified by downstream questions, graded jointly on timing
and semantics.

\begin{table}[!t]
\centering
\caption{\label{tab:benchmark-comparison}\textbf{Benchmark comparison}.
\emph{Proactive}: acting without an explicit query; \emph{Multi-Level}: two
proactive depths; \emph{Tool Lifecycle}: a maintained background lifecycle;
\emph{Mem.\ QA}: questions verify self-acquired memory. (\cmark) marks a
partial case; \method marks all four.}
\resizebox{\textwidth}{!}{%
\renewcommand{\arraystretch}{0.95}%
\begin{tabular}{@{}l c c c c c c r l@{}}
\toprule
\textbf{Benchmark} & \textbf{Video Domain} & \textbf{Setting} & \textbf{Proactive} & \textbf{Multi-Level} & \textbf{Tool Lifecycle} & \textbf{Mem.\ QA} & \textbf{Ave.\ Dur.\ (s)} & \textbf{Response Type} \\
\midrule
\multicolumn{9}{@{}l}{\emph{Offline video QA}}\\
Video-MME \citep{fu2025videomme} & General & Offline & \xmark & \xmark & \xmark & \xmark & 1018 & Multi-choice \\
LongVideoBench \citep{wu2024longvideobench} & General & Offline & \xmark & \xmark & \xmark & \xmark & 473 & Multi-choice \\
MLVU \citep{zhou2025mlvu} & General & Offline & \xmark & \xmark & \xmark & \xmark & 930 & Multi-choice \\
EgoSchema \citep{mangalam2023egoschema} & Egocentric & Offline & \xmark & \xmark & \xmark & \xmark & 180 & Multi-choice \\
HourVideo \citep{chandrasegaran2024hourvideo} & Egocentric & Offline & \xmark & \xmark & \xmark & \xmark & 2742 & Multi-choice \\
EgoLifeQA \citep{yang2025egolife} & Egocentric & Offline & \xmark & \xmark & \xmark & \xmark & $\sim$44\,h & Multi-choice \\
HoloAssist \citep{wang2023holoassist} & Egocentric & Offline & \xmark & \xmark & \xmark & \xmark & \na & Intervention type \\
Ego2Web \citep{yu2026ego2web} & Egocentric & Offline & \xmark & \xmark & \xmark & \xmark & 180 & Web actions \\
\midrule
\multicolumn{9}{@{}l}{\emph{Streaming QA (prompted)}}\\
StreamingBench \citep{lin2024streamingbench} & General & Streaming & (\cmark) & \xmark & \xmark & \xmark & \na & Multi-choice \\
OVO-Bench \citep{li2025ovobench} & General & Streaming & (\cmark) & \xmark & \xmark & \xmark & $\sim$429 & Multi-choice \\
\midrule
\multicolumn{9}{@{}l}{\emph{Streaming episodic memory}}\\
EgoStream \citep{forte2026egostream} & Egocentric & Streaming & (\cmark) & \xmark & \xmark & (\cmark) & \na & Open-ended \\
S-EMBER \citep{wang2026sember} & Egocentric & Streaming & (\cmark) & \xmark & \xmark & (\cmark) & $\sim$445 & Open-ended \\
\midrule
\multicolumn{9}{@{}l}{\emph{Proactive interaction}}\\
OmniMMI \citep{wang2025omnimmi} & General & Streaming & \cmark & \xmark & \xmark & \xmark & $\sim$611 & Open-ended \\
ProAssist \citep{zhang2025proassist} & Egocentric & Streaming & \cmark & \xmark & \xmark & \xmark & $\sim$438 & Dialogue \\
ProactiveVideoQA \citep{wang2025proactivevideoqa} & General & Streaming & \cmark & \xmark & \xmark & \xmark & $\sim$126 & Open-ended \\
EgoServe (Vinci2) \citep{gong2026vinci2} & Egocentric & Streaming & \cmark & \xmark & \xmark & \xmark & \na & Service response \\
EgoPro-Bench \citep{ran2026egoprobench} & Egocentric & Streaming & \cmark & \xmark & \xmark & \xmark & \na & Dialogue \\
Pro$^2$Bench / EgoProactive \citep{kundu2026planwatchrecover} & Egocentric & Streaming & \cmark & \xmark & \xmark & \xmark & $\sim$2310 & Guidance \\
LiveProBench \citep{du2026proactivebench} & General & Streaming & \cmark & \xmark & \xmark & \xmark & \na & Open-ended \\
\midrule
\method (ours) & Egocentric (Daily + Sci) & Streaming & \cmark & \cmark & \cmark & \cmark & 202 & Timed actions + Multi-choice \\
\bottomrule
\end{tabular}}
\end{table}

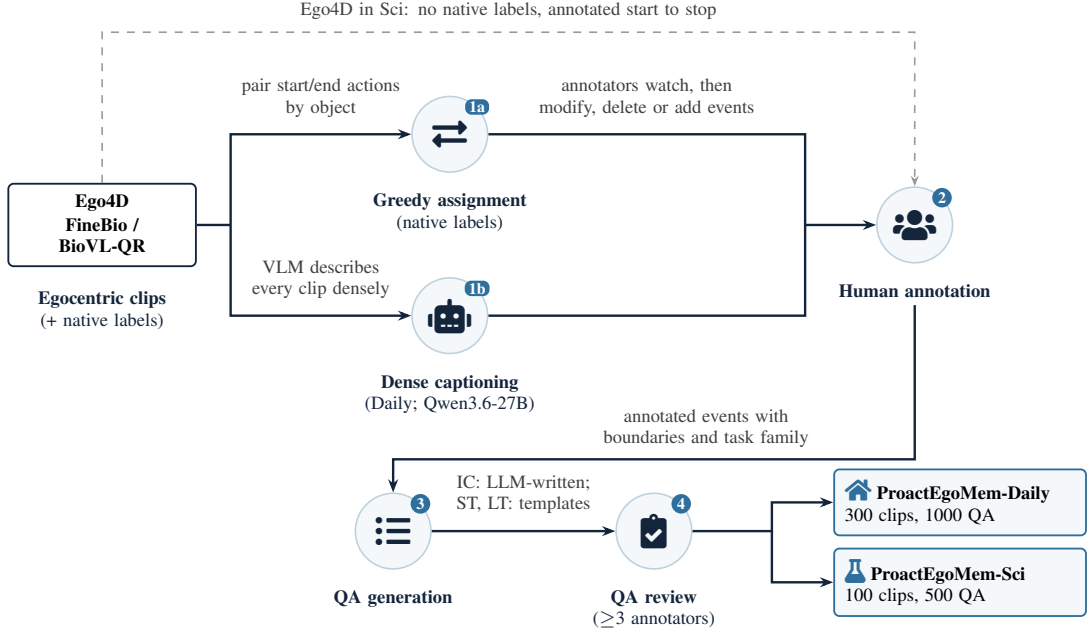
\begin{figure}[!t]
\centering
\begin{tikzpicture}[
  font=\scriptsize,
  icon/.style={circle, fill=teaserSoft, draw=teaserLine, line width=0.6pt,
    minimum size=1.0cm, inner sep=0pt, text=teaserNavy,
    font=\fontsize{13}{13}\selectfont},
  badge/.style={circle, fill=teaserBlue, text=white, inner sep=0.8pt,
    font=\fontsize{6}{6}\selectfont\bfseries},
  pbadge/.style={badge, rectangle, rounded corners=2.4pt, inner xsep=1.5pt},
  lbl/.style={align=center, text width=2.6cm, font=\scriptsize\bfseries,
    text=teaserNavy},
  acttxt/.style={align=center, text width=1.95cm, font=\scriptsize, text=black!75},
  flow/.style={-{Stealth[length=5pt]}, draw=teaserNavy, line width=0.9pt},
  link/.style={draw=teaserNavy, line width=0.9pt},
  cut/.style={-{Stealth[length=4pt]}, draw=black!40, line width=0.6pt, dashed},
  card/.style={rounded corners=2pt, draw=teaserBlue, line width=0.7pt,
    fill=teaserSoft, align=left, text width=3.05cm, inner sep=4pt},
  inbox/.style={rounded corners=2pt, draw=teaserNavy, line width=0.7pt,
    fill=white, align=center, text width=2.2cm, inner sep=4pt},
]
\node[inbox] (in) at (0,0)
  {\textbf{Ego4D}\\[1pt]\textbf{FineBio / BioVL-QR}};
\node[lbl, below=2mm of in] {Egocentric clips\\\mdseries(+ native labels)};
\node[icon] (pair)  at (4.6,1.2)  {\faIcon{exchange-alt}};
\node[icon] (cap)   at (4.6,-1.2) {\faRobot};
\node[icon] (annot) at (10.75,0)  {\faUsers};
\node[pbadge] at (pair.north east) {1a};
\node[pbadge] at (cap.north east)  {1b};
\node[badge]  at (annot.north east) {2};
\node[lbl, below=1.5mm of pair]  {Greedy assignment\\\mdseries(native labels)};
\node[lbl, below=1.5mm of cap]   {Dense captioning\\\mdseries(Daily; Qwen3.6-27B)};
\node[lbl, below=1.5mm of annot] {Human annotation};

\coordinate (fork) at (1.7,0);
\draw[link] (in.east) -- (fork);
\draw[flow] (fork) |- node[acttxt, above=1mm, pos=0.75, text width=2.1cm]
  {pair start/end actions\\by object} (pair.west);
\draw[flow] (fork) |- node[acttxt, above=1mm, pos=0.75, text width=2.1cm]
  {VLM describes\\every clip densely} (cap.west);

\coordinate (merge) at (9.3,0);
\draw[link] (pair.east) -- node[acttxt, above=1mm, text width=3.4cm]
  {annotators watch, then\\modify, delete or add events} (pair.east -| merge) -- (merge);
\draw[link] (cap.east) -- (cap.east -| merge) -- (merge);
\draw[flow] (merge) -- (annot.west);

\draw[cut] (in.north) -- (in.north |- 0,2.55)
  -- node[acttxt, above=0.5mm, text width=7cm]
     {Ego4D in Sci: no native labels, annotated start to stop}
  (annot.north |- 0,2.55) -- (annot.north);

\node[icon] (qa)     at (3.85,-4.05) {\faListUl};
\node[icon] (review) at (7.3,-4.05)  {\faClipboardCheck};
\node[badge] at (qa.north east)     {3};
\node[badge] at (review.north east) {4};
\node[lbl, below=1.5mm of qa]     {QA generation};
\node[lbl, below=1.5mm of review] {QA review\\\mdseries($\geq$3 annotators)};

\draw[flow] ($(annot.south)+(0,-0.55)$) |- (qa.north |- 0,-3.1) coordinate (turn)
  node[acttxt, above=0.3mm, text width=4.8cm, pos=0.7]
     {annotated events with boundaries and task family}
  -- (qa.north);

\draw[flow] (qa.east) -- node[acttxt, above=1mm] {IC: LLM-written;\\ST, LT: templates} (review.west);

\node[card] (daily) at (11.35,-3.67)
  {\textcolor{teaserBlue}{\fontsize{9}{9}\selectfont\faHome}\;\textbf{\dailymethod{}}\\
   \dailyclips{} clips, \dailyqa{} QA};
\node[card, below=1.5mm of daily] (sci)
  {\textcolor{teaserBlue}{\fontsize{9}{9}\selectfont\faFlask}\;\textbf{\scimethod{}}\\
   \sciclips{} clips, \sciqa{} QA};
\coordinate (outbus) at ($(review.east)+(1.05,0)$);
\draw[flow] (review.east) -- (outbus) |- (daily.west);
\draw[flow] (outbus) |- (sci.west);
\end{tikzpicture}
\caption{\label{fig:construction-pipeline}
\textbf{\method construction pipeline}. Native labels are greedily paired into candidate events; for Daily, a VLM densely captions every clip in addition to the greedy pairing. Annotators then watch each clip and modify, delete, or add events (Ego4D clips in Sci are annotated from scratch because native labels are missing). Every QA is reviewed by at least three annotators before being finalized.}
\end{figure}

\Needspace{14\baselineskip}

\section{Data Leakage Mitigation and Annotation Novelty}
\label{sec:data-leakage-mitigation}

Although raw video streams in \method{} are sampled from public egocentric repositories (e.g., Ego4D~\citep{9879279}, FineBio~\citep{yagi2025finebio}), the benchmark evaluates capabilities that cannot be solved through visual frame memorization or pre-training textual priors. We guarantee that \method{} serves as an out-of-distribution evaluation through three key structural mechanisms:

\begin{enumerate}
    \item \textbf{Novel Proactive Task Formulations:} Existing video datasets focus on passive captioning or action segmentation. In contrast, \method{} evaluates proactive tool lifecycles, real-time trigger notifications, and longitudinal tracking (\vctask{}, \sttask{}, \lttask{}). The defined target action boundaries ($Start@\tau$), tool activation states, and intervention logic do not exist in the source datasets.
    
    \item \textbf{Newly Generated QA Pairs and Templates:} All downstream evaluation questions are entirely original. \vctask{} questions are synthesized via LLM generation from dense captions and subjected to a three-round human review. \sttask{} and \lttask{} QA items are programmatically instantiated over human-verified event timelines using custom slot templates (e.g., tracking sequence order, active state durations, and destination locations).
    
    \item \textbf{Dense, Multi-Stage Human Grounding:} Every clip undergoes fresh annotation by three independent annotators (holding a bachelor's degree or higher) to establish precise temporal boundaries and ground-truth answers. Because the evaluation protocol assesses timing precision and proactive state tracking simultaneously, pre-trained models cannot rely on web-scraped visual captions or pre-training text to succeed.
\end{enumerate}

\section{Tool Interface}
\label{app:tool-interface}
The benchmark exposes three core tools for proactive ambient memory:

\vspace{-8pt}
\noindent\begin{minipage}[t]{\linewidth}
    \begin{minipage}[t]{0.51\linewidth}\footnotesize
    \vspace{0pt}\raggedright
        \begin{itemize}[leftmargin=1.2em, labelsep=0.45em, itemsep=1pt,
      topsep=0pt, parsep=0pt, partopsep=0pt]
      \item \textbf{Recorder:} records visible information.
      \item \textbf{State Tracker:} keeps an object state until resolved.
      \item \textbf{Location Tracker:} tracks an object until it moves.
      \item \textbf{Tool-call content:} a start call names the target with its description, state, or location; a tracker's start returns an \texttt{id} that its stop call closes.
      \item \textbf{Minimal interface:} the difficulty lies in tool calling maintenance; the API surface is deliberately small.
    \end{itemize}
  \end{minipage}\hfill
  \begin{minipage}[t]{0.47\linewidth}
    \vspace{0pt}
    \toolcard{
      \textbf{Recording Tool}\hfill\textbf{\vctask{}}\\
      \texttt{start\_recording(name, description)}\\
      \texttt{stop\_recording()}}

    \vspace{2pt}
    \toolcard{
      \textbf{State Tracking Tool}\hfill\textbf{\sttask{}}\\
      \mbox{\texttt{start\_state\_tracking(name, state)}$\,\rightarrow\,$\texttt{id}}\\
      \texttt{stop\_state\_tracking(id)}}

    \vspace{2pt}
    \toolcard{
      \textbf{Location Tracking Tool}\hfill\textbf{\lttask{}}\\
      \mbox{\texttt{start\_location\_tracking(name, location)}$\,\rightarrow\,$\texttt{id}}\\
      \texttt{stop\_location\_tracking(id)}}
  \end{minipage}

\end{minipage}

\section{Quality of the Proactive Message}
\label{sec:proactive-message}
We further analyze the quality of the message being sent at the end of the tool call at the \IntervLevelPlain{}. 

\noindent\textbf{Protocol.} For every ground-truth message, we build a
reference message by template from its arguments, including object names, tool, and status, building strings such as \emph{``[Nutrition facts] has been recorded''},
\emph{``Keys location has been placed at [a new location]''}, or \emph{``[Stove status] tracked until [turned off]''}. The candidate is the message that the model emitted at the end of its tool call. A separate text-only judge, DeepSeek-V4-Flash, returns a binary
score for whether the message conveys the correct information. We condition on the model's own true positives and report $P(\text{correct msg}\mid\text{recall})$; a model is included in a task only if its \IntervLevelPlain{} recall there exceeds a minimum threshold of $5\%$.

\noindent\textbf{Detection is the bottleneck; Models tend to emit correct messages conditional on successful detection. Quality of proactive messages still varies by model}. Across the
$18$ model in Figure~\ref{fig:conditional-stopmsg}, a recalled
intervention is correctly explained above $76\%$. Once a
model notices that a lifecycle should end, it usually names the right object and
the right outcome. The quality of the conditional message is also close to independent of recall value
(with Spearman value of $r=0.23$), showing that models that fire rarely are not thereby more careful about
what they say. Accuracy of proactive messages still varies by model, causing conditional message accuracy to range from 58\% (VideoLLaMA3-7B on \lttask{}) to 97\% (Inkling-small on \sttask{}). VideoLLaMA3-7B is the
substantive case: it recalls $19.8\%$ of \lttask{} lifecycles, more than
Qwen3.6-27B does, yet explains barely half of them correctly, so its joint rate ($11.5\%$) falls below Qwen3.6-27B's ($12.5\%$).

\takeaway{Detection Is the Bottleneck and Proactive Message Accuracy Varies by Models}{Once a model
notices that a lifecycle should end, it names the right object and outcome
on average $87.8\%$ of the time. Model performance varies across tasks and models, and some fail to represent what happened at the right moments, which remains as a challenge.}

\section{Effects of Timing in Proactivity}
\label{sec:timing-thresholds}

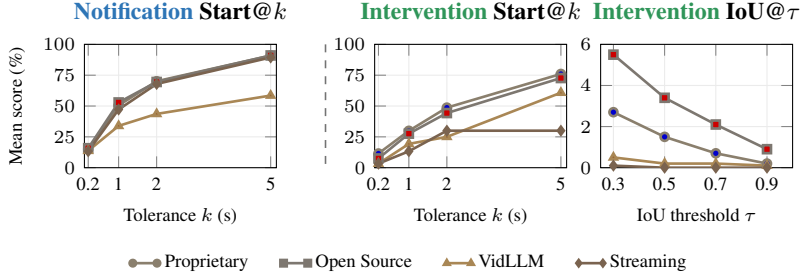
\begin{figure}[!t]
\centering
\definecolor{thresholdProp}{HTML}{8A7F6D}
\definecolor{thresholdOpen}{HTML}{7D7770}
\definecolor{thresholdVid}{HTML}{A98755}
\definecolor{thresholdStream}{HTML}{7A634F}
\begin{tikzpicture}
\begin{axis}[
  name=notifstart,
  width=0.295\textwidth,
  height=0.23\textwidth,
  xmin=0.1, xmax=5.15,
  ymin=0, ymax=100,
  xtick={0.2,1,2,5},
  xticklabels={0.2,1,2,5},
  ytick={0,25,50,75,100},
  grid=major,
  grid style={black!8},
  xlabel={Tolerance $k$ (s)},
  ylabel={Mean score (\%)},
  title={\Notif Start@$k$},
  title style={font=\small\bfseries},
  tick label style={font=\scriptsize},
  label style={font=\scriptsize},
  every axis plot/.append style={line width=1pt, mark options={solid, scale=0.75}}
]
\addplot+[color=thresholdProp, mark=*] coordinates {(0.2,15.0) (1,50.5) (2,70.1) (5,90.9)};
\addplot+[color=thresholdOpen, mark=square*] coordinates {(0.2,15.8) (1,52.9) (2,69.5) (5,91.0)};
\addplot+[color=thresholdVid, mark=triangle*] coordinates {(0.2,13.7) (1,33.9) (2,43.6) (5,58.5)};
\addplot+[color=thresholdStream, mark=diamond*] coordinates {(0.2,13.3) (1,47.2) (2,67.9) (5,89.4)};
\end{axis}

\begin{axis}[
  name=intervstart,
  at={(notifstart.east)},
  anchor=west,
  xshift=1.30cm,
  width=0.295\textwidth,
  height=0.23\textwidth,
  xmin=0.1, xmax=5.15,
  ymin=0, ymax=100,
  xtick={0.2,1,2,5},
  xticklabels={0.2,1,2,5},
  ytick={0,25,50,75,100},
  grid=major,
  grid style={black!8},
  xlabel={Tolerance $k$ (s)},
  title={\Interv Start@$k$},
  title style={font=\small\bfseries},
  tick label style={font=\scriptsize},
  label style={font=\scriptsize},
  every axis plot/.append style={line width=1pt, mark options={solid, scale=0.75}}
]
\addplot+[color=thresholdProp, mark=*] coordinates {(0.2,11.5) (1,29.9) (2,48.7) (5,76.0)};
\addplot+[color=thresholdOpen, mark=square*] coordinates {(0.2,7.5) (1,27.6) (2,44.3) (5,72.7)};
\addplot+[color=thresholdVid, mark=triangle*] coordinates {(0.2,3.3) (1,19.3) (2,25.0) (5,60.8)};
\addplot+[color=thresholdStream, mark=diamond*] coordinates {(0.2,3.3) (1,13.3) (2,30.0) (5,30.0)};
\end{axis}

\begin{axis}[
  name=interviou,
  at={(intervstart.east)},
  anchor=west,
  xshift=0.45cm,
  width=0.295\textwidth,
  height=0.23\textwidth,
  xmin=0.25, xmax=1.0,
  ymin=0, ymax=6,
  xtick={0.3,0.5,0.7,0.9},
  xticklabels={0.3,0.5,0.7,0.9},
  ytick={0,2,4,6},
  grid=major,
  grid style={black!8},
  xlabel={IoU threshold $\tau$},
  title={\Interv IoU@$\tau$},
  title style={font=\small\bfseries},
  tick label style={font=\scriptsize},
  label style={font=\scriptsize},
  every axis plot/.append style={line width=1pt, mark options={solid, scale=0.75}}
]
\addplot+[color=thresholdProp, mark=*] coordinates {(0.3,2.7) (0.5,1.5) (0.7,0.7) (0.9,0.2)};
\addplot+[color=thresholdOpen, mark=square*] coordinates {(0.3,5.5) (0.5,3.4) (0.7,2.1) (0.9,0.9)};
\addplot+[color=thresholdVid, mark=triangle*] coordinates {(0.3,0.5) (0.5,0.2) (0.7,0.2) (0.9,0.1)};
\addplot+[color=thresholdStream, mark=diamond*] coordinates {(0.3,0.1) (0.5,0.0) (0.7,0.0) (0.9,0.0)};
\end{axis}
\draw[dashed, line width=0.6pt, black!55]
  ($(notifstart.north east)!0.5!(intervstart.north west)$)
  --
  ($(notifstart.south east)!0.5!(intervstart.south west)$);
\end{tikzpicture}
\vspace{2pt}

{\scriptsize
\begin{tabular}{@{}cccc@{}}
\tikz[baseline=-0.5ex]\draw[color=thresholdProp, line width=1pt, mark=*, mark size=1.2pt] plot coordinates {(0,0) (0.32,0)}; Proprietary &
\tikz[baseline=-0.5ex]\draw[color=thresholdOpen, line width=1pt, mark=square*, mark size=1.2pt] plot coordinates {(0,0) (0.32,0)}; Open Source &
\tikz[baseline=-0.5ex]\draw[color=thresholdVid, line width=1pt, mark=triangle*, mark size=1.2pt] plot coordinates {(0,0) (0.32,0)}; VidLLM &
\tikz[baseline=-0.5ex]\draw[color=thresholdStream, line width=1pt, mark=diamond*, mark size=1.2pt] plot coordinates {(0,0) (0.32,0)}; Streaming
\end{tabular}}
\caption{\label{fig:threshold-curves}
Group-level mean Start@$k$ and intervention IoU@$\tau$ over tasks and models
as each criterion is relaxed. Note the collapsed scale of the rightmost
panel: the best group mean stays below 6\%.}
\end{figure}

Start@0.2 measures whether a matched action lands at interactive latency, but a
single tolerance does not show whether failures are narrowly late or far from
the intended boundary. Figure~\ref{fig:threshold-curves} therefore sweeps the
start tolerance and the intervention-overlap criterion before we examine the
signed error distributions.

\paragraph{Effect of the Timing Threshold.} 

(1)Given loose tolerances, the models can achieve high coverage: Start@1.0 reaches $50$--$80\%$ for most models at the \NotifLevelPlain{} and $30$--$60\%$ at the \IntervLevelPlain{}. However, the curves are steep: Start@0.5 drops to $20$--$40\%$ and $10$--$30\%$, respectively, and Start@0.2 falls to $10$--$30\%$ and $5$--$20\%$. The steepness indicates that most response are hard to arrive on time. (2) VLMs, no matter open-source or proprietary, perform better than VidLMs in general, and the proprietary models do not show a consistent advantage over open-source models. (3) The \IntervLevelPlain{} is uniformly harder than the \NotifLevelPlain{}, and the curves are steeper, so no model is at once timely, able to maintain tool use, and proactive in memory acquisition.

\begin{figure}[!t]
\centering
{\bfseries ProactEgoMem -- predicted start error per model}\par\vspace{0.35em}
\includegraphics[width=\textwidth]{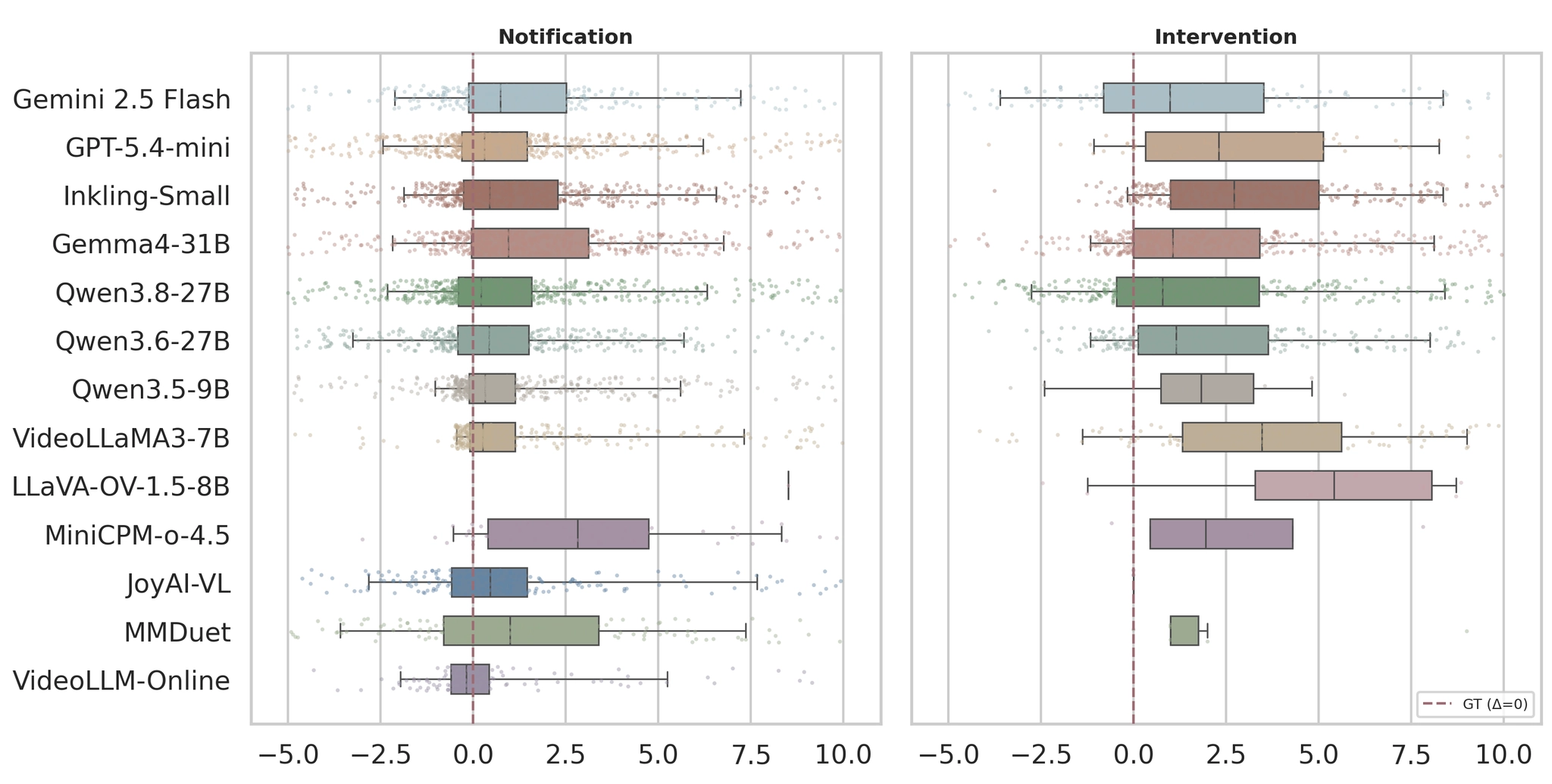}\par\vspace{0.15em}
{\small start error $\Delta$ = predicted $-$ GT (s) \quad $\leftarrow$ early \quad late $\rightarrow$}
\caption{\label{fig:start-boxplots}
Predicted start-error distributions by model, pooled across the three tasks,
for notification and intervention levels. Points are individual matched
starts; the dashed vertical line marks the ground-truth boundary, with positive
values late and negative values early. Empty intervention rows indicate that
no valid matched intervention was available for that model.}
\end{figure}
\FloatBarrier

\noindent\textbf{Timing is uncalibrated and responses usually arrive late.} Even a
semantically matched notification rarely lands on time: group-mean Start@0.2
sits near $13$--$16\%$, so fewer than one matched notification in six falls
within the $200$\,ms tolerance. Across emitting model--task cells, mean
absolute error is $2.07$\,s (range $0.73$--$8.5$\,s), far beyond responsive
sub-second latency. Figure~\ref{fig:start-boxplots} shows notification medians
often near the boundary but with long right tails; at the \IntervLevelPlain{}, all
$11$ valid pooled distributions have a positive median, placing the typical
tool start after the moment when it was needed.

\noindent\textbf{Committing to a tool generally costs about another second of
delay, but not for every model--task pair.} Mean absolute start error rises
from $2.07$\,s at the \NotifLevelPlain{} to $3.02$\,s at the \IntervLevelPlain{}; among
the $28$ cells with predictions at both levels, $22$ become later and six
improve. The pooled boxes expose the same asymmetry: intervention distributions
shift visibly right for GPT-5.4-mini, Inkling-Small, VideoLLaMA3-7B, and
Qwen3.5-9B, while JoyAI-VL and VideoLLM-Online produce no valid
intervention distribution. Tool commitment therefore usually adds delay and
can also end in abstention.

\noindent\textbf{The lag varies too much to be calibrated away.} Start-error spread stays near $2$\,s at both levels (mean $\sigma$ of
$2.01$\,s and $2.21$\,s), and the boxes differ in both width and location. Some
models have broad late tails, whereas MMDuet's intervention box is compact and
LLaVA-OV-1.5-8B's notification estimate collapses to one very late value. A
model-level mean shift would leave substantial dispersion and would not
reliably bring starts inside the $200$\,ms tolerance; what is missing is a
reliable boundary signal.

\takeaway{Proactive Timing Is Late and Uncalibrated}{Fewer than one matched
notification in six lands within the $200$\,ms tolerance, and committing to a
tool adds about another second (mean absolute start error $2.07$\,s
$\rightarrow$ $3.02$\,s). Start-error spread stays near $2$\,s at both levels,
so no constant offset could calibrate the lag away; what is missing is a
reliable boundary signal.}

\section{\scimethod: from Daily Life to the Laboratory}
\label{sec:sci-transfer}

\scimethod is the parallel Science track of \method: \sciclips{} egocentric
laboratory clips (\sciicclips{} \vctask{}, \scistclips{} \sttask{}, and
\sciltclips{} \lttask{}) with \sciqa{} QA questions, evaluated with identical
task settings, tool interfaces, and matching protocol as \dailymethod{}. Full
per-model results are reported in Appendix~\ref{app:sci-main-results}
(Table~\ref{tab:sci-merged-main-results}), with the tolerance sweeps in
Figure~\ref{fig:sci-threshold-curves} and the QA memory conditions in
Figure~\ref{fig:sci-qa-results}.

\begin{figure}[!t]
\centering
\begin{tikzpicture}
\begin{axis}[
  name=panela, xbar, width=0.40\textwidth, height=0.36\textwidth,
  bar width=5pt, xmin=-17.5, xmax=7, xtick={-15,-10,-5,0,5},
  symbolic y coords={QA Acc,Interv S@0.2,Interv F1,Interv mIoU,Notif S@0.2,Notif Prec,Notif F1,Notif Recall}, ytick=data,
  xlabel={Science $-$ Daily (points)},
  title={(a) Paired change per metric},
  title style={font=\small\bfseries}, tick label style={font=\scriptsize},
  label style={font=\scriptsize}, grid=major, grid style={black!8},
  nodes near coords, nodes near coords style={font=\scriptsize,
    /pgf/number format/.cd, fixed, fixed zerofill, precision=1, /tikz/.cd},
  nodes near coords align={horizontal}, enlarge y limits=0.09,
]
\addplot+[draw=teaserNavy, fill=teaserNavy!65] coordinates {(-1.2,QA Acc) (2.3,Interv S@0.2) (0.1,Interv F1) (-0.1,Interv mIoU) (0.1,Notif S@0.2) (-1.3,Notif Prec) (-1.8,Notif F1) (-14.3,Notif Recall)};
\end{axis}

\begin{axis}[
  name=panelb, at={(panela.east)}, anchor=west, xshift=1.45cm,
  ybar, width=0.42\textwidth, height=0.36\textwidth, bar width=4pt,
  ymin=-18, ymax=12, ytick={-15,-10,-5,0,5,10},
  symbolic x coords={IC,ST,LT}, xtick=data,
  ylabel={Science $-$ Daily (points)},
  title={(b) Change by task},
  title style={font=\small\bfseries}, tick label style={font=\scriptsize},
  label style={font=\scriptsize}, grid=major, grid style={black!8},
  enlarge x limits=0.32,
  legend style={at={(0.5,1.22)}, anchor=south, legend columns=4,
    font=\scriptsize, draw=none, /tikz/every even column/.append style={column sep=4pt}},
]
\addplot+[draw=teaserNavy, fill=teaserNavy!70] coordinates {(IC,8.1) (ST,-7.4) (LT,-0.3)};
\addplot+[draw=teaserBlue, fill=teaserBlue!70] coordinates {(IC,-14.9) (ST,-17.8) (LT,-10.2)};
\addplot+[draw=teaserTeal, fill=teaserTeal!65] coordinates {(IC,-1.2) (ST,0.0) (LT,1.4)};
\addplot+[draw=teaserGold, fill=teaserGold!80] coordinates {(IC,-6.3) (ST,-5.9) (LT,8.7)};
\legend{Notif S@0.2,Notif Recall,Interv F1,QA Acc}
\end{axis}

\begin{axis}[
  name=panelc, at={(panela.south west)}, anchor=north west, yshift=-1.9cm,
  ybar, width=0.40\textwidth, height=0.32\textwidth, bar width=6pt,
  ymin=0, ymax=13.5, ytick={0,4,8,12},
  symbolic x coords={Propr.,Open,VidLLM,Stream.}, xtick=data,
  ylabel={Notif.\ F1 (\%)},
  title={(c) Family transfer},
  title style={font=\small\bfseries}, tick label style={font=\scriptsize},
  label style={font=\scriptsize}, grid=major, grid style={black!8},
  nodes near coords, nodes near coords style={font=\scriptsize,
    /pgf/number format/.cd, fixed, fixed zerofill, precision=1, /tikz/.cd},
  enlarge x limits=0.18,
]
\addplot+[draw=teaserBlue, fill=teaserBlue!75, every node near coord/.append style={xshift=-4pt}] coordinates {(Propr.,6.4) (Open,9.7) (VidLLM,2.4) (Stream.,4.1)};
\addplot+[draw=teaserGold, fill=teaserGold!80, every node near coord/.append style={xshift=4pt}] coordinates {(Propr.,6.8) (Open,6.0) (VidLLM,0.3) (Stream.,3.3)};
\end{axis}

\begin{axis}[
  name=paneld, at={(panelc.east)}, anchor=west, xshift=1.45cm,
  ybar, width=0.42\textwidth, height=0.32\textwidth, bar width=8pt,
  ymin=0, ymax=62, ytick={0,20,40,60},
  symbolic x coords={IC,ST,LT}, xtick=data,
  ylabel={Oracle $-$ Proact (points)},
  title={(d) Acquisition gap},
  title style={font=\small\bfseries}, tick label style={font=\scriptsize},
  label style={font=\scriptsize}, grid=major, grid style={black!8},
  nodes near coords, nodes near coords style={font=\scriptsize,
    /pgf/number format/.cd, fixed, fixed zerofill, precision=1, /tikz/.cd},
  enlarge x limits=0.3,
]
\addplot+[draw=teaserBlue, fill=teaserBlue!75] coordinates {(IC,31.3) (ST,27.9) (LT,36.2)};
\addplot+[draw=teaserGold, fill=teaserGold!80] coordinates {(IC,50.0) (ST,33.1) (LT,22.6)};
\end{axis}
\end{tikzpicture}
\vspace{2pt}

{\scriptsize Panels (c) and (d):\quad
\tikz[baseline=-0.5ex]\fill[draw=teaserBlue, fill=teaserBlue!75] (0,-0.08) rectangle (0.30,0.10); \,\dailymethod{}\quad
\tikz[baseline=-0.5ex]\fill[draw=teaserGold, fill=teaserGold!80] (0,-0.08) rectangle (0.30,0.10); \,\scimethod{}}
\caption{\label{fig:cross-track-findings}
Daily-to-Science transfer over the expanded paired results: $39$
model--task cells for notification and intervention, and $36$ for QA
(VideoLLM-Online excluded).
\textbf{(a)}~Mean change per metric: only opportunity coverage moves
substantially. \textbf{(b)}~The same change split by task: the direction
reverses between tasks for strict timing, intervention F1, and QA.
\textbf{(c)}~\NotifLevelAdj{} F1 by model family: the open-source lead in
daily life does not survive the move to the laboratory. \textbf{(d)}~Mean
gap between oracle and proactively acquired memory: acquisition costs more
in the laboratory on \vctask{} and \sttask{}, less on \lttask{}.}
\end{figure}
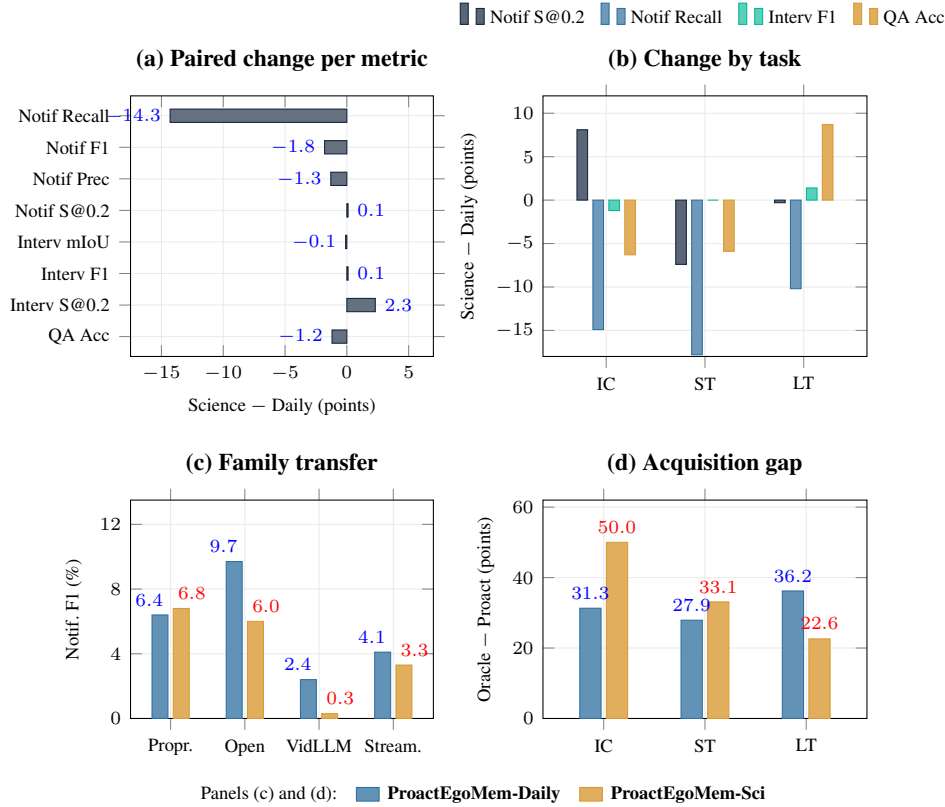

Because ten of the thirteen baselines run on both tracks under one protocol, the
two domains can be compared \emph{paired}: every claim below is a mean over the
same $30$ model--task cells on each track, so no single outlier can move it.
Figure~\ref{fig:cross-track-findings} reports the transfer.

\noindent\textbf{The clearest cross-domain effect is a systematic loss of
opportunity coverage in the laboratory.} At the \NotifLevelPlain{}, mean
recall falls from $28.5\%$ on \dailymethod{} to $15.5\%$ on \scimethod{}.
This decrease is both large and exceptionally consistent: recall is lower in
$28$ of the $30$ paired model--task cells, higher in none, and tied in only
two. Notification F1 consequently falls from $5.9\%$ to $3.9\%$, with $21$ of
$30$ cells decreasing and only four increasing. By contrast, the directions
of the intervention, conditional-timing, and QA changes are mixed, so we do
not treat their small aggregate shifts as strong evidence of a general domain
effect.

\noindent\textbf{The coverage loss spans all three tasks while largely
preserving which models are strongest.} Notification recall decreases by
$13.3$ points on \vctask{} ($38.7\!\rightarrow\!25.4\%$), $14.8$ points on
\sttask{} ($22.7\!\rightarrow\!8.0\%$), and $10.8$ points on \lttask{}
($24.0\!\rightarrow\!13.3\%$). The decline is therefore not caused by one task
or one failed model family. At the same time, the cross-track rank correlation
of recall remains very strong for \vctask{} ($\rho=0.930$), \sttask{}
($\rho=0.828$), and \lttask{} ($\rho=0.900$). Science changes the attainable
operating point much more than the relative ordering: models that cover more
daily-life opportunities usually still cover more laboratory opportunities,
but almost all of them miss substantially more of those opportunities after
the domain changes. The largest loss on \sttask{} further indicates that
recognizing domain-specific state transitions is the most acute transfer
bottleneck.

\begin{figure}[!t]
\centering
\begin{tikzpicture}
\begin{axis}[
  name=quality,
  ybar,
  width=0.54\textwidth, height=0.30\textwidth,
  bar width=7pt,
  ymin=0, ymax=24,
  ytick={0,5,10,15,20},
  ylabel={Best across models (\%)},
  symbolic x coords={Notif F1, Interv mIoU, Interv F1},
  xtick=data,
  title={Detection quality},
  title style={font=\small\bfseries},
  tick label style={font=\scriptsize},
  label style={font=\scriptsize},
  nodes near coords, nodes near coords style={font=\scriptsize},
  enlarge x limits=0.28,
  grid=major, grid style={black!8},
  legend style={at={(0.5,1.24)}, anchor=south, legend columns=2,
    font=\scriptsize, draw=none, /tikz/every even column/.append style={column sep=8pt}},
]
\addplot+[draw=teaserBlue, fill=teaserBlue!75,
  every node near coord/.append style={xshift=-5pt}]
  coordinates {(Notif F1,19.0) (Interv mIoU,17.2) (Interv F1,20.1)};
\addplot+[draw=teaserGold, fill=teaserGold!80,
  every node near coord/.append style={xshift=5pt}]
  coordinates {(Notif F1,10.8) (Interv mIoU,14.6) (Interv F1,20.6)};
\legend{Daily, Science}
\end{axis}

\begin{axis}[
  name=coverage,
  at={(quality.east)}, anchor=west, xshift=1.15cm,
  ybar,
  width=0.44\textwidth, height=0.30\textwidth,
  bar width=7pt,
  ymin=0, ymax=92,
  ytick={0,25,50,75},
  symbolic x coords={Notif Recall, Interv Recall, QA Acc},
  xtick=data,
  title={Coverage \& memory utility},
  title style={font=\small\bfseries},
  tick label style={font=\scriptsize},
  label style={font=\scriptsize},
  nodes near coords, nodes near coords style={font=\scriptsize},
  enlarge x limits=0.28,
  grid=major, grid style={black!8},
]
\addplot+[draw=teaserBlue, fill=teaserBlue!75,
  every node near coord/.append style={xshift=-5pt}]
  coordinates {(Notif Recall,79.5) (Interv Recall,37.9) (QA Acc,66.5)};
\addplot+[draw=teaserGold, fill=teaserGold!80,
  every node near coord/.append style={xshift=5pt}]
  coordinates {(Notif Recall,63.4) (Interv Recall,50.9) (QA Acc,57.2)};
\end{axis}
\end{tikzpicture}
\caption{\label{fig:track-comparison}
Daily vs.\ Science, best value across models per metric. Moving to the
laboratory lowers notification F1, intervention mIoU, notification coverage,
and memory-backed QA. The higher best intervention recall and F1 come from
two different single-model \lttask{} spikes.}
\end{figure}
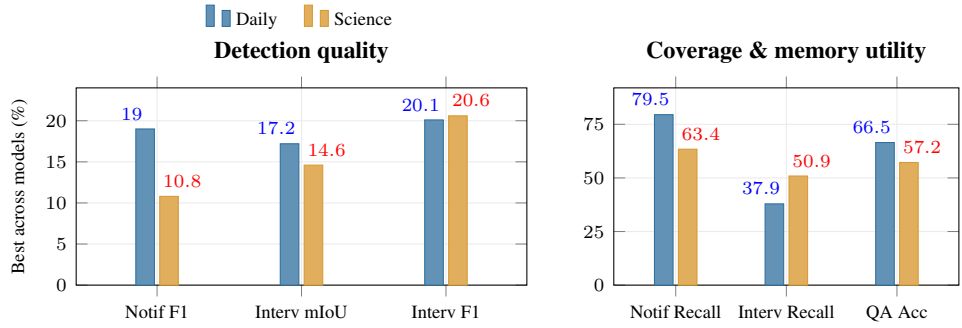

\noindent\textbf{The leaderboard view exaggerates the average effect.}
Figure~\ref{fig:track-comparison} shows the best value across models on each
track. Every joint metric falls sharply at the top: notification F1 from
$19.0\%$ to $10.8\%$, intervention mIoU from $17.2\%$ to $10.2\%$, intervention
F1 from $20.1\%$ to $15.4\%$, notification recall from $79.5\%$ to $53.1\%$, and
memory-backed QA from $66.5\%$ to $57.2\%$, while the paired means move far less
(notification F1 by $1.9$ points, intervention mIoU by $0.2$). The headline
numbers on both tracks belong to the open-source family, whose joint metrics
degrade most in the laboratory (Figure~\ref{fig:cross-track-findings}c): the
laboratory removes the ceiling that one family had set.

\noindent\textbf{Strict timing and intervention recall rise only through
selective firing.} The best Start@0.2 rises ($28.8\%\!\rightarrow\!50.0\%$) and
the best intervention recall edges up ($37.9\%\!\rightarrow\!38.8\%$), both
single-model spikes of VideoLLaMA3-7B at $0.6\%$ and $7.0\%$ precision. Silent
model--task cells rise from $2$ to $7$ of $30$ at the \NotifLevelPlain{} and from $7$
to $10$ at the \IntervLevelPlain{}, and a silent cell contributes no timing error.

\noindent\textbf{Averaged over models, the laboratory shrinks coverage and
leaves the rest of the profile in place.} Figure~\ref{fig:daily-sci-compact}a
overlays the two tracks, each polygon the mean of the $13$ models on shared
axes. At the \NotifLevelPlain{}, recall nearly halves ($32.0\%\!\rightarrow\!17.7\%$),
precision drops from $4.5\%$ to $3.2\%$, and the timing-accuracy score from
$60.7$ to $52.5$, while strict timing ($14.5\%$ against $14.6\%$) and QA accuracy
($31.9\%$ against $30.8\%$) hold. At the \IntervLevelPlain{} the polygons nearly
coincide: overlap stays at $2.3\%$ mIoU and recall slips from $10.4\%$ to
$8.8\%$, while strict timing rises from $6.2\%$ to $8.5\%$ through the fewer,
more selective firings above.

\noindent\textbf{The two tracks fail for different reasons.} \dailymethod{} is
limited by comprehension as much as by acquisition: its oracle accuracy
averages $63.1\%$ and reaches only $56.8$--$91.1\%$ at best. \scimethod{} is
limited by acquisition: its oracle mean is higher ($67.4\%$, up to $99.3\%$ on
\vctask{}) while proactively acquired accuracy is lower, widening the mean
acquisition gap from $29.2$ to $34.7$ points
(Figure~\ref{fig:cross-track-findings}d).

\noindent\textbf{Improving cross-domain proactivity therefore requires
coverage-aware domain adaptation.} Training should expose models to laboratory
objects, instruments, procedures, and hard negatives, while explicitly
supervising whether an actionable opportunity exists before optimizing its
start time or lifecycle boundary. For \sttask{}, a persistent representation
of the previous object state and the observed transition should make the
trigger depend on a change verified across frames. Finally,
coverage or no-output rates should be optimized and reported jointly with
conditional timing metrics, preventing abstention from appearing as successful
temporal transfer.

\section{Results at the Two Proactive Levels}
\label{sec:proactive-level-results}

\noindent\textbf{At the \NotifLevel{}, apparent proactivity is bought by
over-triggering.} The detection columns show how
coverage is earned: recall runs high, up to $79.5\%$ (Inkling-Small on
\vctask{}), while precision is almost uniformly in the single digits, so the
best notification F1 anywhere in the table is only $19.0\%$ (Qwen3.6-27B on
\sttask{}). Models inflate recall by firing indiscriminately and
paying for coverage with false positives, and the few
high-precision cells are the mirror image of the same miscalibration. Nor is the moment well chosen when it is
caught: Start@0.2 never exceeds $28.8\%$, and because that metric is
conditional on matched intervals it flatters the least selective models, the
strict-timing leaders (VideoLLaMA3-7B, VideoLLM-Online, Qwen3.5-9B) all show precision below $2\%$.

\Needspace{6\baselineskip}
\noindent\textbf{At the \IntervLevel{}, no model maintains the full memory
lifecycle in real time, and the over-triggering could not help either.} Maintaining a
start--maintain--stop background tool lifecycle proves far harder than firing a single notification, with overlap in the single digits in almost all model--task cells. Nine scores contain no valid
intervention at all (VideoLLM-Online on all three tasks; Qwen3.5-9B,
LLaVA-OV-1.5-8B, MiniCPM-o-4.5, and JoyAI-VL on \sttask{}; MMDuet and JoyAI-VL
on \lttask{}), and the best Start@0.2 drops to $20.0\%$. Chattiness at the \NotifLevelPlain{} no longer supports substantial event coverage either: recall falls from as high as $79.5\%$ at the
\NotifLevelPlain{} to at most $37.9\%$ (Gemma4-31B on \vctask{}), while
precision remains low once conditioned on recall, as its highest scores are degenerate (JoyAI-VL at $22.2\%$ precision on $0.7\%$ recall, MiniCPM-o-4.5 at
$20.0\%$ on $0.4\%$). Precision, recall, and lifecycle overlap collapse
together at the \IntervLevelPlain.

\noindent\textbf{Intervention demands background co-existence.} A notification
is a single shot; an intervention must survive the stream---the agent commits
to a tool call, holds it open as the scene changes, keeps perceiving in
parallel, and judges when to stop (Section~\ref{sec:modes}). The two-level gap
measures that co-existence.

\section{Qualitative Failure Cases}
\label{app:failure-cases}

Figures~\ref{fig:fc-lifecycle}--\ref{fig:fc-protocol} walk through
representative traces behind the failure modes of
Section~\ref{sec:failure-modes}. Each case gives the setting, the model's
timestamped reasoning with the tool calls it emitted and the tool's response,
the behavior the ground truth asks for, and an error analysis.

\newtcolorbox{fcsetting}[1][]{colback=black!5, colframe=black!75, boxrule=0.6pt,
  arc=0pt, left=3pt, right=3pt, top=2pt, bottom=2pt, before skip=0pt,
  after skip=3pt, fontupper=\small, fontlower=\small, #1}
\newtcolorbox{fctrace}{breakable, colback=orange!8, colframe=orange!75!black,
  boxrule=0.6pt, arc=0pt, left=3pt, right=3pt, top=2pt, bottom=2pt,
  before skip=0pt, after skip=3pt, fontupper=\small\raggedright}
\newtcolorbox{fcexpect}{breakable, colback=green!6, colframe=green!45!black,
  boxrule=0.6pt, arc=0pt, left=3pt, right=3pt, top=2pt, bottom=2pt,
  before skip=0pt, after skip=3pt, fontupper=\small}
\newtcolorbox{fcerror}{breakable, colback=blue!5, colframe=blue!55!black, boxrule=0.6pt,
  arc=0pt, left=3pt, right=3pt, top=2pt, bottom=2pt, before skip=0pt,
  after skip=7pt, fontupper=\small}
\newcommand{\fcts}[1]{\textbf{#1}\enspace}
\newcommand{\fccall}[1]{\\\hspace*{1.6em}$\hookrightarrow$\ {\footnotesize\fontfamily{lmtt}\selectfont #1}}
\newcommand{\fcbad}[1]{\textcolor{red!70!black}{#1}}

\Needspace{11\baselineskip}
\begingroup
\begin{fcsetting}
\textbf{Setting:} VideoLLaMA3-7B, \dailymethod{}, \lttask{}, \IntervLevelPlain{}; a
3:58 clip. \textbf{Situation:} the user installs a wooden floor, repeatedly
picking up and placing planks and a hammer.
\end{fcsetting}
\begin{fctrace}
\textbf{VideoLLaMA3-7B trace:}\\
\fcts{0:46}\emph{``I decided to start tracking an object's location because
the user is placing a plank on the floor, which could be a personal object.''}
\fccall{start\_location\_tracking(name="plank", location="on wooden surface")
$\rightarrow$ reminder-16}\\
\fcts{0:46--3:58}The same call recurs at almost every second the plank is in
view, and hammer tracks pile up alongside: the timeline holds parallel plank
and hammer tracks, \fcbad{every one still open when the clip ends; no stop
call is issued}.
\end{fctrace}
\begin{fcexpect}
\textbf{Expected behavior:} open one location track when an object is set
down, keep it open while the object stays put, and close it once when the
object is moved; a new track starts only for a new placement.
\end{fcexpect}
\begin{fcerror}
\textbf{Error analysis:} (1) \textbf{No lifecycle state.} The model decides
from the current frame alone and never consults the tracks it already holds,
so every glimpse of the plank opens a duplicate. (2) \textbf{No closing
decision.} Nothing in the stream prompts a stop, so the background memory
grows without bound and never yields a final location.
\end{fcerror}
\captionsetup{position=below,skip=0pt}\captionof{figure}{\label{fig:fc-lifecycle}\textbf{Lifecycle maintenance failure:
VideoLLaMA3-7B opens tracks without ever closing them.}}
\par\endgroup\vspace{8pt}

\Needspace{11\baselineskip}
\begingroup
\begin{fcsetting}
\textbf{Setting:} VideoLLaMA3-7B, \dailymethod{}, \IntervLevelPlain{}.
\textbf{Situation (top):} the same floor installation, \lttask{}; a hammer
lies on the wooden surface. \textbf{Situation (bottom):} a driving clip,
\vctask{}; the route is being recorded.
\end{fcsetting}
\begin{fctrace}
\textbf{VideoLLaMA3-7B trace (hammer):}\\
\fcts{0:10}\emph{``I decided to start tracking an object's location because
the user is working on a floor installation project and needs to keep track of
tools and materials.''}
\fccall{start\_location\_tracking(name="hammer", location="on wooden surface")
$\rightarrow$ reminder-6}\\
\fcts{0:11}\emph{``I decided to stop tracking the hammer's location because
it is no longer visible in the current frame.''}
\fccall{stop\_location\_tracking(id=reminder-6) $\rightarrow$ closed after
1.0\,s}\\
\fcts{0:12}\emph{(same reasoning)}
\fccall{stop\_location\_tracking() $\rightarrow$ \fcbad{REJECTED: already
closed at 0:11}}\\[3pt]
\textbf{VideoLLaMA3-7B trace (recording):}\\
\fcts{0:09}\emph{``I decided to stop recording because the current driving
route has already been recorded.''}
\fccall{stop\_recording() $\rightarrow$ closed after 4.0\,s}\\
\fcts{0:10--0:12}\emph{``I decided to stop recording because the current
driving route has been captured.''} (three times)
\fccall{stop\_recording() $\rightarrow$ \fcbad{REJECTED: no recording
interval is active} (three times)}
\end{fctrace}
\begin{fcexpect}
\textbf{Expected behavior:} keep the hammer's location while it is out of
view, since that is when the user needs it, and issue each stop exactly once,
against a live track or recording.
\end{fcexpect}
\begin{fcerror}
\textbf{Error analysis:} (1) \textbf{Visibility stands in for state.} The
model closes the hammer track because the hammer left the frame, discarding
the memory at the moment it becomes useful. (2) \textbf{Stale tool state.}
It repeats stop calls on a track and a recording it has already closed,
reasoning about the scene while ignoring the tool responses it receives.
\end{fcerror}
\captionsetup{position=below,skip=0pt}\captionof{figure}{\label{fig:fc-stop}\textbf{Lifecycle maintenance failure:
VideoLLaMA3-7B stops tools that are no longer live.}}
\par\endgroup\vspace{8pt}

\Needspace{11\baselineskip}
\begingroup
\begin{fcsetting}[sidebyside, lefthand width=0.66\linewidth, sidebyside gap=6pt,
  segmentation style={draw=none}]
\textbf{Setting:} Gemma4-31B, \dailymethod{}, \lttask{}, \IntervLevelPlain{}; a
4:54 cooking clip. \textbf{Situation:} the user cooks at the counter, pouring
and blending while lids and containers move between the mixer, jar, cutting
board, and table.
\tcblower
\centering\includegraphics[width=\linewidth]{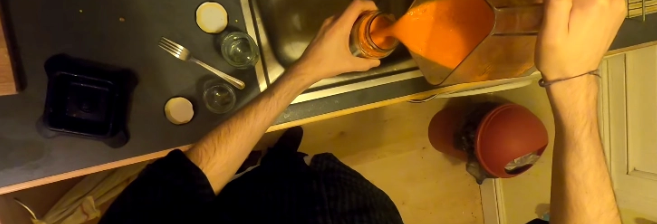}
\end{fcsetting}
\begin{fctrace}
\textbf{Gemma4-31B trace:}\\
\fcts{0:04}\emph{``I decided to start tracking the location of the open
notebook on the counter as the user is actively referring to it while
cooking\ldots''}
\fccall{start\_location\_tracking(name="open notebook", location="on the
kitchen counter next to the stove") $\rightarrow$ reminder-1}\\
\fcts{0:06}\emph{``\ldots the user has finished interacting with it and is now
focused on cooking\ldots''}
\fccall{stop\_location\_tracking(id=reminder-1) $\rightarrow$ closed after
2.0\,s}\\
\fcts{0:13}\emph{(same notebook, same reason)}
\fccall{start\_location\_tracking(name="open notebook", \ldots) $\rightarrow$
reminder-2}\\
\fcts{0:22}\emph{``\ldots the user has moved away from it\ldots''}
\fccall{stop\_location\_tracking(id=reminder-2) $\rightarrow$ closed after
9.0\,s}\\
\fcts{2:51}\emph{``I decided to start tracking the black electronic device's
location because it is sitting on the kitchen counter and the user is active in
the area\ldots''}\\
\fcts{0:00--4:54}\fcbad{$111$ tracking intervals and $222$ messages, most
intervals lasting about one second.}
\end{fctrace}
\begin{fcexpect}
\textbf{Expected behavior:} the ground truth holds $7$ location events in this
clip, each a lid or container moved between the mixer, jar, cutting board, and
table; one track per event, opened when the object is moved.
\end{fcexpect}
\begin{fcerror}
\textbf{Error analysis:} (1) \textbf{Salience triggers action.} The model
tracks whatever is prominent on the counter, a notebook or a device that is
never moved, and misses the lids and containers that are. (2)
\textbf{Interaction drives the lifecycle.} Each track closes when the user
looks away and reopens when the object returns, fragmenting one object into
many one-second intervals.
\end{fcerror}
\captionsetup{position=below,skip=0pt}\captionof{figure}{\label{fig:fc-overtrigger}\textbf{Uncalibrated interruption:
Gemma4-31B over-triggers location tracking.}}
\par\endgroup\vspace{8pt}

\Needspace{11\baselineskip}
\begingroup
\begin{fcsetting}[sidebyside, lefthand width=0.66\linewidth, sidebyside gap=6pt,
  segmentation style={draw=none}]
\textbf{Setting:} VideoLLM-Online, \dailymethod{}, \sttask{}, \IntervLevelPlain{};
a 4:56 clip. \textbf{Situation:} the user prepares food in a home kitchen,
opening the refrigerator, running the tap, and handling a cucumber.
\tcblower
\centering\includegraphics[width=\linewidth]{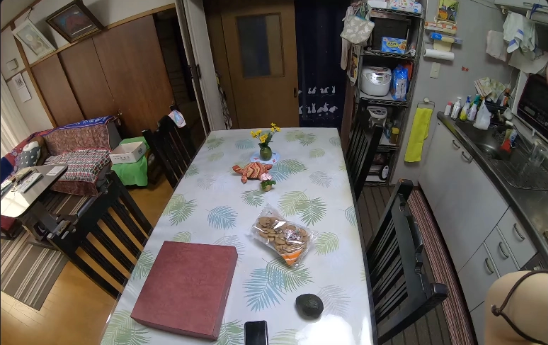}
\end{fcsetting}
\begin{fctrace}
\textbf{VideoLLM-Online trace:}\\
\fcts{0:00}\emph{``(Video Time = 0.0s) Assistant: You are in a room.''}
\fcbad{[unparsed]}\\
\fcts{0:02}\emph{``(Video Time = 2.0s) Assistant: You open the fridge.''}
\fcbad{[unparsed]}\\
\fcts{0:12}\emph{``\ldots You open the tap.''} \fcbad{[unparsed]}\quad
\fcts{0:15}\emph{``\ldots You close the tap.''} \fcbad{[unparsed]}\\
\fcts{0:00--4:56}\fcbad{$248$ messages, all unparsed; $0$ tool intervals.}\\[3pt]
\textbf{MMDuet trace (another \sttask{} clip, the user cutting potatoes):}\\
\fcts{1:00}\emph{``I decided to start tracking a state because the person is
cutting potatoes.''} \fcbad{[unparsed]}\\
\fcts{1:01}\emph{(identical sentence)} \fcbad{[unparsed]}
\end{fctrace}
\begin{fcexpect}
\textbf{Expected behavior:} answer each moment with a JSON tool call; here,
start status tracking when the refrigerator opens (the ground truth marks it
open at 0:02--0:07) and stop it when it closes.
\end{fcexpect}
\begin{fcerror}
\textbf{Error analysis:} (1) \textbf{Perception succeeds, the protocol
fails.} VideoLLM-Online sees the refrigerator open at the annotated moment and
says so, yet answers in its pre-trained narration format, so no tool ever
starts. (2) \textbf{Intent without action.} MMDuet states the right decision in
words and never emits the call, repeating it every second while nothing
executes.
\end{fcerror}
\captionsetup{position=below,skip=0pt}
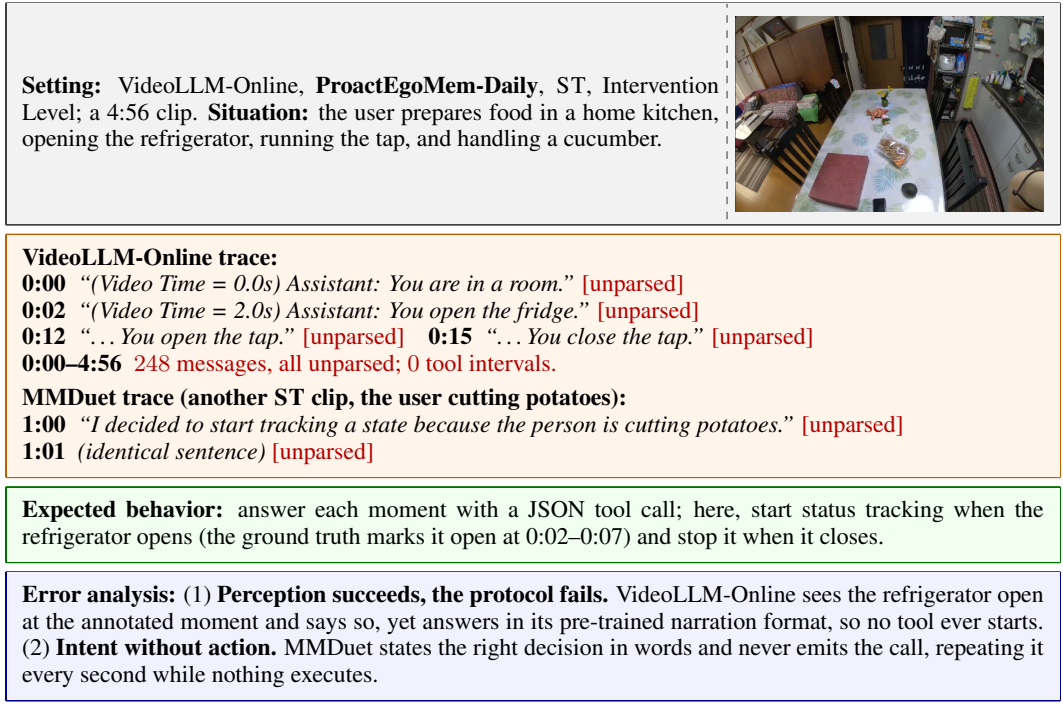
\captionof{figure}{\label{fig:fc-protocol}\textbf{Video-language models fail to follow
instructions: VideoLLM-Online narrates and MMDuet describes its decision
instead of calling the tool.}}
\par\endgroup\vspace{8pt}

\section{Results Across Ambient-Memory Tasks}
\label{sec:task-results}

\noindent\textbf{Difficulty scales with the memory horizon
(\vctask{}\,$<$\,\sttask{}\,$<$\,\lttask{}).} \vctask{} is the easiest task:
memory-backed QA accuracy reaches $36$--$66\%$, well above the $18$--$34\%$ seen
on \sttask{} and \lttask{}.
\begin{itemize}[leftmargin=1.3em, itemsep=2pt, topsep=3pt, parsep=0pt]
\item \textbf{\vctask{}} is the easiest because reading dense text off video
frames lands in the strengths of current video/visual-language models, as they
are trained on dense visual content.
\item \textbf{\sttask{}} demands temporal reasoning with parallelism: states change concurrently during multitasking for several object status and the model could not hold multiple state lifecycles open at once, interleaving their updates to know what remains active, in which order, and for how long.
\item \textbf{\lttask{}} is the hardest, as it requires tracking the dynamics of
an individual object over the longest horizon. The model must overwrite the object locations at every move while still remembering the original location, the last location, and whether the object has been moved elsewhere or stays in the same place.
\end{itemize}

\section{\scimethod: Full Main Results}
\label{app:sci-main-results}

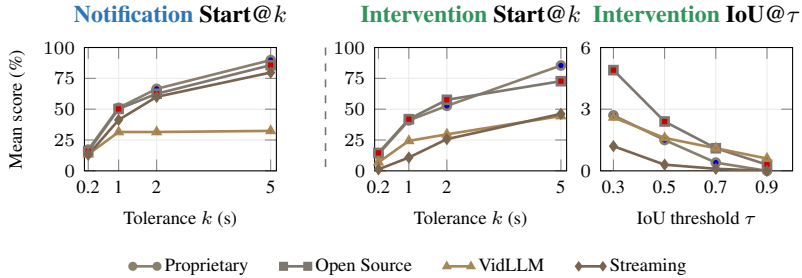
\begin{figure}[!t]
\centering
\providecolor{thresholdProp}{HTML}{8A7F6D}
\providecolor{thresholdOpen}{HTML}{7D7770}
\providecolor{thresholdVid}{HTML}{A98755}
\providecolor{thresholdStream}{HTML}{7A634F}
\begin{tikzpicture}
\begin{axis}[
  name=scinotifstart,
  width=0.295\textwidth,
  height=0.23\textwidth,
  xmin=0.1, xmax=5.15,
  ymin=0, ymax=100,
  xtick={0.2,1,2,5},
  xticklabels={0.2,1,2,5},
  ytick={0,25,50,75,100},
  grid=major,
  grid style={black!8},
  xlabel={Tolerance $k$ (s)},
  ylabel={Mean score (\%)},
  title={\Notif Start@$k$},
  title style={font=\small\bfseries},
  tick label style={font=\scriptsize},
  label style={font=\scriptsize},
  every axis plot/.append style={line width=1pt, mark options={solid, scale=0.75}}
]
\addplot+[color=thresholdProp, mark=*] coordinates {(0.2,17.2) (1,51.1) (2,66.5) (5,89.8)};
\addplot+[color=thresholdOpen, mark=square*] coordinates {(0.2,15.8) (1,50.2) (2,62.5) (5,85.7)};
\addplot+[color=thresholdVid, mark=triangle*] coordinates {(0.2,13.4) (1,31.5) (2,31.5) (5,32.4)};
\addplot+[color=thresholdStream, mark=diamond*] coordinates {(0.2,13.0) (1,41.3) (2,60.0) (5,79.7)};
\end{axis}

\begin{axis}[
  name=sciintervstart,
  at={(scinotifstart.east)},
  anchor=west,
  xshift=1.30cm,
  width=0.295\textwidth,
  height=0.23\textwidth,
  xmin=0.1, xmax=5.15,
  ymin=0, ymax=100,
  xtick={0.2,1,2,5},
  xticklabels={0.2,1,2,5},
  ytick={0,25,50,75,100},
  grid=major,
  grid style={black!8},
  xlabel={Tolerance $k$ (s)},
  title={\Interv Start@$k$},
  title style={font=\small\bfseries},
  tick label style={font=\scriptsize},
  label style={font=\scriptsize},
  every axis plot/.append style={line width=1pt, mark options={solid, scale=0.75}}
]
\addplot+[color=thresholdProp, mark=*] coordinates {(0.2,13.6) (1,40.9) (2,52.7) (5,85.4)};
\addplot+[color=thresholdOpen, mark=square*] coordinates {(0.2,14.5) (1,41.9) (2,57.6) (5,72.7)};
\addplot+[color=thresholdVid, mark=triangle*] coordinates {(0.2,6.9) (1,24.3) (2,29.6) (5,44.4)};
\addplot+[color=thresholdStream, mark=diamond*] coordinates {(0.2,1.2) (1,10.8) (2,25.7) (5,46.2)};
\end{axis}

\begin{axis}[
  name=sciinterviou,
  at={(sciintervstart.east)},
  anchor=west,
  xshift=0.45cm,
  width=0.295\textwidth,
  height=0.23\textwidth,
  xmin=0.25, xmax=1.0,
  ymin=0, ymax=6,
  xtick={0.3,0.5,0.7,0.9},
  xticklabels={0.3,0.5,0.7,0.9},
  ytick={0,3,6},
  grid=major,
  grid style={black!8},
  xlabel={IoU threshold $\tau$},
  title={\Interv IoU@$\tau$},
  title style={font=\small\bfseries},
  tick label style={font=\scriptsize},
  label style={font=\scriptsize},
  every axis plot/.append style={line width=1pt, mark options={solid, scale=0.75}}
]
\addplot+[color=thresholdProp, mark=*] coordinates {(0.3,2.7) (0.5,1.5) (0.7,0.4) (0.9,0.0)};
\addplot+[color=thresholdOpen, mark=square*] coordinates {(0.3,4.9) (0.5,2.4) (0.7,1.1) (0.9,0.3)};
\addplot+[color=thresholdVid, mark=triangle*] coordinates {(0.3,2.6) (0.5,1.6) (0.7,1.1) (0.9,0.6)};
\addplot+[color=thresholdStream, mark=diamond*] coordinates {(0.3,1.2) (0.5,0.3) (0.7,0.1) (0.9,0.0)};
\end{axis}
\draw[dashed, line width=0.6pt, black!55]
  ($(scinotifstart.north east)!0.5!(sciintervstart.north west)$)
  --
  ($(scinotifstart.south east)!0.5!(sciintervstart.south west)$);
\end{tikzpicture}
\vspace{2pt}

{\scriptsize
\begin{tabular}{@{}cccc@{}}
\tikz[baseline=-0.5ex]\draw[color=thresholdProp, line width=1pt, mark=*, mark size=1.2pt] plot coordinates {(0,0) (0.32,0)}; Proprietary &
\tikz[baseline=-0.5ex]\draw[color=thresholdOpen, line width=1pt, mark=square*, mark size=1.2pt] plot coordinates {(0,0) (0.32,0)}; Open Source &
\tikz[baseline=-0.5ex]\draw[color=thresholdVid, line width=1pt, mark=triangle*, mark size=1.2pt] plot coordinates {(0,0) (0.32,0)}; VidLLM &
\tikz[baseline=-0.5ex]\draw[color=thresholdStream, line width=1pt, mark=diamond*, mark size=1.2pt] plot coordinates {(0,0) (0.32,0)}; Streaming
\end{tabular}}
\caption{\label{fig:sci-threshold-curves}
\scimethod{} group-level mean Start@$k$ and intervention IoU@$\tau$ over
tasks and models as each criterion is relaxed. Note the collapsed scale of the
rightmost panel: the best group mean stays below 6\%.}
\end{figure}

Table~\ref{tab:sci-merged-main-results} reports the full notification- and
intervention-level results on \scimethod{} under the metrics and layout of
Table~\ref{tab:merged-main-results}. Ten of the thirteen baselines have a
completed streaming run at those two levels; Qwen3.8-27B, Inkling-Small, and
JoyAI-VL are evaluated on the QA memory study only
(Appendix~\ref{app:sci-memory}), which covers all thirteen. Figure~\ref{fig:sci-threshold-curves} relaxes the timing and overlap
criteria, Figure~\ref{fig:radars-sci} compares the per-axis tradeoffs, and
Figure~\ref{fig:sci-qa-results} evaluates whether the acquired memory supports
downstream questions.

\providecommand{\scigrouprule}{%
  \arrayrulecolor{black!30}\cmidrule[0.02em](lr){2-17}\arrayrulecolor{black}}

\begin{table}[!t]
\centering
\caption{\label{tab:sci-merged-main-results}%
\textbf{\scimethod{} main results} for both levels, in the layout of Table~\ref{tab:merged-main-results}.
\textbullet~\textbf{Start@0.2}, \textbf{mIoU} and \textbf{Acc} are defined as in Table~\ref{tab:merged-main-results}.
\textbullet~Error columns are seconds, all others percentages; \textbf{bold}: best in a task.}
\vspace{2pt}
\setlength{\tabcolsep}{4pt}
\resizebox{\textwidth}{!}{%
\small
\begin{tabular}{@{}lllrrrrrr!{\vrule width 1.1pt}rrrrrrrr@{}}
\toprule
\multirow{3}{*}{Task} & \multirow{3}{*}{Group} & \multirow{3}{*}{Model}
& \multicolumn{6}{c!{\vrule width 1.1pt}}{\textbf{\NotifLevel}}
& \multicolumn{8}{c}{\textbf{\IntervLevel}} \\
\cmidrule(lr){4-9}\cmidrule(l){10-17}
& & & \multicolumn{1}{c}{Timing} & \multicolumn{2}{c}{Abs Error} & \multicolumn{3}{c!{\vrule width 1.1pt}}{Detection}
& \multicolumn{1}{c}{Overlap} & \multicolumn{1}{c}{Timing} & \multicolumn{2}{c}{Abs Error} & \multicolumn{3}{c}{Detection} & \multicolumn{1}{c}{QA} \\
\cmidrule(lr){4-4}\cmidrule(lr){5-6}\cmidrule(lr){7-9}\cmidrule(lr){10-10}\cmidrule(lr){11-11}\cmidrule(lr){12-13}\cmidrule(lr){14-16}\cmidrule(l){17-17}
& & & S@0.2$\uparrow$ & $|\Delta_s|{\downarrow}$ & $\sigma{\downarrow}$ & Prec$\uparrow$ & Recall$\uparrow$ & F1$\uparrow$
& mIoU$\uparrow$ & S@0.2$\uparrow$ & $|\Delta_s|{\downarrow}$ & $\sigma{\downarrow}$ & Prec$\uparrow$ & Recall$\uparrow$ & F1$\uparrow$ & Acc$\uparrow$ \\
\midrule
\multirow{13}{*}{IC}
& \multirow{2}{*}{Proprietary} & Gemini 2.5 Flash & \rcbar{31.6} & 1.23 & 1.92 & \rcbar{2.2} & \rcbar{43.4} & 4.2 & \rcbar{2.7} & \rcbar{36.4} & 1.39 & 2.11 & \rcbar{2.5} & \rcbar{6.3} & 3.6 & \rcbar{41.8} \\
& & GPT-5.4-mini & \rcbar{30.1} & 1.28 & 1.94 & \rcbar{1.6} & \rcbar{47.4} & 3.2 & \rcbar{1.9} & \rcbar{20.0} & 3.41 & 3.56 & \rcbar{4.6} & \rcbar{5.7} & 5.1 & \rcbar{46.5} \\
\scigrouprule
& \multirow{4}{*}{Open Source} & Gemma4-31B & \rcbar{21.4} & 1.36 & 1.71 & \rcbar{4.0} & \rcbar{48.0} & 7.4 & \rcbarb{5.4} & \rcbar{13.6} & 2.43 & 2.75 & \rcbar{2.0} & \rcbarb{25.1} & 3.8 & \rcbarb{57.2} \\
& & Qwen3.8-27B & \rcbar{22.8} & 1.39 & \textbf{1.63} & \rcbar{2.4} & \rcbar{52.6} & 4.7 & \rcbar{3.7} & \rcbar{46.2} & \textbf{0.93} & 1.49 & \rcbar{5.9} & \rcbar{7.4} & \textbf{6.6} & \rcbar{55.9} \\
& & Qwen3.6-27B & \rcbar{23.7} & 1.44 & 1.84 & \rcbarb{4.9} & \rcbar{53.1} & \textbf{8.9} & \rcbar{2.9} & \rcbar{12.5} & 2.65 & 2.59 & \rcbar{1.7} & \rcbar{18.3} & 3.2 & \rcbar{54.5} \\
& & Qwen3.5-9B & \rcbar{28.1} & 1.27 & 1.89 & \rcbar{1.6} & \rcbar{32.6} & 3.1 & \rcbar{0.0} & \rcbar{0.0} & 0.00 & 0.00 & \rcbar{0.0} & \rcbar{0.0} & 0.0 & \rcbar{46.5} \\
\scigrouprule
& \multirow{3}{*}{VidLLM} & VideoLLaMA3-7B & \rcbarb{50.0} & \textbf{0.93} & 1.92 & \rcbar{0.6} & \rcbar{20.6} & 1.1 & \rcbar{2.0} & \rcbarb{50.0} & 1.53 & 2.55 & \rcbar{0.5} & \rcbar{5.7} & 1.0 & \rcbar{40.5} \\
& & LLaVA-OV-1.5-8B & \rcbar{0.0} & 0.00 & 0.00 & \rcbar{0.0} & \rcbar{0.0} & 0.0 & \rcbar{0.0} & \rcbar{0.0} & 4.15 & 1.73 & \rcbar{7.1} & \rcbar{1.1} & 2.0 & \rcbar{34.4} \\
& & MiniCPM-o-4.5 & \rcbar{0.0} & 0.00 & 0.00 & \rcbar{0.0} & \rcbar{0.0} & 0.0 & \rcbar{0.0} & \rcbar{0.0} & 0.00 & 0.00 & \rcbar{0.0} & \rcbar{0.0} & 0.0 & \rcbar{32.4} \\
\scigrouprule
& \multirow{4}{*}{Streaming} & MMDuet & \rcbar{10.0} & 2.50 & 2.04 & \rcbar{1.4} & \rcbar{5.7} & 2.3 & \rcbar{0.3} & \rcbar{0.0} & 3.06 & \textbf{0.87} & \rcbarb{8.3} & \rcbar{1.7} & 2.8 & \rcbar{39.1} \\
& & VideoLLM-Online & \rcbarb{50.0} & 2.06 & 3.34 & \rcbar{0.7} & \rcbar{3.4} & 1.1 & \rcbar{0.0} & \rcbar{0.0} & 0.00 & 0.00 & \rcbar{0.0} & \rcbar{0.0} & 0.0 & \rcbar{0.0} \\
& & Inkling-Small & \rcbar{16.2} & 1.56 & 1.95 & \rcbar{2.5} & \rcbarb{63.4} & 4.8 & \rcbar{1.4} & \rcbar{0.0} & 1.98 & 1.52 & \rcbar{1.3} & \rcbar{7.4} & 2.3 & \rcbar{42.8} \\
& & JoyAI-VL & \rcbar{11.1} & 1.64 & 2.32 & \rcbar{1.3} & \rcbar{5.1} & 2.1 & \rcbar{0.0} & \rcbar{0.0} & 0.00 & 0.00 & \rcbar{0.0} & \rcbar{0.0} & 0.0 & \rcbar{52.8} \\
\midrule[\heavyrulewidth]
\multirow{13}{*}{ST}
& \multirow{2}{*}{Proprietary} & Gemini 2.5 Flash & \rcbar{9.5} & 2.94 & 2.65 & \rcbar{6.3} & \rcbar{16.5} & 9.1 & \rcbar{3.7} & \rcbar{9.3} & 2.28 & 1.97 & \rcbar{13.2} & \rcbar{13.0} & 13.1 & \rcbar{20.2} \\
& & GPT-5.4-mini & \rcbar{12.8} & 2.06 & 2.15 & \rcbar{5.5} & \rcbarb{24.4} & 9.0 & \rcbar{1.2} & \rcbar{10.0} & 1.40 & 1.32 & \rcbar{7.9} & \rcbar{3.6} & 5.0 & \rcbarb{26.0} \\
\scigrouprule
& \multirow{4}{*}{Open Source} & Gemma4-31B & \rcbar{6.3} & 3.34 & 2.63 & \rcbar{5.7} & \rcbar{15.3} & 8.3 & \rcbarb{10.1} & \rcbar{12.3} & 1.81 & 2.04 & \rcbar{9.9} & \rcbarb{30.4} & \textbf{14.9} & \rcbar{25.0} \\
& & Qwen3.8-27B & \rcbar{9.1} & 2.61 & 2.28 & \rcbar{1.4} & \rcbar{4.0} & 2.1 & \rcbar{5.0} & \rcbar{9.3} & 2.12 & 2.12 & \rcbar{10.2} & \rcbar{11.7} & 10.9 & \rcbar{19.2} \\
& & Qwen3.6-27B & \rcbar{2.0} & 5.29 & 2.70 & \rcbarb{25.9} & \rcbar{6.1} & \textbf{9.9} & \rcbar{3.6} & \rcbar{12.7} & 1.39 & 1.43 & \rcbar{6.1} & \rcbar{12.3} & 8.2 & \rcbar{17.3} \\
& & Qwen3.5-9B & \rcbar{10.7} & 2.62 & 2.71 & \rcbar{1.0} & \rcbar{10.1} & 1.7 & \rcbar{0.1} & \rcbarb{50.0} & \textbf{0.33} & \textbf{0.15} & \rcbarb{28.6} & \rcbar{0.2} & 0.5 & \rcbar{18.3} \\
\scigrouprule
& \multirow{3}{*}{VidLLM} & VideoLLaMA3-7B & \rcbarb{30.0} & \textbf{0.28} & \textbf{0.15} & \rcbar{0.1} & \rcbar{1.2} & 0.2 & \rcbar{4.0} & \rcbar{8.6} & 2.56 & 2.26 & \rcbar{10.7} & \rcbar{15.4} & 12.6 & \rcbar{22.1} \\
& & LLaVA-OV-1.5-8B & \rcbar{0.0} & 0.00 & 0.00 & \rcbar{0.0} & \rcbar{0.0} & 0.0 & \rcbar{0.0} & \rcbar{0.0} & 0.00 & 0.00 & \rcbar{0.0} & \rcbar{0.0} & 0.0 & \rcbar{18.3} \\
& & MiniCPM-o-4.5 & \rcbar{0.0} & 0.00 & 0.00 & \rcbar{0.0} & \rcbar{0.0} & 0.0 & \rcbar{0.0} & \rcbar{0.0} & 0.00 & 0.00 & \rcbar{0.0} & \rcbar{0.0} & 0.0 & \rcbar{16.3} \\
\scigrouprule
& \multirow{4}{*}{Streaming} & MMDuet & \rcbar{0.0} & 0.00 & 0.00 & \rcbar{0.0} & \rcbar{0.0} & 0.0 & \rcbar{0.2} & \rcbar{0.0} & 1.78 & 1.37 & \rcbar{13.5} & \rcbar{1.2} & 2.2 & \rcbar{19.2} \\
& & VideoLLM-Online & \rcbar{6.1} & 2.35 & 2.33 & \rcbar{3.5} & \rcbar{5.9} & 4.4 & \rcbar{0.0} & \rcbar{0.0} & 0.00 & 0.00 & \rcbar{0.0} & \rcbar{0.0} & 0.0 & \rcbar{0.0} \\
& & Inkling-Small & \rcbar{8.0} & 2.16 & 2.34 & \rcbar{3.9} & \rcbar{15.0} & 6.2 & \rcbar{3.6} & \rcbar{10.3} & 1.93 & 1.83 & \rcbar{5.6} & \rcbar{16.4} & 8.4 & \rcbar{19.2} \\
& & JoyAI-VL & \rcbar{10.0} & 1.84 & 1.78 & \rcbar{4.6} & \rcbar{2.4} & 3.2 & \rcbar{0.0} & \rcbar{0.0} & 0.00 & 0.00 & \rcbar{0.0} & \rcbar{0.0} & 0.0 & \rcbar{15.4} \\
\midrule[\heavyrulewidth]
\multirow{13}{*}{LT}
& \multirow{2}{*}{Proprietary} & Gemini 2.5 Flash & \rcbar{8.5} & 2.03 & 1.94 & \rcbar{6.3} & \rcbar{25.3} & 10.1 & \rcbar{1.1} & \rcbar{5.9} & 2.96 & 2.75 & \rcbar{6.4} & \rcbar{9.2} & 7.5 & \rcbar{26.1} \\
& & GPT-5.4-mini & \rcbar{10.9} & 1.91 & 2.19 & \rcbar{2.7} & \rcbar{36.9} & 5.0 & \rcbar{0.9} & \rcbar{0.0} & 3.55 & 2.68 & \rcbar{8.1} & \rcbar{3.2} & 4.6 & \rcbar{35.0} \\
\scigrouprule
& \multirow{4}{*}{Open Source} & Gemma4-31B & \rcbar{9.1} & 2.41 & 2.53 & \rcbarb{9.8} & \rcbar{11.9} & \textbf{10.8} & \rcbar{2.7} & \rcbar{6.4} & 2.14 & 2.28 & \rcbar{2.0} & \rcbar{12.7} & 3.4 & \rcbarb{45.2} \\
& & Qwen3.8-27B & \rcbar{20.7} & 1.69 & 2.31 & \rcbar{3.2} & \rcbar{31.3} & 5.8 & \rcbar{9.0} & \rcbarb{6.6} & 2.79 & 2.57 & \rcbarb{14.3} & \rcbar{36.7} & \textbf{20.6} & \rcbar{40.1} \\
& & Qwen3.6-27B & \rcbar{9.1} & 1.59 & 1.55 & \rcbar{9.4} & \rcbar{5.9} & 7.3 & \rcbar{2.2} & \rcbar{4.5} & 3.09 & 2.82 & \rcbar{9.1} & \rcbar{11.9} & 10.3 & \rcbar{40.8} \\
& & Qwen3.5-9B & \rcbar{27.1} & 1.01 & 1.68 & \rcbar{1.3} & \rcbar{22.9} & 2.5 & \rcbar{0.0} & \rcbar{0.0} & 0.00 & 0.00 & \rcbar{0.0} & \rcbar{0.0} & 0.0 & \rcbar{35.0} \\
\scigrouprule
& \multirow{3}{*}{VidLLM} & VideoLLaMA3-7B & \rcbarb{40.3} & \textbf{0.33} & \textbf{0.55} & \rcbar{0.9} & \rcbar{18.1} & 1.7 & \rcbarb{14.6} & \rcbar{3.2} & 3.71 & 2.73 & \rcbar{12.4} & \rcbarb{50.9} & 20.0 & \rcbar{33.1} \\
& & LLaVA-OV-1.5-8B & \rcbar{0.0} & 0.00 & 0.00 & \rcbar{0.0} & \rcbar{0.0} & 0.0 & \rcbar{0.2} & \rcbar{0.0} & \textbf{0.96} & 0.00 & \rcbar{12.5} & \rcbar{0.3} & 0.5 & \rcbar{31.8} \\
& & MiniCPM-o-4.5 & \rcbar{0.0} & 0.00 & 0.00 & \rcbar{0.0} & \rcbar{0.0} & 0.0 & \rcbar{0.0} & \rcbar{0.0} & 0.00 & 0.00 & \rcbar{0.0} & \rcbar{0.0} & 0.0 & \rcbar{27.4} \\
\scigrouprule
& \multirow{4}{*}{Streaming} & MMDuet & \rcbar{9.4} & 2.92 & 2.61 & \rcbar{3.7} & \rcbar{8.6} & 5.2 & \rcbar{0.8} & \rcbar{0.0} & 3.39 & \textbf{1.58} & \rcbar{7.7} & \rcbar{1.9} & 3.0 & \rcbar{34.4} \\
& & VideoLLM-Online & \rcbar{16.7} & 1.81 & 2.11 & \rcbar{1.1} & \rcbar{3.2} & 1.7 & \rcbar{0.0} & \rcbar{0.0} & 0.00 & 0.00 & \rcbar{0.0} & \rcbar{0.0} & 0.0 & \rcbar{0.0} \\
& & Inkling-Small & \rcbar{9.5} & 2.41 & 2.44 & \rcbar{3.2} & \rcbarb{42.6} & 5.9 & \rcbar{5.1} & \rcbar{4.1} & 2.80 & 2.61 & \rcbar{3.4} & \rcbar{32.6} & 6.1 & \rcbar{33.8} \\
& & JoyAI-VL & \rcbar{8.7} & 2.35 & 1.89 & \rcbar{2.1} & \rcbar{6.2} & 3.2 & \rcbar{0.0} & \rcbar{0.0} & 0.00 & 0.00 & \rcbar{0.0} & \rcbar{0.0} & 0.0 & \rcbar{36.9} \\
\bottomrule
\end{tabular}%
}
\end{table}

\noindent\textbf{Conclusion (Table~\ref{tab:sci-merged-main-results}).}
Transferring from daily life to the laboratory amplifies every joint failure
mode: the best notification F1 drops to $10.8\%$ (Gemma4-31B on \lttask{}) and
the best intervention overlap to $10.2\%$ mIoU (Gemma4-31B on \sttask{}), while
$7$ of $30$ model--task cells emit no valid notification and $10$ of $30$ no
valid intervention at all.

\noindent\textbf{Proprietary access does not rescue the laboratory track.}
The proprietary pair buys coverage but not competence. Averaged over models and
tasks, it leads notification recall ($32.3\%$ against $22.9\%$ for the
open-source group) yet trails on notification precision ($4.1$ against $7.1$),
intervention overlap ($1.9$ against $2.9$ mIoU), intervention recall ($7.2$
against $12.1$), and memory-backed QA ($32.6$ against $37.8$), with the two
families level on notification F1 ($6.8$ against $6.7$). No proprietary model
holds the best value of any joint timing-and-semantics metric on any task.

\noindent\textbf{Strict timing is task-dependent.}
Mean notification Start@0.2 over all models is $24.5\%$ on \vctask{} in the
laboratory against $14.6\%$ on the same task in daily life, but falls to
$7.7\%$ on \sttask{} and $13.1\%$ on \lttask{} (daily: $15.4\%$ and $13.4\%$).
Laboratory \vctask{} moments---a label written, a reading taken---are sharper
and easier to pin down than their everyday counterparts, whereas the diffuse
onsets of laboratory status changes make \sttask{} timing markedly worse. The
joint metrics still fall on all three tasks, so the laboratory is harder
overall despite this one easier axis.

\noindent\textbf{Models land in the right neighborhood but not at the right
instant.} Figure~\ref{fig:sci-threshold-curves} sweeps both criteria.
Conditional on a semantic match, notifications arrive within $5$\,s at group
means of $70.7$--$89.8\%$ for every family except the video-language one, yet
the same families sit at $13.4$--$17.2\%$ within $200$\,ms. The steepest drop
falls between $0.2$ and $1$\,s, exactly the range separating an assistant that
feels responsive from one that feels late, so the deficit lies in calibration.

\noindent\textbf{Lifecycle overlap collapses at every threshold.} Relaxing the overlap criterion reveals no hidden competence: the
best group mean is $3.8\%$ of opportunities at IoU$\,\ge\!0.3$ (open source),
falling to $1.8\%$ at $0.5$, $0.8\%$ at $0.7$, and $0.2\%$ at $0.9$. Per model,
the single strongest cell covers $13.8\%$ of opportunities at
IoU$\,\ge\!0.3$ and $0.4\%$ at $0.9$. Models fail to bracket the opportunity at all.

\noindent\textbf{Group means confirm the weakness is field-wide.} The strongest family on any given axis still averages only
$6.8\%$ notification F1 (proprietary), $2.9\%$ intervention mIoU (open source),
and $6.4\%$ intervention F1 (proprietary); the video-language and streaming
families are lower still. No family exceeds single digits on any joint
timing-and-semantics metric, and the overall means---$3.9\%$ notification F1,
$1.8\%$ intervention mIoU---leave no candidate for a usable laboratory
assistant.
\FloatBarrier

\subsection{Family Tradeoffs on the Science Track}
\label{app:sci-radars}

\begin{figure}[!t]
\centering
\includegraphics{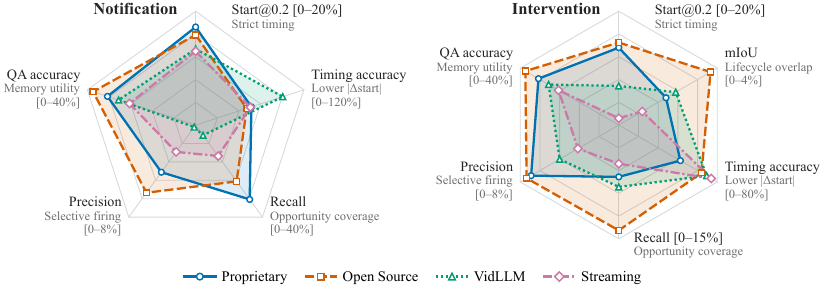}
\caption{\label{fig:radars-sci}
Notification- and intervention-level radar summaries for \scimethod{},
task-averaged per model family. Axes and normalization follow
Figure~\ref{fig:radars}.}
\end{figure}

Figure~\ref{fig:radars-sci} gives the family-level, task-averaged view of the
Science track over the same six axes used in Figure~\ref{fig:radars}.

\noindent\textbf{No model family is a rounded proactive assistant on the
Science track either.} In Figure~\ref{fig:radars-sci}, open-source models lead
memory-backed QA ($37.9\%$), intervention overlap, and intervention recall;
proprietary models lead notification recall ($32.3\%$) and strict timing; and
video-language models spike only on notification timing accuracy. The best notification
F1 reaches only $10.8\%$ (Gemma4-31B on \lttask{}) and the best intervention
overlap only $10.2\%$ mIoU (Gemma4-31B on \sttask{}), both below their
daily-life counterparts, and the per-metric leaders remain
scattered---VideoLLaMA3-7B owns strict timing and intervention F1, Gemma4-31B
overlap and QA, Qwen3.6-27B notification precision, GPT-5.4-mini notification
recall.

\noindent\textbf{The intervention level stays far harder than the notification
level.} Intervention overlap remains in the single digits for every
model--task cell, and a third of the grid produces nothing at all: MiniCPM-o-4.5
and VideoLLM-Online emit no valid intervention on any of the three tasks,
LLaVA-OV-1.5-8B on two, and Qwen3.5-9B---one of the stronger notification-level
timers---collapses to zero on \vctask{} and \lttask{}. Detection recall falls
from up to $53.1\%$ at the \NotifLevelPlain{} to at most $38.8\%$ at the
\IntervLevelPlain{}.

\noindent\textbf{Apparent proactivity is again bought by over-triggering.}
Notification precision stays at or below $9.8\%$ in $29$ of $30$ cells; the lone
exception, Qwen3.6-27B at $25.9\%$ on \sttask{}, comes with only $6.1\%$ recall.
Recall runs far higher (up to $53.1\%$, Qwen3.6-27B on \vctask{})---models
inflate coverage by firing indiscriminately.

\noindent\textbf{The smallest video-language model tops every timing axis.}
The $7$B VideoLLaMA3-7B dominates strict
timing on all three tasks (Start@0.2 up to $50.0\%$) and posts the best
intervention recall and F1 ($38.8\%$ and $15.4\%$), while the $31$B Gemma4-31B
leads overlap and QA and the proprietary pair leads none of them; the
purpose-built streaming systems again largely collapse at the intervention
level. Proactive competence tracks neither scale, nor API access, nor
streaming-specific design here.

\subsection{Effects of Proactive Memory on the Science Track}
\label{app:sci-memory}

Proactive acquisition matters only if the stored memory supports later
questions, so every \scimethod{} question is answered under the three memory
conditions of Section~\ref{sec:proactive-memory-effect}: \emph{No Memory},
where the model stores nothing and answers from frames sampled uniformly across
the whole clip; \emph{Proact Memory}, where the frames available at answering
time are exactly those the model's own tool calls chose to keep; and
\emph{Oracle Memory}, which substitutes the ground-truth events and bounds
what perfect proactivity could deliver. VideoLLM-Online returns no valid answer
under any condition and is excluded from the counts below, leaving twelve
models per task.

\begin{figure}[!t]
\centering
\definecolor{qaPropGroup}{HTML}{F8F2E5}
\definecolor{qaOpenGroup}{HTML}{E9E3DC}
\definecolor{qaVidGroup}{HTML}{DDCDB0}
\definecolor{qaStreamGroup}{HTML}{CBBAA8}

\begin{tikzpicture}[
  font=\scriptsize,
  qaNo/.style={draw=black, line width=1.5pt, line cap=round},
  qaMem/.style={draw=orange!45!black, line width=1.5pt, line cap=round},
  qaOracle/.style={draw=black!45, line width=1.5pt, line cap=round},
  qaNoDot/.style={circle, draw=black, fill=white, line width=0.5pt,
    inner sep=0.9pt},
  qaMemDot/.style={circle, draw=orange!45!black, fill=white, line width=0.5pt,
    inner sep=0.9pt},
  qaOracleDot/.style={circle, draw=black!45, fill=white, line width=0.5pt,
    inner sep=0.9pt}
]
\def\RowH{0.25}
\def\LabelW{1.45}     %
\def\PlotW{3.80}      %
\def\PanelStep{4.10}  %
\pgfmathsetmacro{\Bottom}{-12*\RowH-0.5*\RowH}
\pgfmathsetmacro{\Top}{0.5*\RowH}
\pgfmathsetmacro{\XU}{0.01*\PlotW}  %

\providecommand{\QARow}{}\renewcommand{\QARow}[4]{%
  \pgfmathsetmacro{\topnm}{max(#2,#3)}
  \draw[qaNo] (0,#1) -- (#2,#1);
  \draw[qaMem] (#2,#1) -- (#3,#1);
  \draw[qaOracle] (\topnm,#1) -- (#4,#1);
  \node[qaNoDot] at (#2,#1) {};
  \node[qaMemDot] at (#3,#1) {};
  \node[qaOracleDot] at (#4,#1) {};
}

\providecommand{\QAPanel}{}\renewcommand{\QAPanel}[3]{%
  \begin{scope}[shift={({\LabelW+#1*\PanelStep},0)}]
    \foreach \ra/\rb/\col in {0/1/qaPropGroup,2/5/qaOpenGroup,6/8/qaVidGroup,9/12/qaStreamGroup} {
      \fill[\col] (0,{-\ra*\RowH+0.5*\RowH}) rectangle (\PlotW,{-\rb*\RowH-0.5*\RowH});
    }
    \node[font=\bfseries\footnotesize, anchor=south] at (0.5*\PlotW,\Top) {#2};
    \begin{scope}[x=\XU cm]
      \foreach \xt/\anc in {0/north west,50/north,100/north east} {
        \draw[black!25, densely dashed] (\xt,\Top) -- (\xt,\Bottom);
        \node[anchor=\anc, text=black!70, inner xsep=0pt, inner ysep=1.5pt] at (\xt,\Bottom) {\xt};
      }
      \draw[black!60] (0,\Bottom) -- (100,\Bottom);
      \draw[black!60] (0,\Bottom) -- (0,\Top);
      \foreach \i/\nomem/\mem/\oracle in {#3} {
        \QARow{-\i*\RowH}{\nomem}{\mem}{\oracle}
      }
    \end{scope}
    \node[anchor=north, text=black!70, inner sep=0pt] at (0.5*\PlotW,{\Bottom-0.27})
      {Accuracy (\%)};
  \end{scope}
}

\QAPanel{0}{\vctask{}}{
  0/46.8/41.8/97.7,
  1/47.5/46.5/95.0,
  2/54.2/57.2/99.0,
  3/54.5/55.9/99.0,
  4/54.5/54.5/99.3,
  5/42.8/46.5/97.3,
  6/35.5/40.5/92.3,
  7/34.8/34.4/94.6,
  8/32.8/32.4/97.0,
  9/37.5/39.1/80.3,
  10/0.0/0.0/0.0,
  11/49.2/42.8/97.7,
  12/50.8/52.8/95.7}

\QAPanel{1}{\sttask{}}{
  0/22.1/20.2/58.7,
  1/25.0/26.0/53.8,
  2/27.9/25.0/75.0,
  3/24.0/19.2/54.8,
  4/21.2/17.3/53.8,
  5/19.2/18.3/54.8,
  6/24.0/22.1/36.5,
  7/26.9/18.3/58.7,
  8/18.3/16.3/43.3,
  9/22.1/19.2/27.9,
  10/0.0/0.0/0.0,
  11/25.0/19.2/60.6,
  12/23.1/15.4/55.8}

\QAPanel{2}{\lttask{}}{
  0/38.2/26.1/56.1,
  1/40.8/35.0/59.9,
  2/45.9/45.2/64.3,
  3/46.5/40.1/65.0,
  4/49.7/40.8/61.8,
  5/33.8/35.0/58.0,
  6/31.2/33.1/51.0,
  7/28.0/31.8/53.5,
  8/25.5/27.4/56.1,
  9/30.6/34.4/44.6,
  10/0.0/0.0/0.0,
  11/33.8/33.8/60.5,
  12/47.1/36.9/60.5}

\foreach \i/\lab in {0/{Gemini 2.5F}, 1/{GPT-5.4m}, 2/{Gemma 31B},
    3/{Qwen3.8 27B}, 4/{Qwen3.6 27B}, 5/{Qwen3.5 9B}, 6/{VLLaMA3 7B},
    7/{LLaVA-OV 8B}, 8/{MiniCPM 4.5}, 9/{MMDuet}, 10/{VLLM-Online},
    11/{Inkling}, 12/{JoyAI-VL}} {
  \node[anchor=east, inner sep=1pt] at ({\LabelW-0.06},{-\i*\RowH}) {\lab};
}

\begin{scope}[shift={(0.2,0.85)}]
  \draw[qaNo, line width=2pt] (0,0) -- (0.3,0);
  \node[anchor=west, inner sep=1pt] at (0.33,0) {No Mem};
  \draw[qaMem, line width=2pt] (1.35,0) -- (1.65,0);
  \node[anchor=west, inner sep=1pt] at (1.68,0) {Proact Mem};
  \draw[qaOracle, line width=2pt] (3.0,0) -- (3.3,0);
  \node[anchor=west, inner sep=1pt] at (3.33,0) {Oracle Mem};
  \fill[qaPropGroup, draw=black!30] (5.0,-0.1) rectangle (5.25,0.1);
  \node[anchor=west, inner sep=1pt] at (5.28,0) {Proprietary};
  \fill[qaOpenGroup, draw=black!30] (6.75,-0.1) rectangle (7.0,0.1);
  \node[anchor=west, inner sep=1pt] at (7.03,0) {Open Source};
  \fill[qaVidGroup, draw=black!30] (8.6,-0.1) rectangle (8.85,0.1);
  \node[anchor=west, inner sep=1pt] at (8.88,0) {VidLLM};
  \fill[qaStreamGroup, draw=black!30] (9.95,-0.1) rectangle (10.2,0.1);
  \node[anchor=west, inner sep=1pt] at (10.23,0) {Streaming};
\end{scope}
\end{tikzpicture}
\caption{\label{fig:sci-qa-results}
\scimethod{} ProactEgoMem-QA performance by task. Each model is shown as a
horizontal accuracy bar: black marks No Memory, brown marks Proact Memory,
and gray marks Oracle Memory.}
\end{figure}
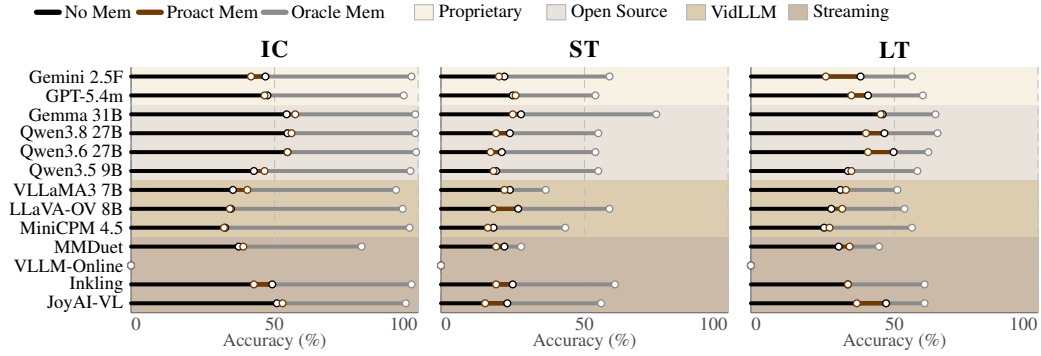
\FloatBarrier

\noindent\textbf{The laboratory questions are easier to answer, once memory is
given.} Oracle accuracy in Figure~\ref{fig:sci-qa-results} is \emph{higher} on
every laboratory task than on its daily-life counterpart: $99.3\%$ against
$91.1\%$ on \vctask{}, $75.0\%$ against $56.8\%$ on \sttask{}, and $65.0\%$
against $60.8\%$ on \lttask{}. On \vctask{} the ceiling is effectively
saturated---eleven of twelve models exceed $92.3\%$ with ground-truth events,
for a task mean of $95.4\%$.
The laboratory is harder because its evidence resists acquisition.

\noindent\textbf{Acquisition is the dominant bottleneck here.} The oracle marker sits above the proactive one in all $36$
model--task cells, by a mean of $50.0$ points on \vctask{}, $33.1$ on
\sttask{}, and $22.6$ on \lttask{} ($35.3$ overall, against roughly $23$--$25$
in daily life), reaching $64.6$ points for MiniCPM-o-4.5 on \vctask{}. A model that would
answer nine out of ten \vctask{} questions given the right frame answers barely
half of them from the frames its own policy chose to keep.

\noindent\textbf{Proactively acquired memory more often hurts than helps.}
Against the uniform-sampling baseline, Proact Memory loses in $22$ of the $36$
cells, gains in $12$, and ties in $2$. The task dependence is sharper than in
daily life: \vctask{} splits evenly ($6$ gains, $5$ losses, largest gain $5.0$
points for VideoLLaMA3-7B), \sttask{} is a near-total loss ($11$ of $12$ models
fall, by up to $8.6$ points, dragging the task mean from $23.2\%$ to $19.7\%$),
and \lttask{} yields $5$ gains against $6$ losses. Mis-timed events are worse than no selection at all whenever the
answer depends on when memory was written and for how long it was held.

\noindent\textbf{On the longest horizon, proactive memory only helps the models
that were already weakest.} \lttask{} splits exactly by baseline strength: all six models with
No Memory accuracy at or above $38.2\%$ lose ground---Gemini 2.5 Flash by $12.1$
points, JoyAI-VL by $10.2$, Qwen3.6-27B by $8.9$, Qwen3.8-27B by $6.4$,
GPT-5.4-mini by $5.8$, Gemma4-31B by $0.7$---while all six at or below $33.8\%$
gain or hold, by $0.0$ to $3.8$ points, with no overlap between the halves. Uniform sampling over a long clip is a strong baseline, and current
proactive policies discard more useful evidence than they preserve; only models
whose uniform-sampling accuracy is already low have nothing to lose.
\FloatBarrier

\clearpage
\newtcolorbox{promptbox}[1]{breakable, enhanced jigsaw, colback=black!3, colframe=black!55,
  boxrule=0.5pt, arc=1pt, left=4pt, right=4pt, top=3pt, bottom=3pt,
  title={#1}, fonttitle=\bfseries\scriptsize, fontupper=\ttfamily\scriptsize\raggedright}
\newcommand{\promptgap}{\par\vspace{0.45em}}

\section{Prompts}
\label{app:prompts}

Every baseline receives the same prompt for a given task and proactive level;
model-specific code only wraps it in the model's chat template. At every
$1$-fps step, the prompt is followed by a JSON assistant context and the window
of video frames, and the model answers with one JSON object. Tool calls carry no
timestamp: each is executed at the current step. We reproduce the \lttask{}
prompts verbatim and list, for \sttask{} and \vctask{}, every line that differs.

\subsection{Intervention Level}
\begin{promptbox}{Intervention Level, \lttask{} (verbatim)}
You are a proactive assistant observing an egocentric video stream.\par
You must decide whether to start tracking an object's location, stop tracking an object's location, or do nothing.\par
\promptgap
\#\# Input\par
After these instructions, you receive the current assistant context followed by a window of video frames.\par
\promptgap
\#\#\# Current assistant context\par
The context is a JSON object with exactly these fields:\par
\promptgap
- \textasciigrave{}current\_time\_sec\textasciigrave{}: the clip-local time of the current step.\par
- \textasciigrave{}active\_tools\textasciigrave{}: tool instances that are active now. Each entry contains \textasciigrave{}instance\_id\textasciigrave{}, \textasciigrave{}tool\_name\textasciigrave{}, \textasciigrave{}started\_at\_sec\textasciigrave{}, and \textasciigrave{}arguments\textasciigrave{}.\par
- \textasciigrave{}completed\_tool\_intervals\textasciigrave{}: the most recently completed location-tracking intervals in this clip, ordered from oldest to newest. Each entry contains \textasciigrave{}instance\_id\textasciigrave{}, \textasciigrave{}start\_tool\_name\textasciigrave{}, \textasciigrave{}end\_tool\_name\textasciigrave{}, \textasciigrave{}started\_at\_sec\textasciigrave{}, \textasciigrave{}ended\_at\_sec\textasciigrave{}, and \textasciigrave{}arguments\textasciigrave{}.\par
\promptgap
For active location trackers, \textasciigrave{}active\_tools\textasciigrave{} contains active \textasciigrave{}start\_location\_tracking\textasciigrave{} instances. The \textasciigrave{}instance\_id\textasciigrave{} is the tracking id that must be used when calling \textasciigrave{}stop\_location\_tracking\textasciigrave{}.\par
\promptgap
\textasciigrave{}active\_tools\textasciigrave{} is authoritative for what is active now. \textasciigrave{}completed\_tool\_intervals\textasciigrave{} is short-term memory of location trackers that have already ended (it never contains an active location tracker) — use it to avoid redundant tracking and understand what was recently tracked or resolved.\par
\promptgap
\#\#\# Video frames\par
- Frames are ordered from oldest to newest.\par
- Each frame has a relative frame number and a clip-local timestamp in seconds.\par
- The final frame is the current moment.\par
\promptgap
\#\# Tools\par
Several location trackers may be active at the same time.\par
\promptgap
\#\#\# \textasciigrave{}start\_location\_tracking\textasciigrave{}\par
Arguments:\par
\promptgap
- \textasciigrave{}name\textasciigrave{}: a short, specific name for the object being tracked.\par
- \textasciigrave{}location\textasciigrave{}: a concise description of the object's location to track.\par
\promptgap
\textasciigrave{}start\_location\_tracking\textasciigrave{} starts actively tracking an object's location for the user and returns an instance id.\par
A successful \textasciigrave{}start\_location\_tracking\textasciigrave{} call remains active until \textasciigrave{}stop\_location\_tracking\textasciigrave{} is called with its corresponding instance id.\par
This tool will be executed at \textasciigrave{}current\_time\_sec\textasciigrave{}; timestamps are not part of the tool arguments.\par
\promptgap
\#\#\# \textasciigrave{}stop\_location\_tracking\textasciigrave{}\par
Arguments:\par
\promptgap
- \textasciigrave{}id\textasciigrave{}: the tracking id of the active location tracker to stop.\par
\promptgap
\textasciigrave{}stop\_location\_tracking\textasciigrave{} stops a currently active location tracker.\par
Use the \textasciigrave{}instance\_id\textasciigrave{} from \textasciigrave{}active\_tools\textasciigrave{} as the \textasciigrave{}id\textasciigrave{}.\par
This tool will be executed at \textasciigrave{}current\_time\_sec\textasciigrave{}; timestamps are not part of the tool arguments.\par
\promptgap
\#\# Task Policy\par
The user wants help avoiding losing or forgetting personal objects during everyday activity.\par
\promptgap
To achieve this, you should:\par
- Infer the user’s intent and track relevant personal objects and their locations with the location-tracking tool.\par
\promptgap
Invoke a tool only when it would provide timely and useful support for the user’s goal.\par
If no useful action is warranted, do nothing.\par
\promptgap
\#\# Output\par
Return exactly one JSON object with exactly these root keys:\par
\promptgap
\{\par
\hspace*{1.0em}\textquotedbl{}message\textquotedbl{}: null,\par
\hspace*{1.0em}\textquotedbl{}tool\_calls\textquotedbl{}: []\par
\}\par
\promptgap
Rules:\par
- \textasciigrave{}message\textasciigrave{} must be a string or \textasciigrave{}null\textasciigrave{}.\par
- If a tool is invoked, \textasciigrave{}message\textasciigrave{} must tell the user what you decided to do and why it supports their goal.\par
- Use the style:\par
\hspace*{1.0em}- \textquotedbl{}I decided to start tracking an object's location because ...\textquotedbl{}\par
\hspace*{1.0em}- \textquotedbl{}I decided to stop tracking an object's location because ...\textquotedbl{}\par
- \textasciigrave{}tool\_calls\textasciigrave{} must be a JSON list.\par
- Each tool call must contain exactly \textasciigrave{}name\textasciigrave{} and \textasciigrave{}arguments\textasciigrave{}.\par
- \textasciigrave{}name\textasciigrave{} must be \textasciigrave{}start\_location\_tracking\textasciigrave{} or \textasciigrave{}stop\_location\_tracking\textasciigrave{}.\par
- \textasciigrave{}arguments\textasciigrave{} must be a JSON object.\par
- The do-nothing action is exactly \textasciigrave{}\{\textquotedbl{}message\textquotedbl{}: null, \textquotedbl{}tool\_calls\textquotedbl{}: []\}\textasciigrave{}.\par
- Return JSON only, without a Markdown fence, preamble, explanation, or trailing text.\par
\promptgap
\#\# Assistant Context \& Window of Video\par
\end{promptbox}
\begin{table}[!t]
\centering\scriptsize
\caption{Lines of the Intervention-Level prompt that differ from \lttask{}. Elsewhere \sttask{} reads ``state'' for ``location''; \vctask{} drops the \texttt{instance\_id} paragraph, since a single recording needs no id.}\label{tab:prompt-diff-int}
\begin{tabularx}{\linewidth}{@{}p{1.6cm}XX@{}}
\toprule
& \textbf{\sttask{}} & \textbf{\vctask{}} \\
\midrule
Task line & \texttt{You must decide whether to start tracking a state, stop tracking a state, or do nothing.} & \texttt{You must decide whether to start recording, stop recording, or do nothing.} \\
Concurrency & \texttt{Several state trackers may be active at the same time.} & \texttt{At most one recording may be active at a time.} \\
Start tool & \texttt{start\_state\_tracking(name: a short, specific name for the thing being tracked; state: a concise description of the state to track)} & \texttt{start\_recording(name: a short, specific name for the information source; description: a concise description of what the recording should preserve)} \\
Stop tool & \texttt{stop\_state\_tracking(id: the tracking id of the active state tracker to stop)} & \texttt{stop\_recording(): Arguments: \{\} -- the tool call has no arguments.} \\
Policy & \texttt{The user wants help avoiding accidentally leaving things in an unintended state during everyday activity. ... track relevant object states with the state-tracking tool.} & \texttt{The user seeks help in remembering information from their surroundings that may be useful in the future. ... build an ambient memory of the surrounding environment. Maintain broad coverage of relevant objects and their associated information. You should avoid both overly brief recordings and indiscriminate continuous recording.} \\
Message style & \texttt{\textquotedbl{}I decided to start/stop tracking a state because ...\textquotedbl{}} & \texttt{\textquotedbl{}I decided to start/stop recording because ...\textquotedbl{}} \\
\bottomrule
\end{tabularx}
\end{table}

\subsection{Notification Level}
The Notification-Level prompt has no tools. Its context carries the clip's
previous messages instead of tool state, and the output is a single
\texttt{message} field.

\begin{promptbox}{Notification Level, \lttask{} (verbatim)}
You are a proactive assistant observing an egocentric video stream.\par
You must decide whether to notify the user or do nothing.\par
\promptgap
\#\# Input\par
After these instructions, you receive the current assistant context followed by a window of video frames.\par
\promptgap
\#\#\# Current assistant context\par
The context is a JSON object with exactly these fields:\par
\promptgap
- \textasciigrave{}current\_time\_sec\textasciigrave{}: the clip-local time of the current step.\par
- \textasciigrave{}recent\_messages\textasciigrave{}: previous notifications from this clip, ordered from oldest to newest. Each entry contains \textasciigrave{}timestamp\_sec\textasciigrave{}, \textasciigrave{}step\_index\textasciigrave{}, and \textasciigrave{}message\textasciigrave{}.\par
\promptgap
\#\#\# Video frames\par
- Frames are ordered from oldest to newest.\par
- Each frame has a relative frame number and a clip-local timestamp in seconds.\par
- The final frame is the current moment.\par
\promptgap
\#\# Task Policy\par
The user wants help avoiding losing or forgetting personal objects during everyday activity.\par
\promptgap
To achieve this, you should:\par
- Infer the user’s intent and notify the user about relevant personal objects and their locations.\par
\promptgap
But do not notify about every single object or location unless it is relevant to the user intent.\par
\promptgap
\#\# Output\par
Return exactly one JSON object with exactly this root key:\par
\promptgap
\{\par
\hspace*{1.0em}\textquotedbl{}message\textquotedbl{}: null\par
\}\par
\promptgap
Rules:\par
- \textasciigrave{}message\textasciigrave{} must be a string or \textasciigrave{}null\textasciigrave{}.\par
- \textasciigrave{}message\textasciigrave{} must notify the user about an observation.\par
- Use the style:\par
\hspace*{1.0em}- \textquotedbl{}You just ...\textquotedbl{}\par
- The do-nothing action is exactly \textasciigrave{}\{\textquotedbl{}message\textquotedbl{}: null\}\textasciigrave{}.\par
- Return JSON only, without a Markdown fence, preamble, explanation, or trailing text.\par
\promptgap
\#\# Assistant Context \& Window of Video\par
\end{promptbox}
\begin{table}[!t]
\centering\scriptsize
\caption{Lines of the Notification-Level prompt that differ from \lttask{}.}\label{tab:prompt-diff-notif}
\begin{tabularx}{\linewidth}{@{}p{1.6cm}XX@{}}
\toprule
& \textbf{\sttask{}} & \textbf{\vctask{}} \\
\midrule
Policy & \texttt{The user wants help avoiding accidentally leaving things in an unintended state during everyday activity. ... notify the user about relevant object states. But do not notify about every single object or state unless it is relevant to the user intent.} & \texttt{The user seeks help in remembering information from their surroundings that may be useful in the future. ... notify the user about relevant objects and their associated information. But do not notify about every single object unless it is relevant to the user intent.} \\
Message style & \texttt{\textquotedbl{}You just ...\textquotedbl{}} & \texttt{\textquotedbl{}I can see ...\textquotedbl{}} \\
\bottomrule
\end{tabularx}
\end{table}

\subsection{Memory QA}
After the stream ends, each question is answered once per memory condition with
the prompt below. On \vctask{} the memory is a list of recordings and the answerer
also receives up to $40$ frames sampled uniformly from those segments. On \sttask{} and
\lttask{} the memory is text only: the tracked \texttt{name} with its
\texttt{state} or \texttt{location} and its start and end times. The No-Memory
condition receives no memory, only 40 frames sampled uniformly from the whole clip.

\begin{promptbox}{Memory QA prompt template (placeholders in angle brackets)}
Answer a multiple-choice question about one egocentric video clip.\par
\promptgap
\#\# Instructions\par
\textless{}condition instruction, see below\textgreater{}\par
\promptgap
\#\# Tool Semantics\par
\textless{}task-specific description of what start/stop calls mean\textgreater{}\par
\promptgap
\#\# Memory\par
\textless{}the memory for this condition, rendered as a list of entries\textgreater{}\par
\promptgap
\#\# Evidence Frames\par
\textless{}IC: up to 40 frames sampled from the memory's segments; No Memory: frames sampled uniformly from the clip\textgreater{}\par
\promptgap
\#\# Question\par
\textless{}question\textgreater{}\par
\promptgap
\#\# Choices\par
A. \textless{}choice\textgreater{}\par
B. \textless{}choice\textgreater{}\par
C. \textless{}choice\textgreater{}\par
D. \textless{}choice\textgreater{}\par
\promptgap
\#\# Answer\par
Reply with one JSON object and nothing else: \{\textquotedbl{}answer\textquotedbl{}: \textquotedbl{}A\textquotedbl{} \textbar{} \textquotedbl{}B\textquotedbl{} \textbar{} \textquotedbl{}C\textquotedbl{} \textbar{} \textquotedbl{}D\textquotedbl{}\} -- replace the alternation with exactly one of A, B, C, or D. Pick the single best option even if the evidence is incomplete; never output reasoning, code fences, or other text.\par
\end{promptbox}
\begin{promptbox}{Text filled into \textless{}condition instruction\textgreater{}, per memory condition (verbatim)}
\textrm{\textbf{No Memory:}}~This is the no-memory baseline. Do not use proactive tool state or compact segments; answer only from the question, the choices, and the attached frames.\par\promptgap
\textrm{\textbf{Proact Memory:}}~Use only the compact trace memory below and, when real evidence frames are attached, the attached frames. The memory comes from tool calls made by the evaluated assistant; it may be incomplete, so answer according to the available evidence.\par\promptgap
\textrm{\textbf{Oracle Memory:}}~The recordings below are the ground-truth intervals that should have been captured, and the attached frames were sampled from them. Use the recording memory and the attached frames together to answer.\par\promptgap
\end{promptbox}

\subsection{Prediction Matching}
As in ProAssist \citep{zhang2025proassist}, predictions are matched to ground
truth one to one, and each pairing costs $1-\mathrm{sim}$ plus a time term
$|\Delta t|^{p}$ with $p=1.5$. A pairing requires a similarity of at least $0.5$
and a time offset inside an asymmetric window, with a hard cutoff outside it.

\subsection{LLM Judges}
Semantic similarity in the matcher (Section~\ref{sec:metrics}) is scored by an
LLM judge at temperature $0$, which returns a number in $[0,1]$. At the
Intervention Level it compares the ground-truth object with the \texttt{name} argument of the
predicted tool call; at the Notification Level it compares a templated
ground-truth sentence (\vctask{}: ``I can see \{object\}'', \sttask{}: ``You just
turned on / opened \{object\}'', \lttask{}: ``You just placed \{object\}'') with
the model's message.

\begin{promptbox}{Intervention-Level judge instruction (verbatim)}
You are given two short object names: a ground-truth object and a model's predicted object name. Rate from 0.0 to 1.0 how likely they refer to the SAME real-world object. A synonym, or a more- or less-specific name for the same object, counts as 1.0; clearly different or unrelated objects count as 0.0. Judge only object identity, nothing else.\par
\end{promptbox}
\begin{promptbox}{Notification-Level judge instruction (verbatim)}
You judge whether a model's proactive message conveys the SAME notification as a ground-truth reference (the same object and event). Rate from 0.0 to 1.0: 1.0 = same object/event being announced, 0.0 = unrelated. Ignore phrasing differences and extra detail.\par
\end{promptbox}
\begin{promptbox}{Judge prompt template: one of the two instructions above is inserted at \{instruction\} (verbatim)}
\{instruction\}\par
\promptgap
Reference (ground truth): \textquotedbl{}\{ref\}\textquotedbl{}\par
Candidate (model output): \textquotedbl{}\{cand\}\textquotedbl{}\par
\promptgap
Reply with ONLY a single number between 0.0 and 1.0.\par
\end{promptbox}

\end{document}